\documentclass[twocolumn]{article} 

\usepackage{graphicx}
\usepackage{graphbox}
\usepackage{cite}
\usepackage{amssymb}
\usepackage{latexsym}
\usepackage{graphicx}
\usepackage{subfigure}
\usepackage{amsmath}
\usepackage{mathtools}
\usepackage{url}
\usepackage{pgf,tikz}
\usepackage{tikz}
\usetikzlibrary{spy}
\usepackage{xcolor}

\graphicspath{ {images/} }

\begin{document}
\sloppy

\title{Learning-based Framework for US Signals Super-resolution}%

\makeatletter
\onecolumn
{\fontsize{18pt}{20pt}\selectfont\bfseries\@title\par}
Simone Cammarasana
 \footnote{
\textbf{Simone Cammarasana}
CNR-IMATI, Via De Marini 6, Genova, Italy \\
simone.cammarasana@ge.imati.cnr.it
}
 , 
 Paolo Nicolardi
 \footnote{
 \textbf{Paolo Nicolardi}
Esaote S.p.A., Via E. Melen 77, Genova, Italy 
}
  , 
 Giuseppe Patan\`e
  \footnote{
 \textbf{Giuseppe Patan\`e} 
 CNR-IMATI, Via De Marini 6, Genova, Italy 
}

\makeatother

\begin{abstract}
This paper proposes a novel deep-learning framework for super-resolution ultrasound images and videos in terms of spatial resolution and line reconstruction. To this end, we up-sample the acquired low-resolution image through a vision-based interpolation method; then, we train a learning-based model to improve the quality of the up-sampling. We qualitatively and quantitatively test our model on different anatomical districts (e.g., cardiac, obstetric) images and with different up-sampling resolutions (i.e., 2X, 4X). Our method improves the PSNR median value with respect to SOTA methods of~$1.7\%$ on obstetric 2X raw images,~$6.1\%$ on cardiac 2X raw images, and~$4.4\%$ on abdominal raw 4X images; it also improves the number of pixels with a low prediction error of~$9.0\%$ on obstetric 4X raw images,~$5.2\%$ on cardiac 4X raw images, and~$6.2\%$ on abdominal 4X raw images.

The proposed method is then applied to the spatial super-resolution of 2D videos, by optimising the sampling of lines acquired by the probe in terms of the acquisition frequency. Our method specialises trained networks to predict the high-resolution target through the design of the network architecture and the loss function, taking into account the anatomical district and the up-sampling factor and exploiting a large ultrasound data set.  The use of deep learning on large data sets overcomes the limitations of vision-based algorithms that are general and do not encode the characteristics of the data. Furthermore, the data set can be enriched with images selected by medical experts to further specialise the individual networks. Through learning and high-performance computing, the proposed super-resolution is specialised to different anatomical districts by training multiple networks. Furthermore, the computational demand is shifted to centralised hardware resources with a real-time execution of the network’s prediction on local devices.

\end{abstract}
\twocolumn
\begin{figure*}[t]
\centering
\begin{tabular}{c}
\includegraphics[width=0.75\textwidth]{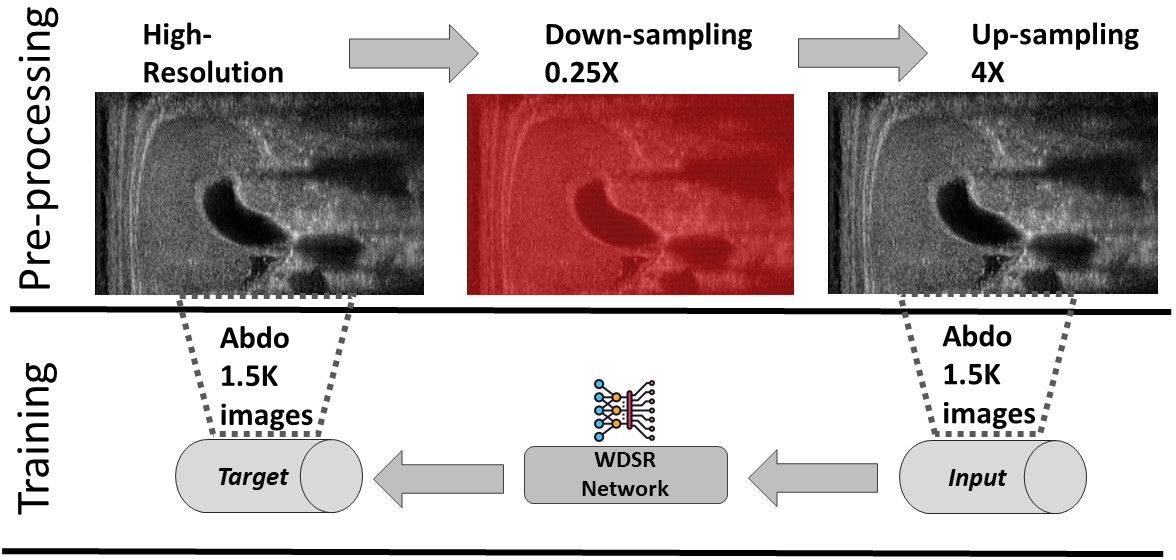}\\
\includegraphics[width=0.75\textwidth]{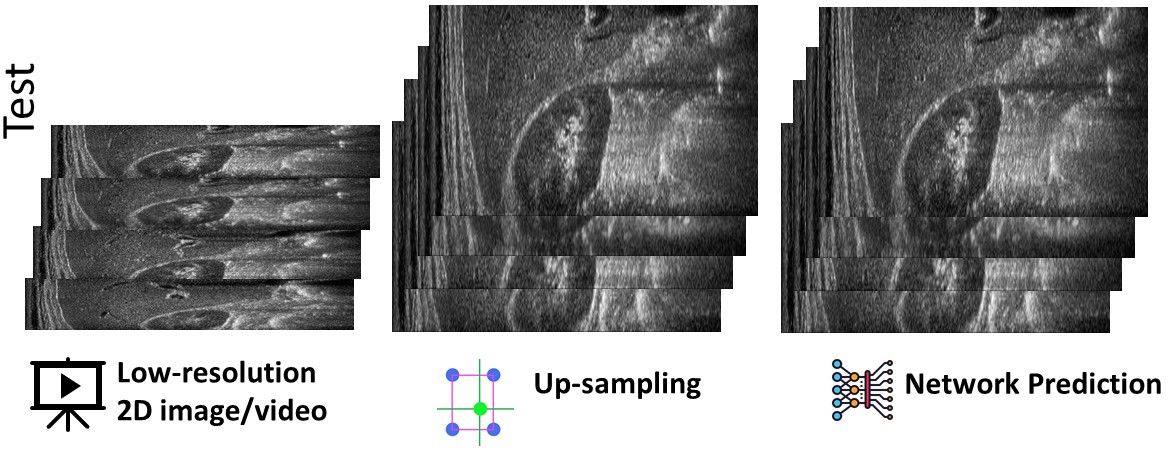} 
\end{tabular}
\caption{Proposed framework (Sect.~\ref{SEC:INTRODUCTION}): training of the learning-based model and spatial up-sampling of US videos. A high-resolution image is down-sampled by removing one line (highlighted in red) each two (0.5X) or four (0.25X) and then up-sampled through the selected interpolation algorithm. Up-sampled images and the corresponding high-resolution images are the input and target to train the neural network, respectively. For the test phase, low-resolution images are acquired during ultrasound acquisition (i.e., images with a reduced number of acquired beam lines); these images are up-sampled through the interpolation algorithm and the neural network predicts the final output that is expected to be similar to the unknown high-resolution target.\label{FIG:TEASER}}
\end{figure*}
\section{Introduction\label{SEC:INTRODUCTION}}
\emph{Ultrasound} (US, for short) acquisition applies high-frequency sound waves to visualise soft tissues and internal organs, and support medical diagnosis for muscle-skeletal, cardiac, and obstetrical diseases. US acquisition has many advantages with respect to magnetic resonance and tomographies, such as its portability, cheapness, and non-invasiveness. Furthermore, its real-time acquisition provides instantaneous feedback to the physician, e.g., during regional anaesthesia. Through US videos, the physician analyses the temporal variation of an anatomical feature (e.g., the movement of a muscle, the volume of the ventricle), which can be generated either by the shift of the probe or by the movement of the anatomical part. 2D US videos are acquired through 2D probes, which capture sequences of images at a given frequency.

The resolution of each image is affected by the required frequency of the video, since some anatomical districts (e.g., cardiac) require a high acquisition frequency, to accurately acquire the behaviour of anatomical features that quickly change over time. For example, US videos of the cardiac district require high temporal frequency, since they need to acquire anatomical parts (e.g., the mitral valve) that move quickly over time; a higher temporal frequency allows the radiologist to better characterise the movement of the anatomical part.

Our goal is the design of a novel deep-learning framework for the super-resolution of 2D US images, by increasing the image resolution and reconstructing non-acquired beamlines. We define the non-acquired beam lines as the intermediate lines to those acquired by the probe. These intermediate lines are not acquired to increase the acquisition time frequency but are approximated by the super-resolution scheme. Applying our approach to US videos with a low spatial resolution and a high frequency (e.g., for the cardiac district), we can generate high-frequency 2D US video with an increased spatial resolution of each frame, thus overcoming the main limits of current US probes, whose spatial resolution decreases as the acquisition frequency increases. Acquiring a 2D video with a low spatial frequency of the single image (i.e., each frame), our method reconstructs the spatially high-resolution video in real-time.

First, we compare several state-of-the-art up-sampling algorithms (Sect.~\ref{SEC:RELATEDWORK}) and identify the best method in terms of quantitative metrics and visual evaluation. Then, we train a neural network to improve the results of the up-sampling to match the target image (i.e., the high-resolution image). Our network does not perform the interpolation of the missing lines; in fact, this task is already performed by up-sampling. In contrast, our network learns how to transform the up-sampled lines into the target lines. To improve the quality of the up-sampling, we train multiple networks, each one specialised to the input anatomical district (e.g., cardiac, abdominal) and its low-resolution image (e.g., 0.5X, 0.25X). This specialisation improves the quality of the up-sampling since we specialise the network to a specific prediction. The execution time of the super-resolution depends on the up-sampling and the network prediction; the prediction is achieved in real-time on standard medical hardware. We summarise the proposed framework (Fig.~\ref{FIG:TEASER}), where we generate the data sets within the pre-processing phase, the learning-based models within the training phase, and the real-time super-resolution prediction within the test phase.
\begin{figure*}[t]
\centering
\begin{tabular}{c}
\includegraphics[width=0.96\textwidth]{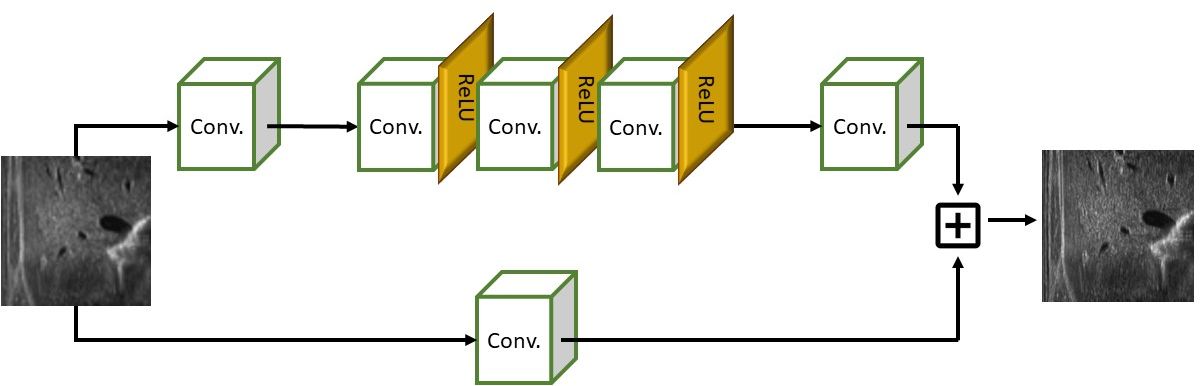}\\
\end{tabular}
\caption{Network's architecture with Convolution layers (Conv.) and ReLU activation functions (ReLU).\label{FIG:NETWORK}}
\end{figure*}

As the main contribution (Sect.~\ref{SEC:METHOD}), we propose a novel learning-based architecture that accounts for convolutional layers and rectified linear unit (ReLU) activation functions (Fig.~\ref{FIG:NETWORK}) and improves the \emph{Wide Activation for Efficient and Accurate Image Super-Resolution} (WDSR)\cite{yu2020wide}. The kernel size is selected according to the dimension of the low-resolution image to guarantee that at least two original lines (i.e., two lines that are acquired by the probe) are always included in the convolution operation. Then, we modify the loss function to improve the visual accuracy of the prediction. Our logarithmic-based loss includes only up-sampled lines, excluding lines acquired by the probe. 
 
The proposed approach is general in terms of the building blocks of the framework; in fact, we can select different up-sampling algorithms, e.g., \emph{Single Image Super Resolution} (SISR)~\cite{peleg2014statistical}, \emph{Enhanced Super Resolution Generative Adversarial Network} (ESRGAN)~\cite{wang2018esrgan} and deep learning architectures, e.g., \emph{Pix2Pix}~\cite{isola2017image} and \emph{VGG19}~\cite{simonyan2014very}. As experimental validation (Sect.~\ref{SEC:RESULTS}), we perform a quantitative and qualitative evaluation of our framework on a large collection of US images acquired from different anatomical (e.g., muscle-skeletal, obstetric, abdominal) districts. Then, we apply our method to the spatial super-resolution of US 2D US videos, and we evaluate the effects of denoising the raw images as pre-processing of our framework. Finally, we present a discussion on main outcomes (Sect.~\ref{SEC:DISCUSSION}), conclusions, and future work (Sect.~\ref{SEC:CONCLUSIONS}).
Trained models and training/test code are available at \url{https://github.com/cammarasana123/US-SuperResolution}.
\begin{figure*}[t]
\centering
\begin{tikzpicture}[spy using outlines={rectangle,magnification=2,size=1cm,width=1cm,height=1cm,every spy on node/.append style={thick}}] 
\node {
\begin{tabular}{c|cc}
Target & Input & Prediction \\
\includegraphics[align=c,width=0.3\textwidth]{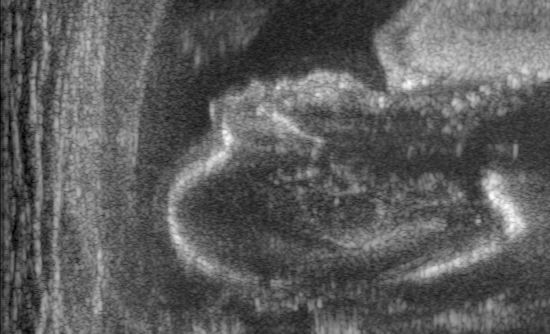} &
\includegraphics[width=0.3\textwidth]{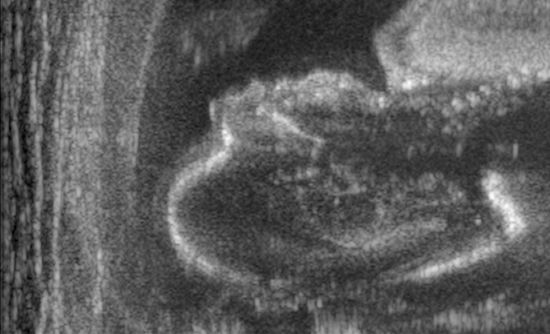} &
\includegraphics[width=0.3\textwidth]{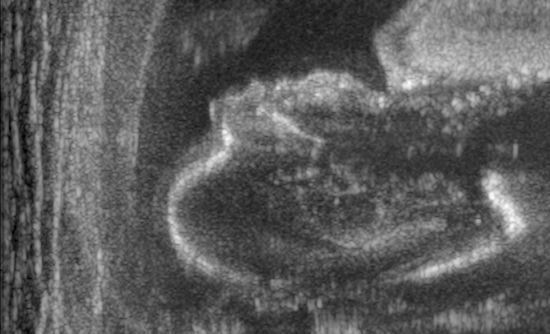} \\ [-13mm]
&
\includegraphics[width=0.3\textwidth]{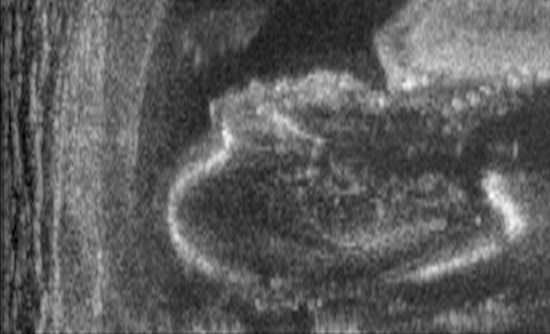} &
\includegraphics[width=0.3\textwidth]{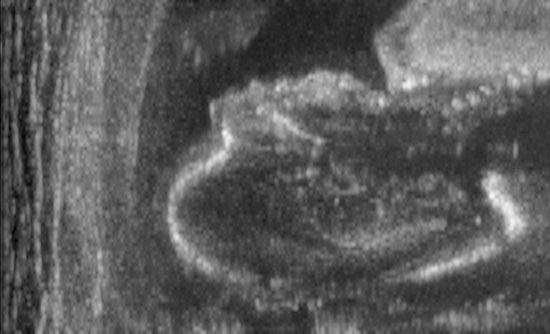} 
\end{tabular}
};
\spy [red] on (-5.3,0.1) in node [left] at (-4.8,0.1); 
\spy [red] on (0.35,-1.75) in node [left] at (0.85,-1.75); 
\spy [red] on (5.97,-1.75) in node [left] at (6.47,-1.75); 
\spy [red] on (0.35,1.55) in node [left] at (0.85,1.55); 
\spy [red] on (5.97,1.55) in node [left] at (6.47,1.55); 
\end{tikzpicture}
\caption{Prediction on the raw images of the obstetric district: 2X up-sampling (first line); 4X up-sampling (second line). The input image (i.e., the input of the neural network) represents the outcome of the up-sampling algorithm; the prediction represents the output of the neural network, which aims at improving the approximation of the target image (i.e., the high-resolution image). The red squares represent a magnification of a portion of the image, to better visualise the results of the prediction of the network.\label{FIG:OBNETWORK}}
\end{figure*}
\section{Related work\label{SEC:RELATEDWORK}}
\paragraph{Learning-based US super-resolution}
The main novelties of the enhanced deep super-resolution network~\cite{lim2017enhanced} are a simplification of the conventional residual network architectures and a multi-scale super-resolution network that reduces the model size. Exploiting the sparsity of the signal in the Fourier domain, the interpolation of missing data~\cite{yoon2018deep} allows reconstructing the high-resolution ultrasound (HR US) image with a low computational cost. A \emph{super-resolution generative adversarial network} (SRGAN)~\cite{ledig2017photo} applies a deep residual network with skip-connection and a perceptual loss between generated and target images. The reduction of artefacts of the previous method is addressed by the Enhanced SRGAN~\cite{wang2018esrgan}, which improves the network architecture, the adversarial and the perceptual loss, removes the batch normalisation layer, and applies the residual scaling and smaller initialisation values. 

The perceptual quality of ESRGAN is improved by the ESRGAN+ method~\cite{rakotonirina2020esrgan+} through a novel \emph{Residual-in-Residual Dense Residual} block, which increases the network capacity without affecting its complexity. The application of the SRGAN to US images~\cite{choi2018deep} preserves both the anatomical structures and the speckle noise pattern, thus improving the perceptual quality of the upsampled images. Dilated convolution~\cite{lu2018unsupervised} extracts the internal recurrence information from the test image; this method upsamples low-resolution (LR) images when LR-HR examples are reduced. Fully convolutional U-net~\cite{van2019deep} obtains high-resolution vascular images from high-density contrast-enhanced US signals. In~\cite{temiz2020super}, the deep learning method exploiting feature extraction blocks, repeating blocks, and upsampling layers apply an up-sampling factor in the range 2-8. A Self-supervised CycleGAN~\cite{liu2021perception} only requires the LR US image,  and generates perceptually consistent up-sampling results. Combining CycleGAN, two-stage GAN, and the zero-shot super resolution~\cite{ding2021ultrasound}, it is possible to obtain super-resolution images with low blurring artefacts.
\begin{figure*}[t]
\centering
\begin{tikzpicture}[spy using outlines={rectangle,magnification=2,size=1cm,width=1cm,height=1cm,every spy on node/.append style={thick}}] 
\node {
\begin{tabular}{c|cc}
Target & Input & Prediction \\
\includegraphics[align=c,width=0.3\textwidth]{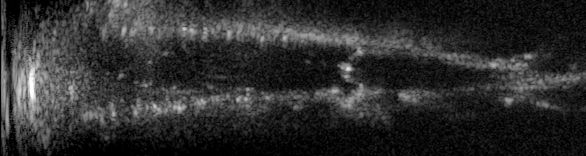} &
\includegraphics[width=0.3\textwidth]{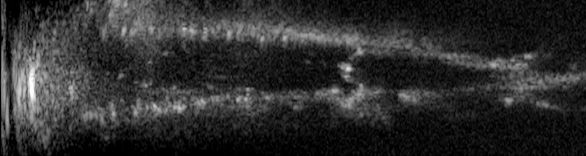} &
\includegraphics[width=0.3\textwidth]{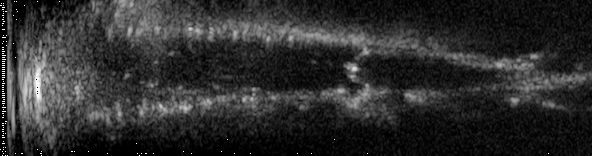} \\ [-5mm]
&
\includegraphics[width=0.3\textwidth]{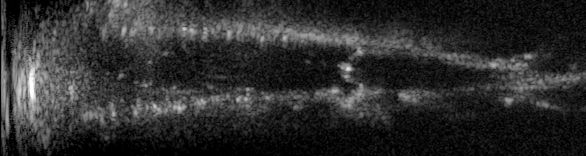} &
\includegraphics[width=0.3\textwidth]{cardio_4X_raw_target_27.jpg} 
\end{tabular}
};
\spy [red] on (-5.1,0.10) in node [left] at (-4.6,0.10); 
\spy [red] on (0.55,-0.75) in node [left] at (1.05,-0.75); 
\spy [red] on (6.17,-0.75) in node [left] at (6.67,-0.75); 
\spy [red] on (0.55,0.70) in node [left] at (1.05,0.70); 
\spy [red] on (6.17,0.70) in node [left] at (6.67,0.70); 
\end{tikzpicture}
\caption{Prediction on raw images of the cardiac district: 2X up-sampling (first line); 4X up-sampling (second line). See also Fig.~\ref{FIG:OBNETWORK}.\label{FIG:CARDIONETWORK}}
\end{figure*}
\paragraph{Vision-based US super-resolution}
Learning-based methods suffer from artefacts and blurring when dealing with noisy signals. Several vision-based methods have been proposed, through the years. The interpolating up-sampling with cubic kernels~\cite{keys1981cubic} offers high accuracy with low computational cost, through appropriate boundary conditions and constraints on the kernel functions. In~\cite{alessandrini2011restoration}, a novel deconvolution-based method applies the maximum a posteriori estimation to the restoration of the tissue response and is validated with several tissue-mimicking phantoms with specific scatterer concentrations. The \emph{Alternating Direction Method of Multipliers}~\cite{ng2010solving} is applied to the super-resolution of US images including deblurring and denoising~\cite{morin2012alternating} through a combination of~$\ell_1$ and~$\ell_2$ minimisation. In~\cite{yu2012envelope}, a deconvolution method models the envelope radiofrequency and point spread function is robust to noise and does not require the knowledge of the centre frequency of the acquired signal. Assuming a Gaussian distribution for both the unknown signal to be restored and the point spread function, in~\cite{zhao2015joint} the reconstructed image is built through a posterior model with hybrid Gibbs sampling~\cite{geman1984stochastic}. The properties of the decimation matrix in the Fourier domain~\cite{zhao2016single} are exploited to solve the super-resolution problem with a~$\ell_p$-norm regulariser, with~$p \in [1,2]$. The envelope of radio frequency signal~\cite{khavari2018non} applies repetitive data in the non-local neighbourhood of samples. 
\begin{figure*}[t]
\centering
\begin{tikzpicture}[spy using outlines={rectangle,magnification=2,size=1cm,width=1cm,height=1cm,every spy on node/.append style={thick}}] 
\node {
\begin{tabular}{c|cc}
Target & Input & Prediction \\
\includegraphics[align=c,width=0.3\textwidth]{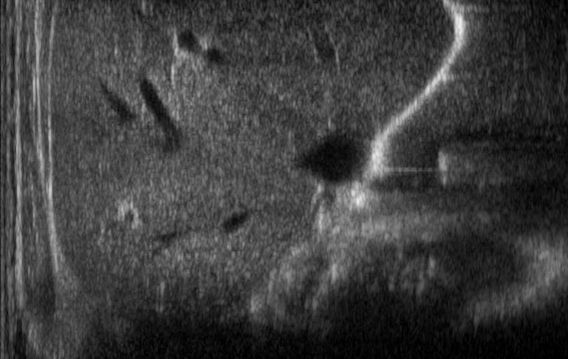} &
\includegraphics[width=0.3\textwidth]{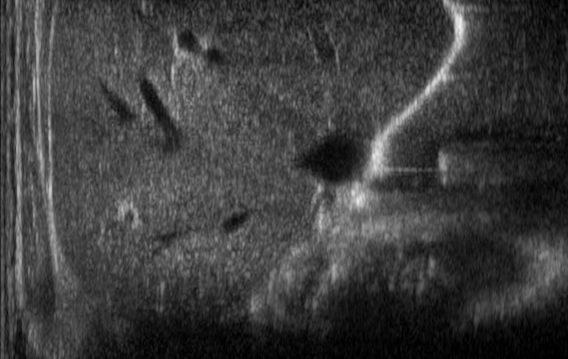} &
\includegraphics[width=0.3\textwidth]{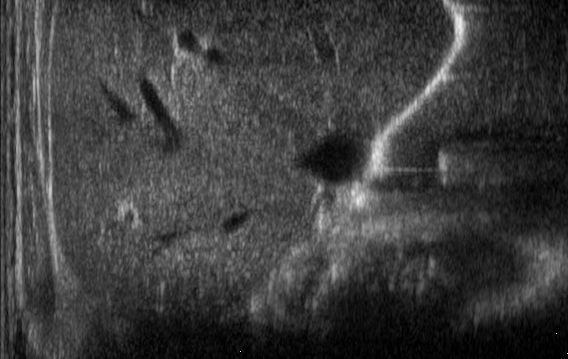} \\ [-13mm]
&
\includegraphics[width=0.3\textwidth]{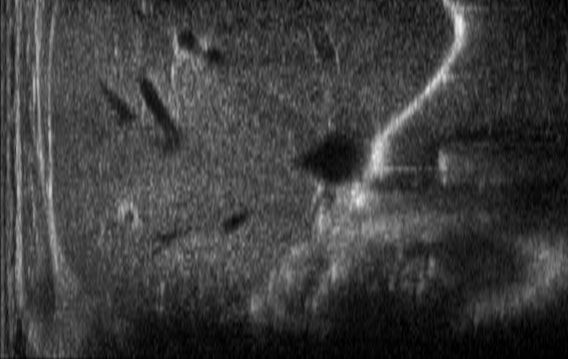} &
\includegraphics[width=0.3\textwidth]{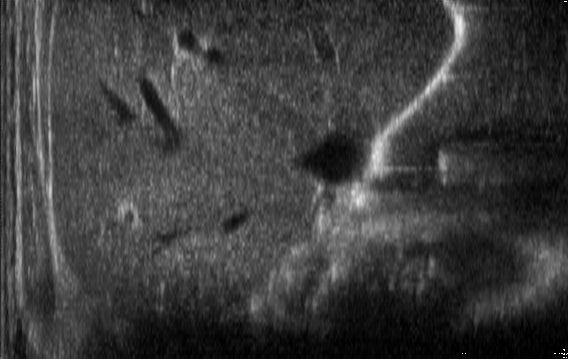} 
\end{tabular}
};
\spy [red] on (-6.1,-0.3) in node [left] at (-5.6,-0.3); 
\spy [red] on (-0.45,-2.3) in node [left] at (0.05,-2.3); 
\spy [red] on (5.20,-2.3) in node [left] at (5.70,-2.3); 
\spy [red] on (-0.45,1.30) in node [left] at (0.05,1.30); 
\spy [red] on (5.20,1.30) in node [left] at (5.70,1.30); 
\end{tikzpicture}
\caption{Prediction on the raw images of the abdominal district: 2X up-sampling (first line); 4X up-sampling (second line). See also Fig.~\ref{FIG:OBNETWORK}.\label{FIG:ABDONETWORK}}
\end{figure*}
\paragraph{Device-based US super-resolution}
The second harmonic image~\cite{taxt2004superresolution} contains less noise and blur than the first harmonic image. Furthermore, the lateral resolution is increased, as the harmonic pulse is auto-focused because the higher harmonics are generated in the centre of the beam. Then, the image super-resolution is achieved by combining the first and second harmonic images. Both spatial and temporal deconvolution operations~\cite{lingvall2004method} are achieved by accounting for the transmit and receive processes, electrical transducer characteristics, and transmit focusing laws. Combining phase-contrast imaging, angular spectral decomposition, and a super-resolution reconstruction technique~\cite{clement2005superresolution}, it is possible to recover the location and dimensions of objects smaller than the imaging wavelength. The reconstruction through generalised Tikhonov regularisation~\cite{lavarello2006regularized} is evaluated as a function of transmit-receive bandwidth and a focal number of the transducer, by comparing the results with traditional B-mode imaging. The \emph{Time-domain Optimized Nearfield Estimator}~\cite{viola2007time} assumes an observation model based on the superposition of spatial responses; then, a maximum a-posteriori estimation finds the distribution and amplitude of hypothetical targets that match the observed data with minimal target energy. As a further improvement, the \emph{Diffuse Time-domain Optimised Near-field Estimator}~\cite{ellis2010super} represents each hypothetical target in the system model as a diffuse region of targets rather than a single discrete target, thus inducing a better signal approximation. The cellular microscopy technique of multi-focal imaging\cite{diamantis2017super} is applied to localise the unique position of the scatterer of the signal; three foci receive multiple overlapping curves, and a maximum likelihood estimation allows the identification of the source of the scatter.
\begin{figure*}[t]
\centering
\begin{tabular}{cc}
2X & 4X \\ \hline
\includegraphics[width=0.45\textwidth]{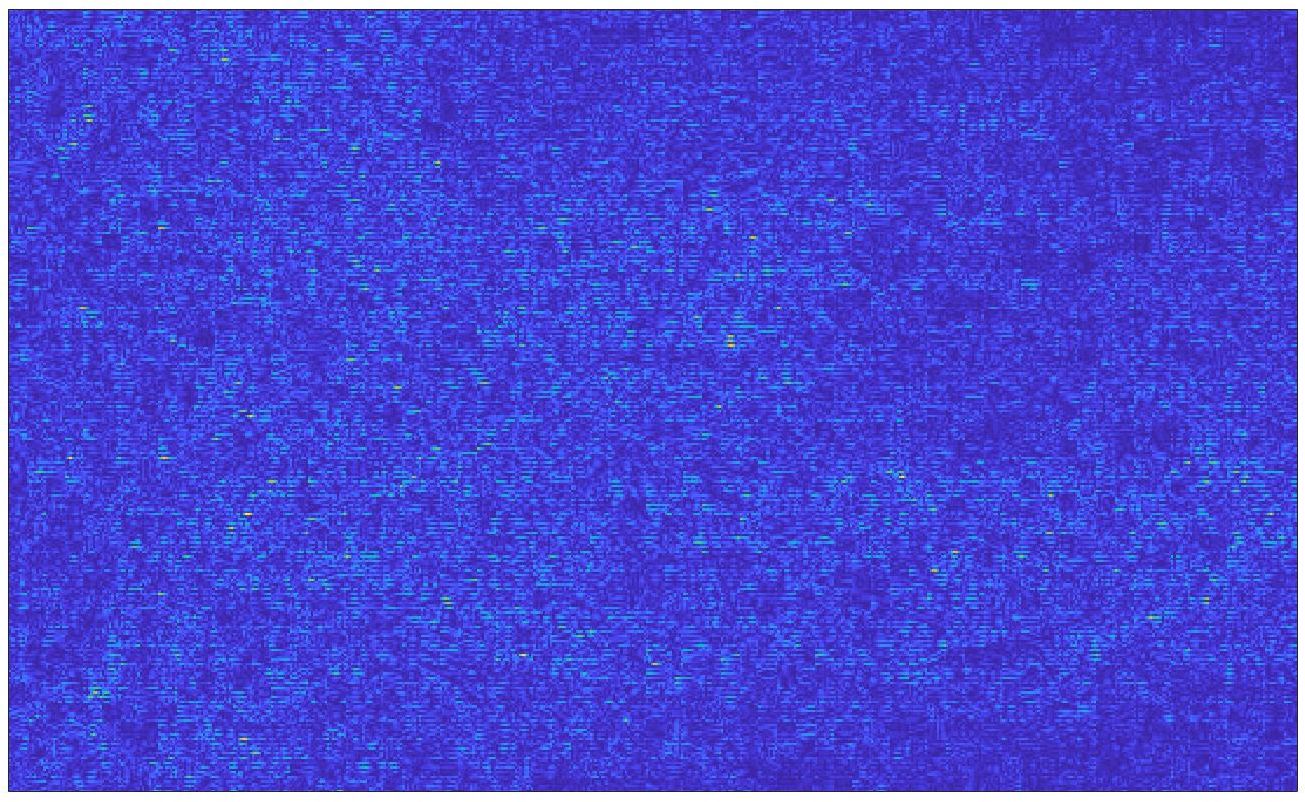} &
\includegraphics[width=0.45\textwidth]{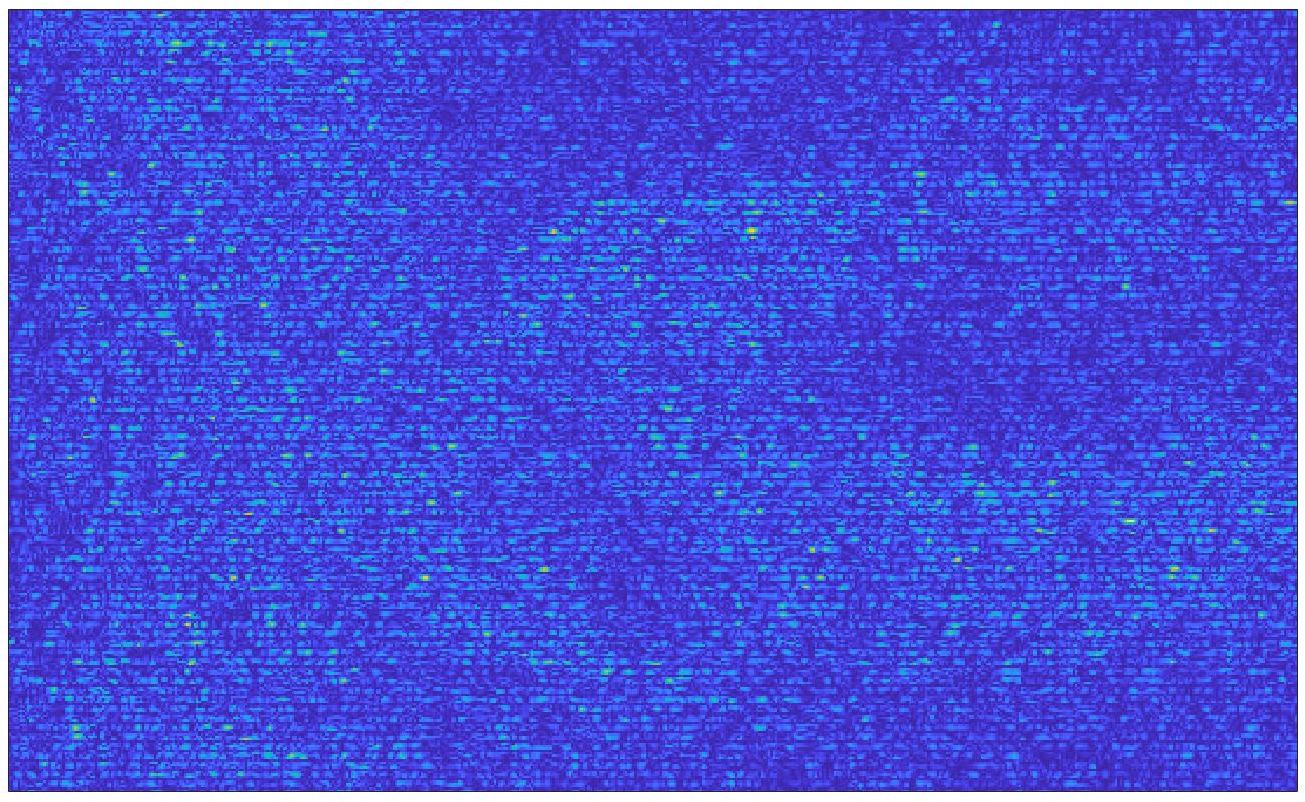} \\ [-0mm]
Max. error: 30 & Max. error: 41 \\ \hline
\includegraphics[width=0.45\textwidth]{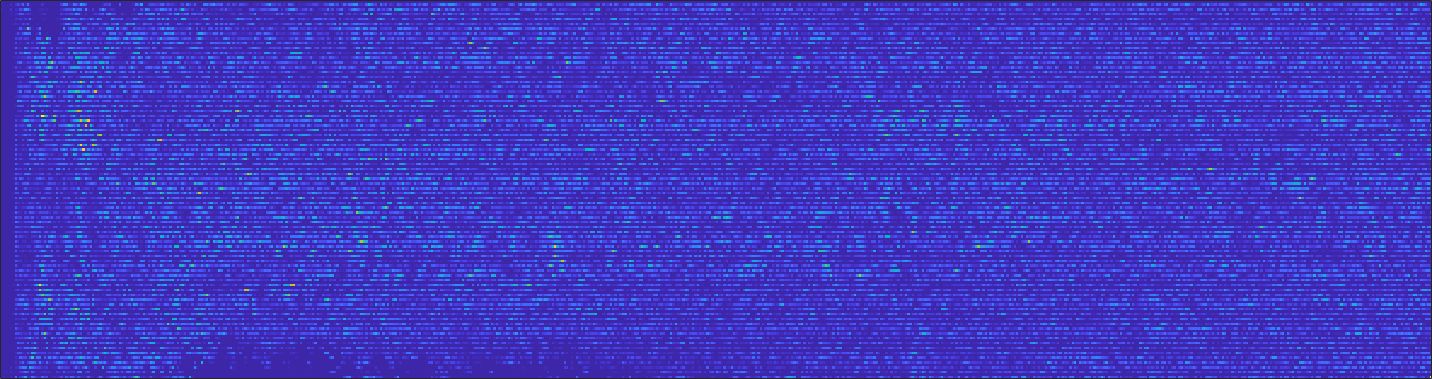} &
\includegraphics[width=0.45\textwidth]{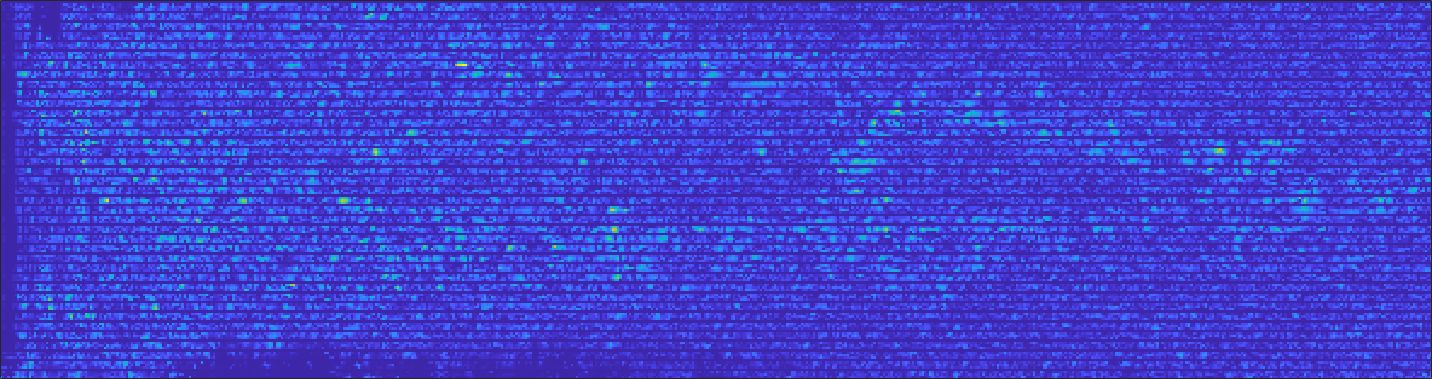} \\ [-0mm]
Max. error: 29 & Max. error: 46 \\ \hline
\includegraphics[width=0.45\textwidth]{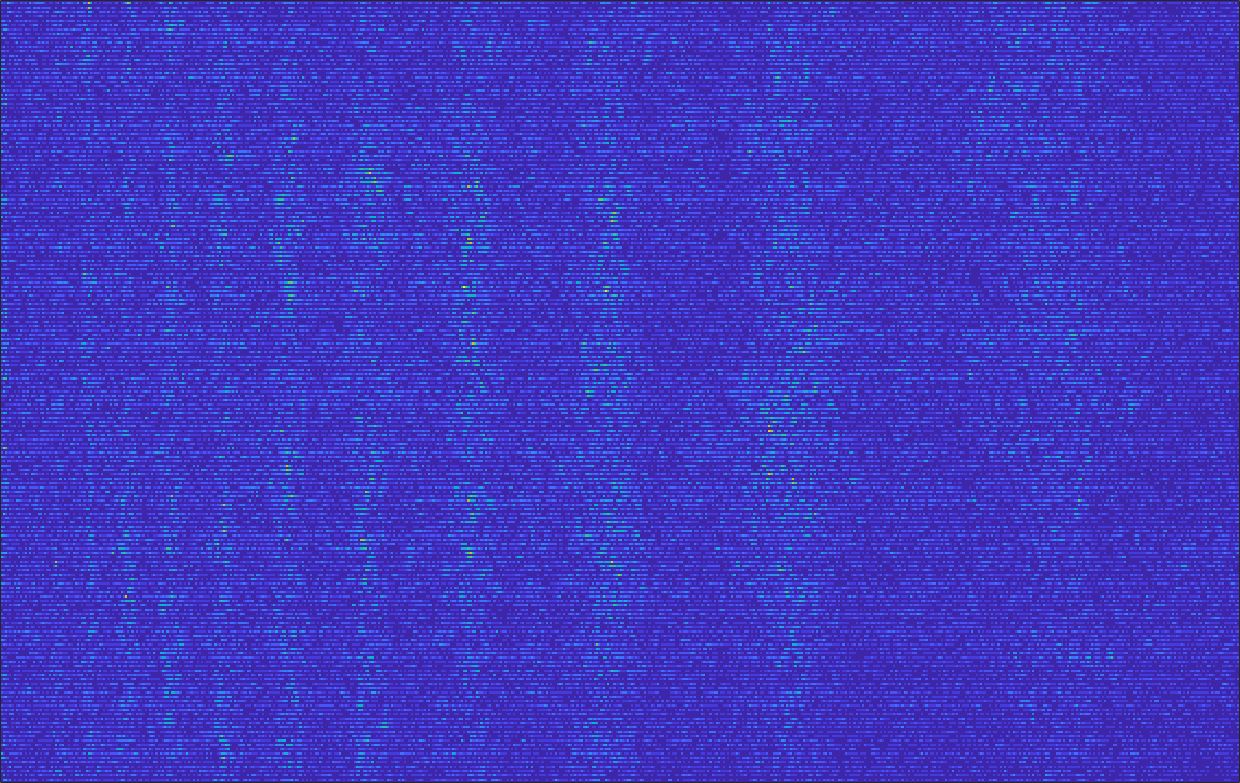} &
\includegraphics[width=0.45\textwidth]{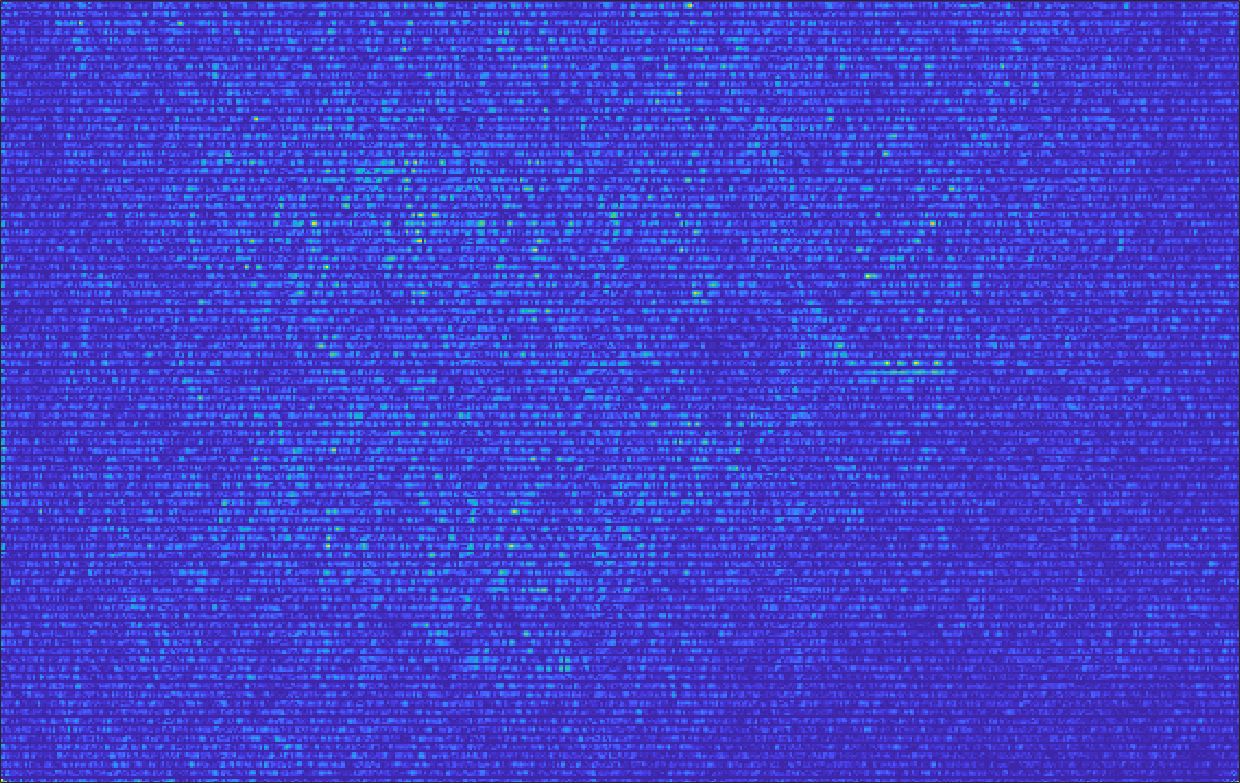} \\
Max. error: 13 & Max. error: 30 \\
\end{tabular}
\caption{With reference to Figs. (\ref{FIG:OBNETWORK},~\ref{FIG:CARDIONETWORK},~\ref{FIG:ABDONETWORK}), we show the error image of our method with respect to the target image with both 2X and 4X up-sampling factors: obstetric district (first row), cardiac district (second row), and abdominal district (third row). For each image, we report the maximum error in the scale~$0-255$.\label{FIG:ERRORIMAGE}}
\end{figure*}
\paragraph{Multi-frame US super-resolution}
The \emph{Bilinear Deformable Block Matching}~\cite{basarab2008method}, which is a registration method that accounts for the complex and deformable motion of soft tissues, is applied to reconstruct the HR image by exploiting the shifting property of the Fourier transform and the aliasing relationship between the continuous Fourier transform of the HR image and the discrete Fourier transform of LR images~\cite{morin2012post}. The use of deep learning for motion estimation among different frames~\cite{abdel2016ultrasound} reduces the effect of noise and artefacts and reconstructs HR images from a sequence of LR images. The modelling of the spatial correlation of the speckle noise~\cite{cuneyitouglu2019single} is applied to standard reconstruction methods, with tissue-mimicking phantom and co-registered multi-images.
\begin{figure*}[t]
\centering
\begin{tabular}{cc|cc|cc}
\multicolumn{2}{c|}{(a) Obstetric district}
&\multicolumn{2}{c|}{(b) Cardiac district}
&\multicolumn{2}{c}{(c) Abdominal district}\\
\hline
\includegraphics[width=0.28\columnwidth]{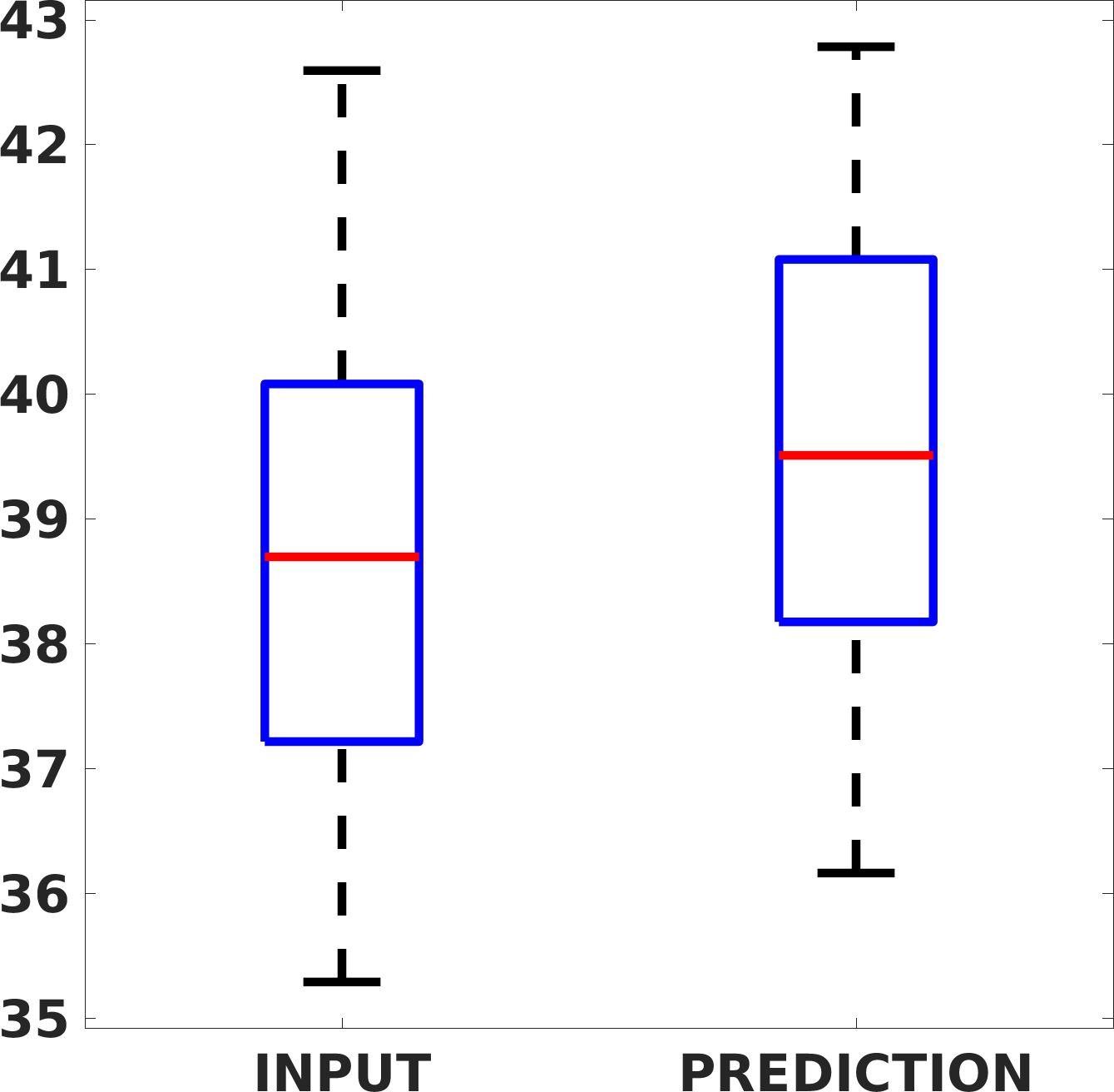} &
\includegraphics[width=0.28\columnwidth]{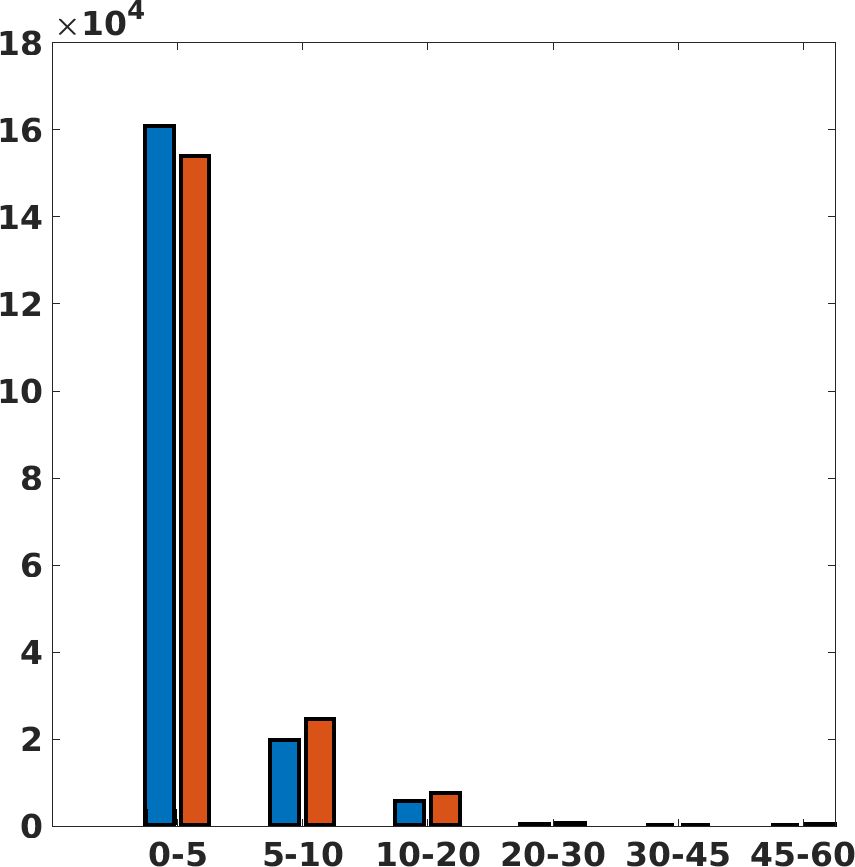} &
\includegraphics[width=0.28\columnwidth]{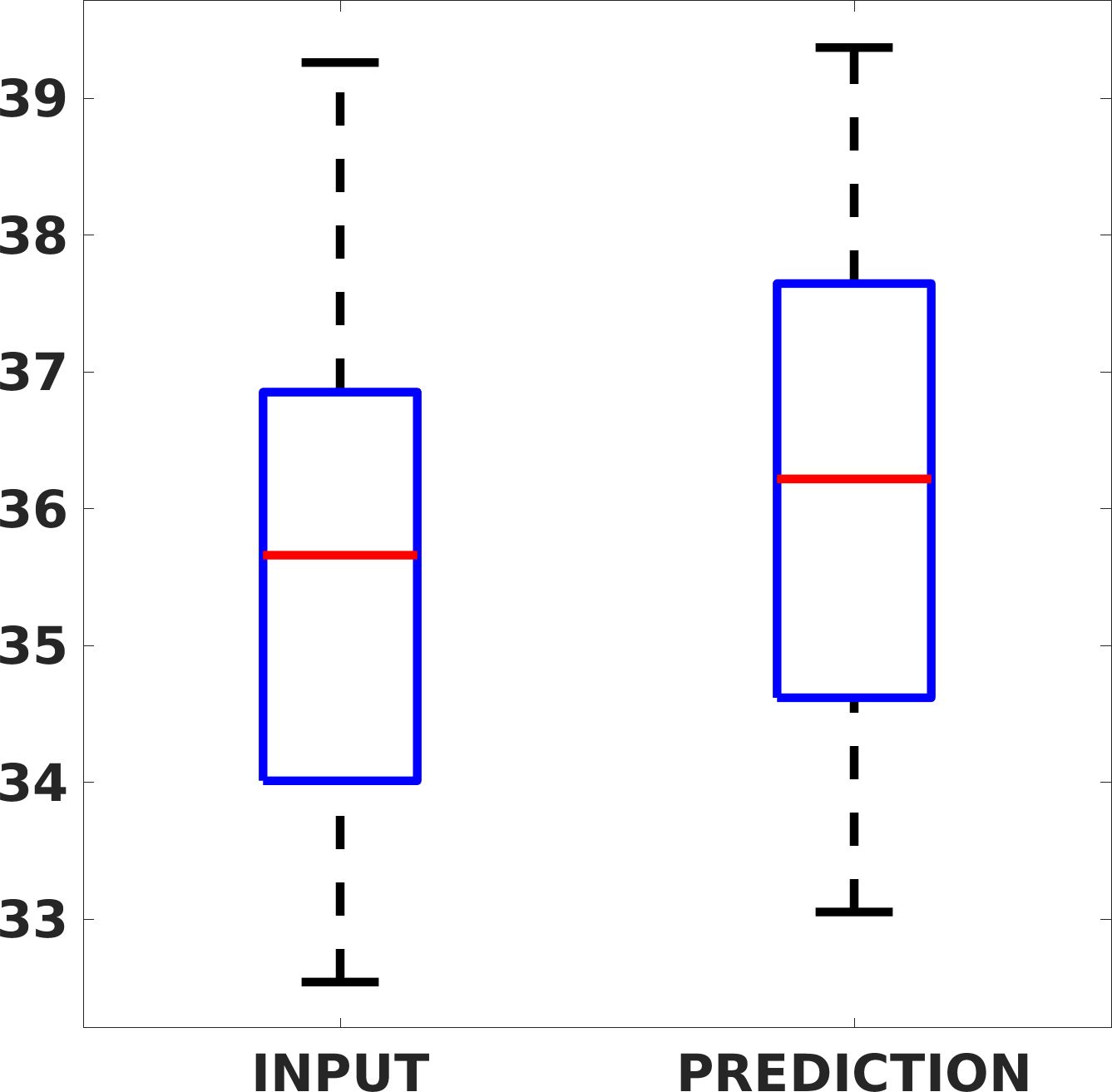} &
\includegraphics[width=0.28\columnwidth]{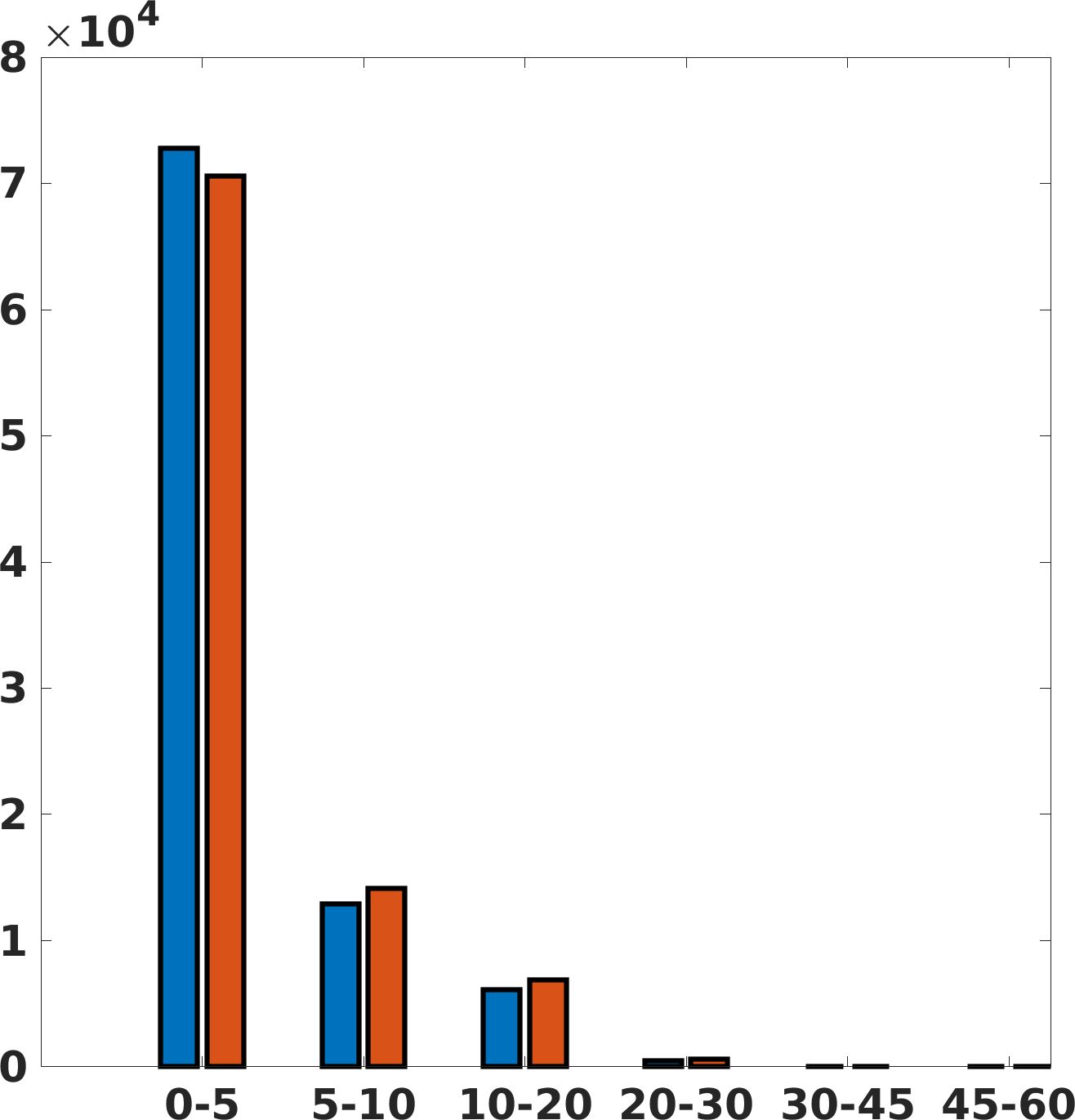} &
\includegraphics[width=0.28\columnwidth]{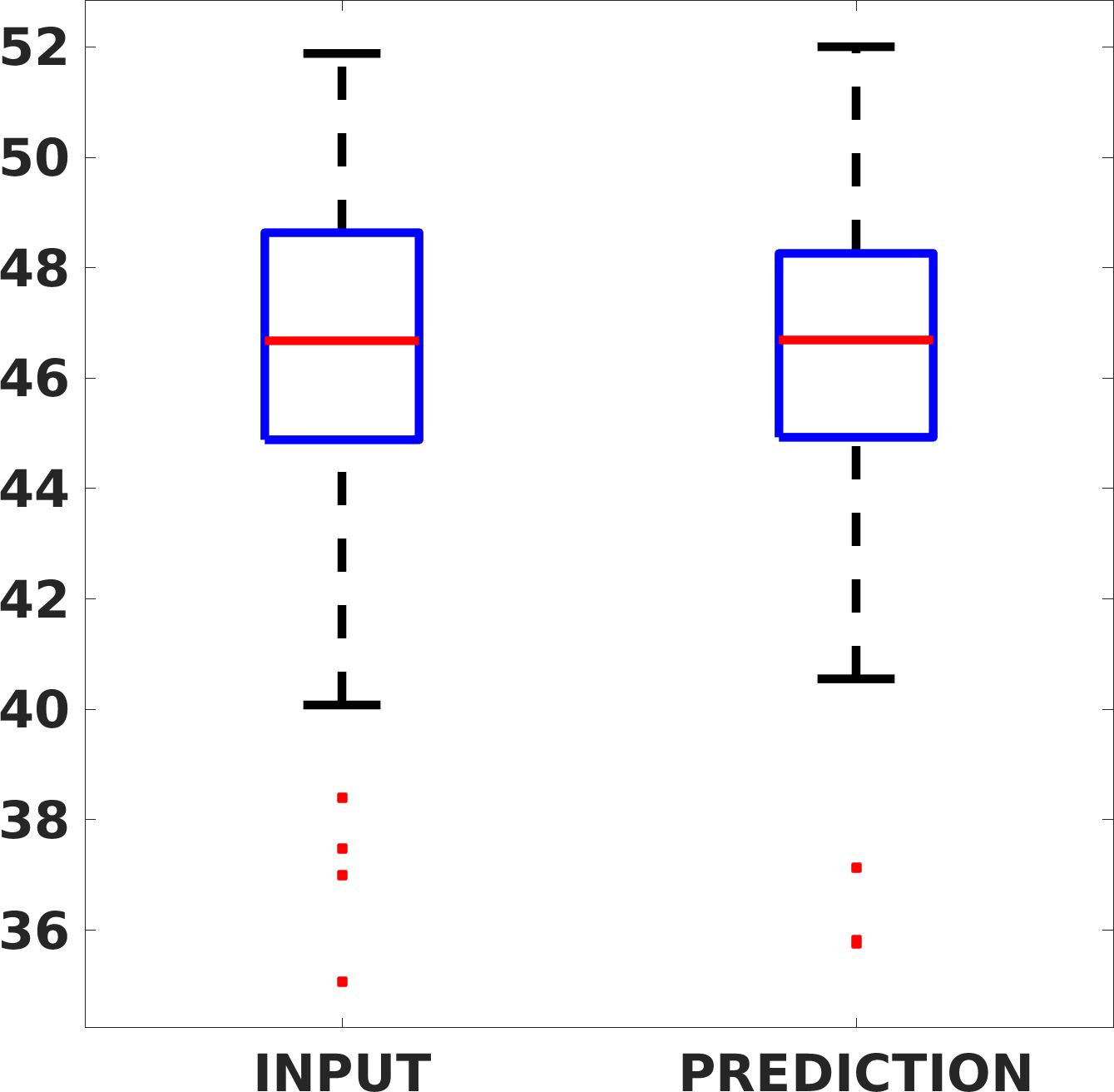} &
\includegraphics[width=0.28\columnwidth]{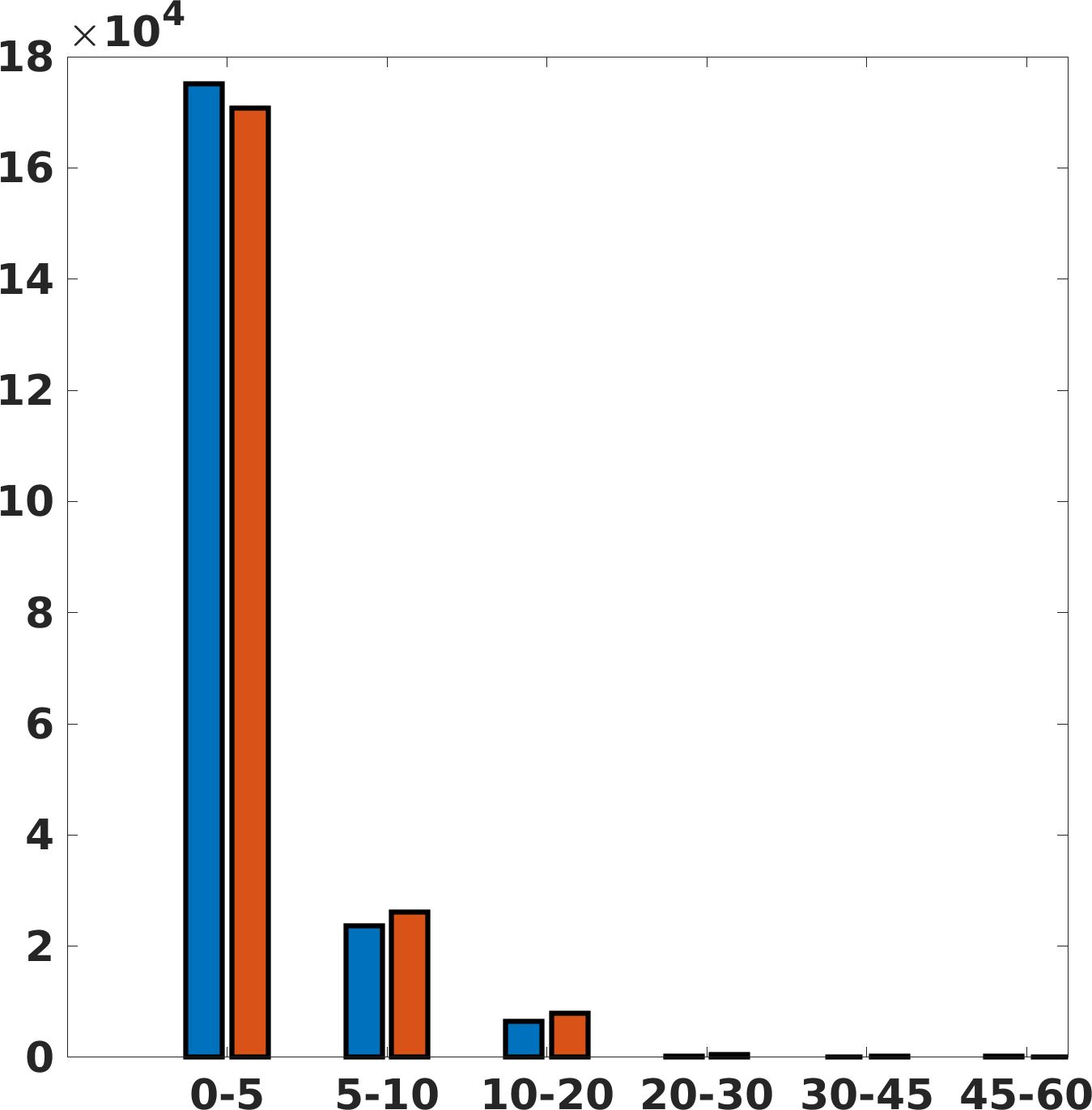} \\ 
\includegraphics[width=0.28\columnwidth]{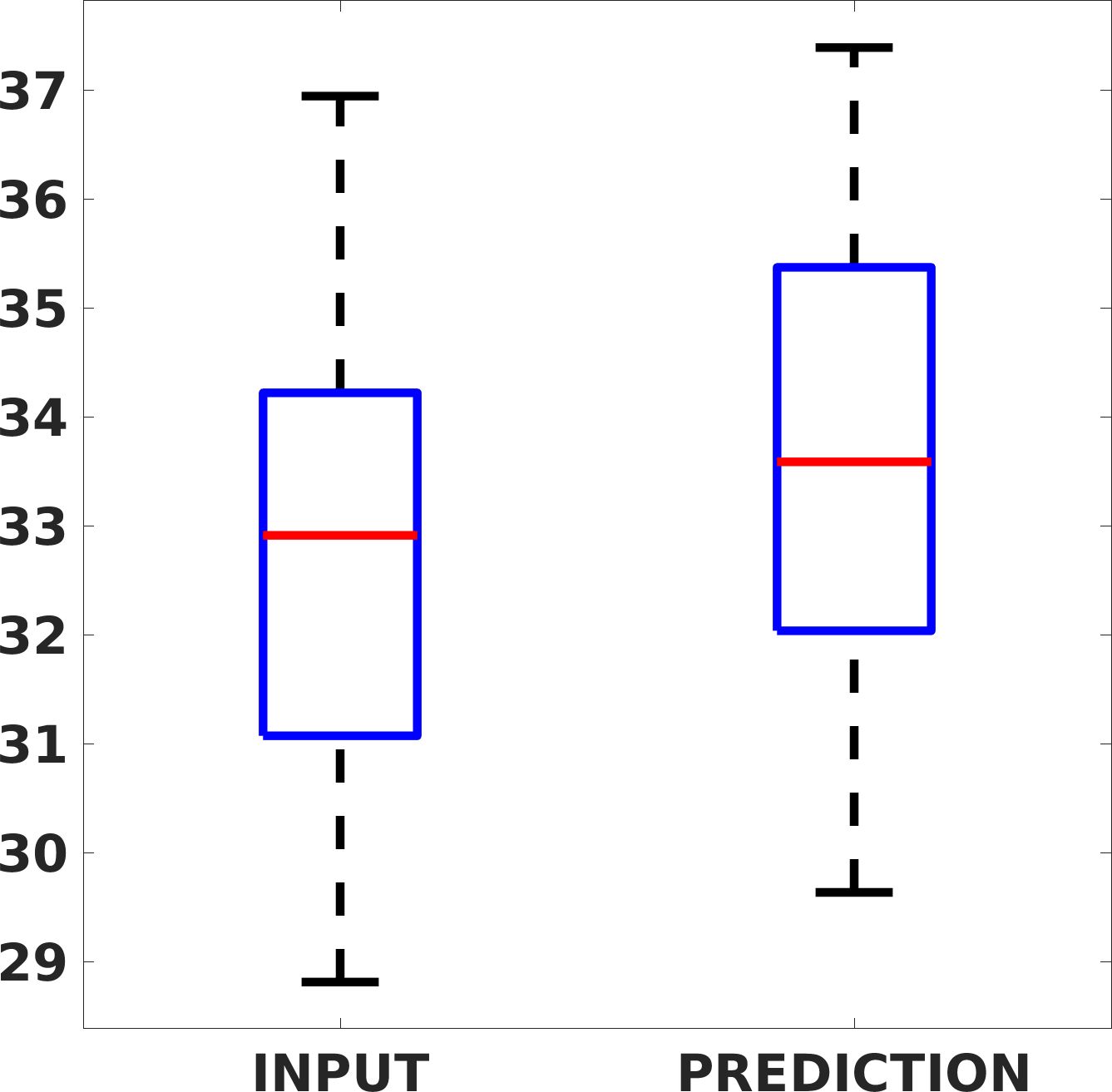} &
\includegraphics[width=0.28\columnwidth]{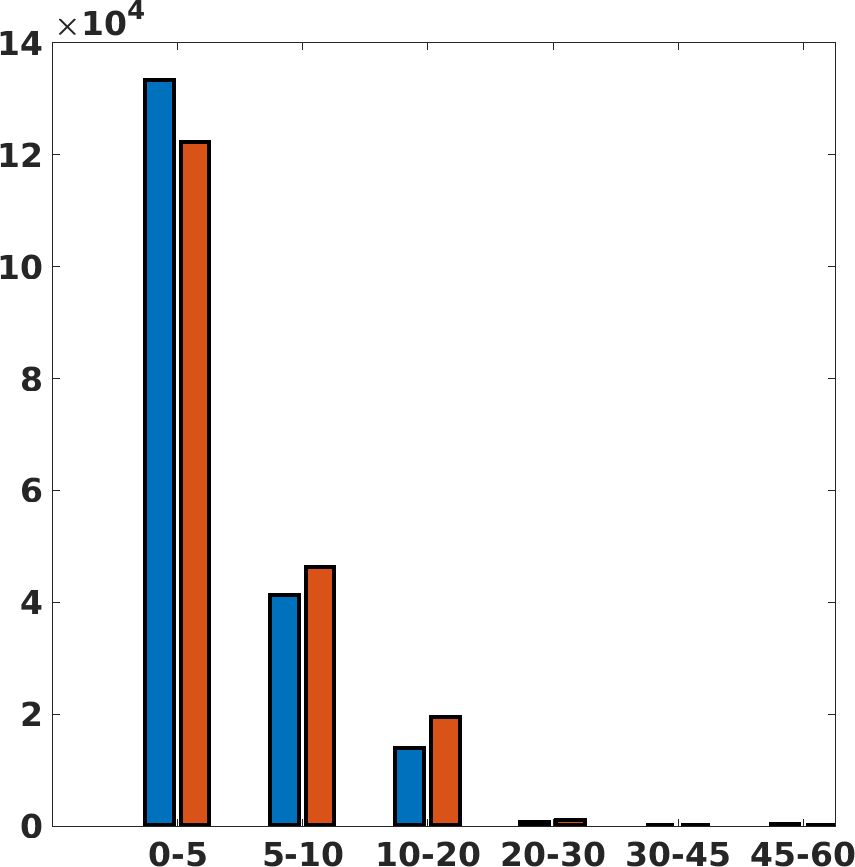} &
\includegraphics[width=0.28\columnwidth]{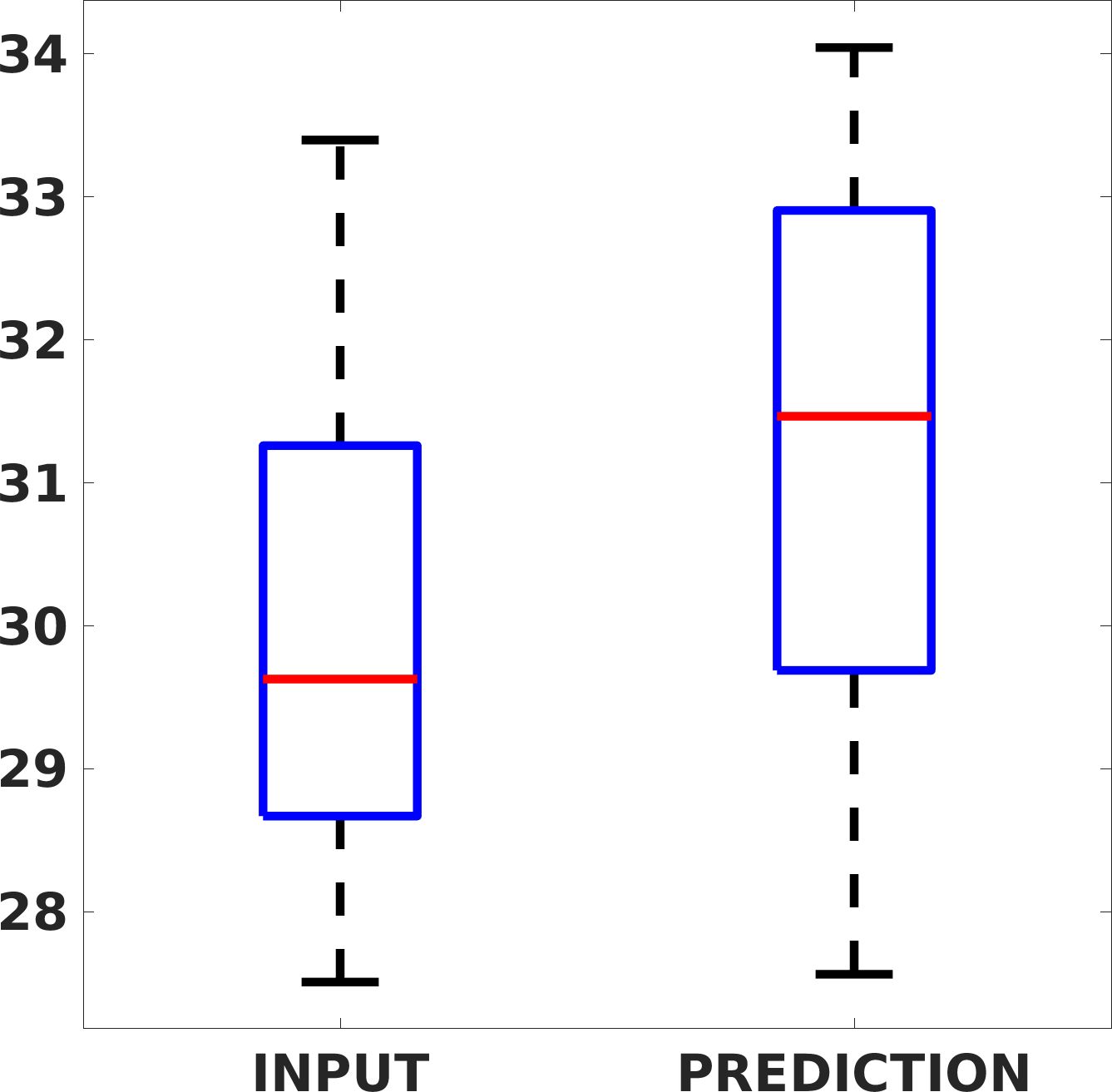} &
\includegraphics[width=0.28\columnwidth]{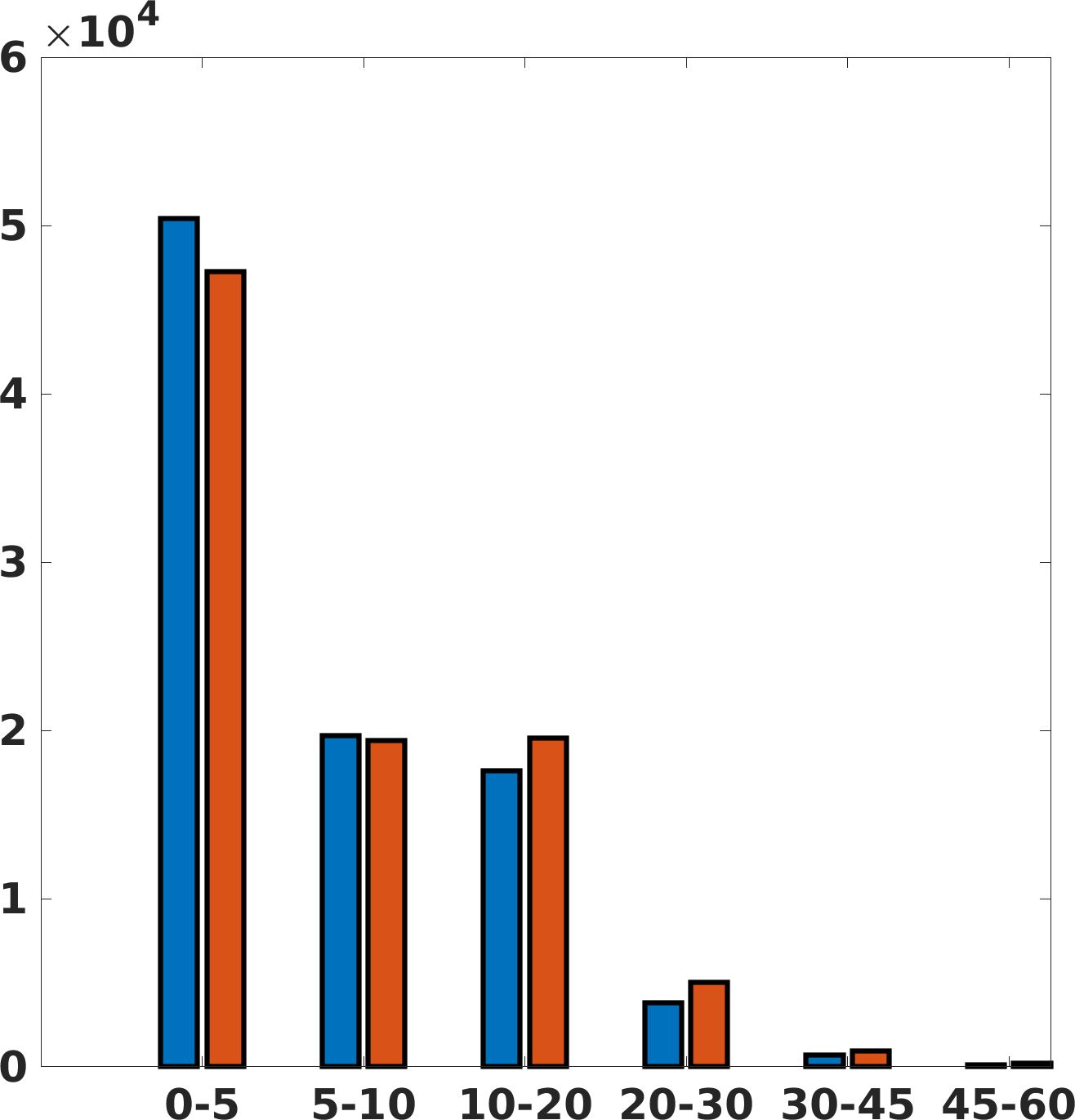} &
\includegraphics[width=0.28\columnwidth]{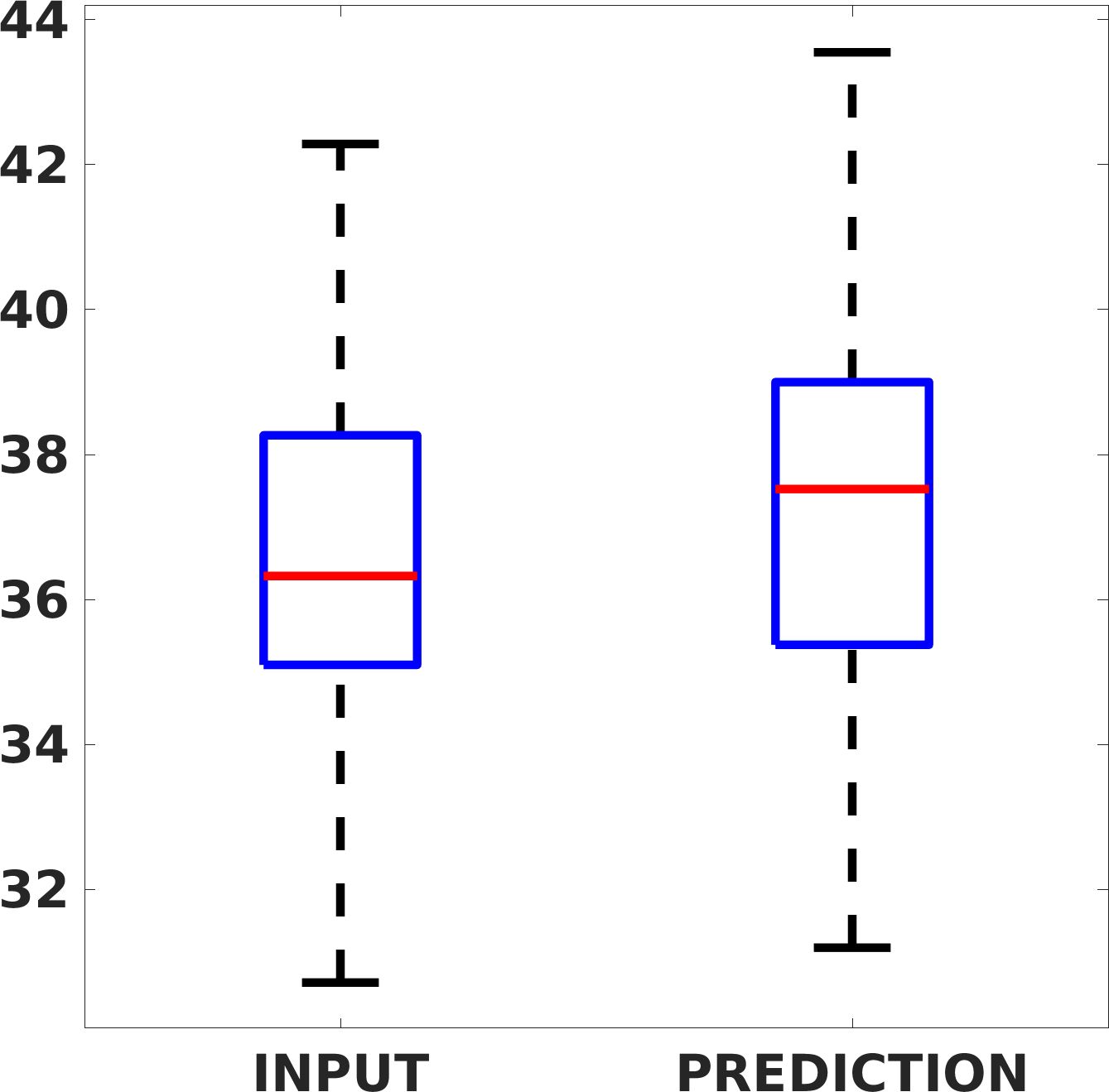} &
\includegraphics[width=0.28\columnwidth]{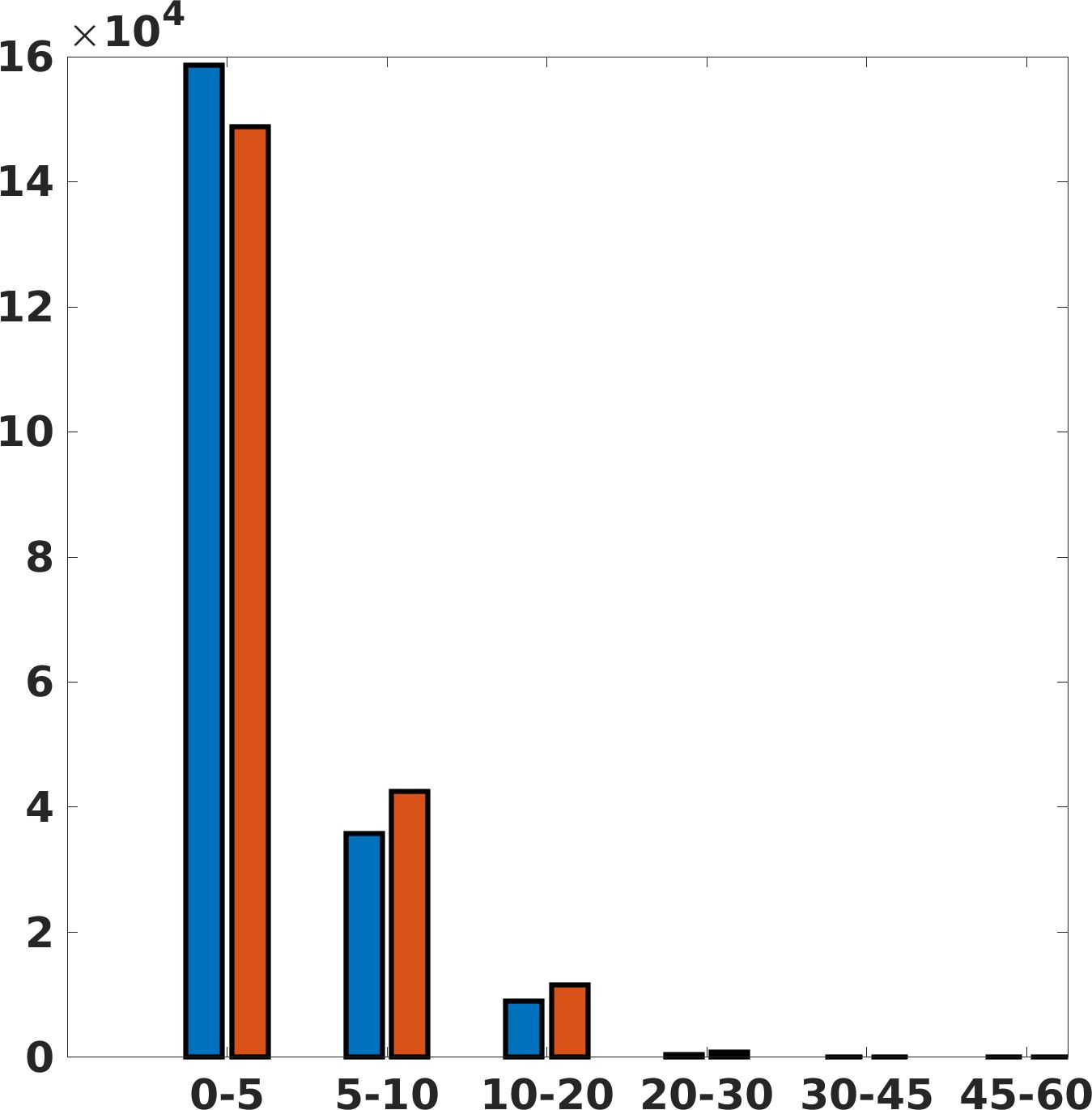} 
\end{tabular}
\caption{PSNR box-plot (left) of the (a) obstetric, (b) cardiac, and (c) abdominal districts, and error histogram (right): prediction (blue) vs. input (red): 2X (first line) and 4X (second line) results. The box-plot represents the statistic of the PSNR on the 200 images test data set; the improvement of the network prediction with respect to the up-sampled image ranges from lower than~$1\%$  (abdominal district, 2X) to~$6.1\%$ (cardiac district, 4X). \label{FIG:OBQUANTITATIVE}}
\end{figure*}
\section{Proposed super-resolution of US signals\label{SEC:METHOD}}
The ultrasonic waves are emitted by the probe and straightly penetrate the body structures along their path; when they pass through adjacent parts of the body with a different acoustic impedance, a fraction of the ultrasound pulse returns as a reflected wave, generating an echo that returns to the probe, while the rest of the wave continues to penetrate along the beam to greater tissue depths. The amplitude of the reflected echo is proportional to the difference in acoustic impedance between two adjacent media. For example, interfaces between soft tissue and dense organs (e.g., bones) generate very strong echoes due to a large acoustic impedance gradient. The acoustic impedance is a physical property of a medium defined as the density of the medium times the velocity of the wave propagation. Human body tissues have different acoustic impedances: for example, air-containing organs (such as the lung) have the lowest acoustic impedance, while dense organs have a higher acoustic impedance.
\begin{figure*}[t]
\centering
\begin{tabular}{cc|cc|cc}
\multicolumn{2}{c|}{(a) Obstetric district}
&\multicolumn{2}{c|}{(b) Cardiac district}
&\multicolumn{2}{c}{(c) Abdominal district}\\
\hline
\includegraphics[width=0.28\columnwidth]{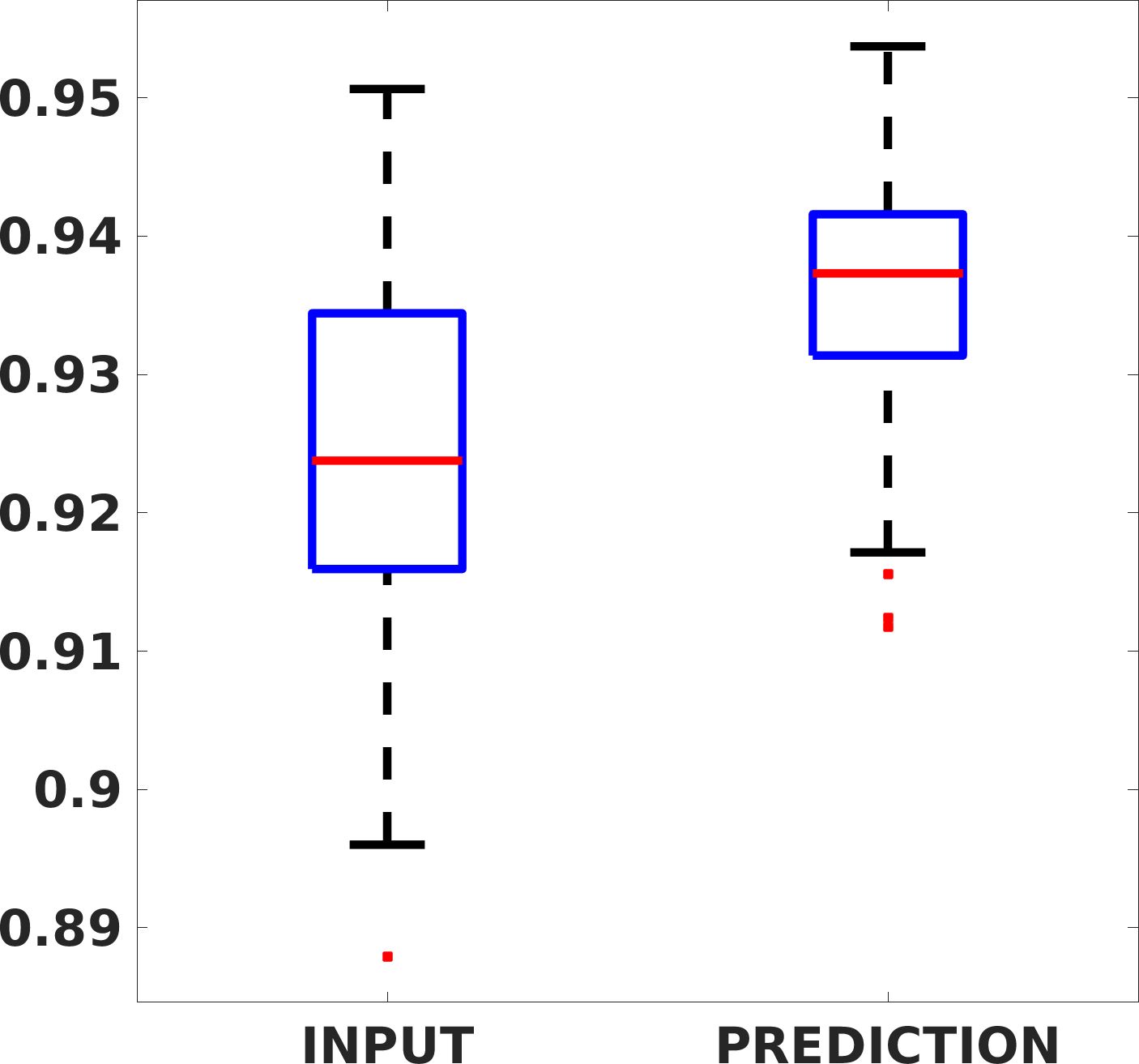} &
\includegraphics[width=0.28\columnwidth]{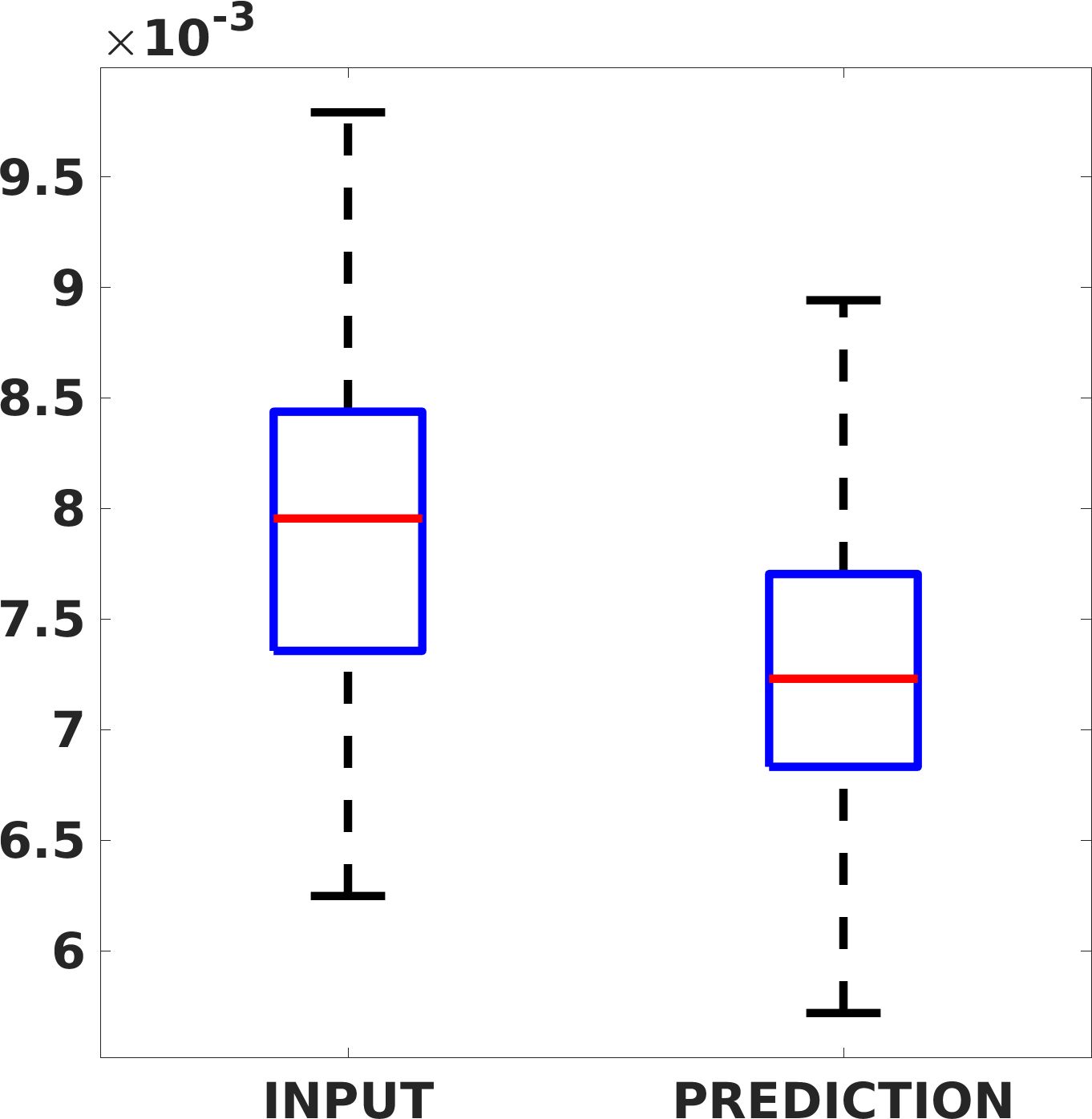} &
\includegraphics[width=0.28\columnwidth]{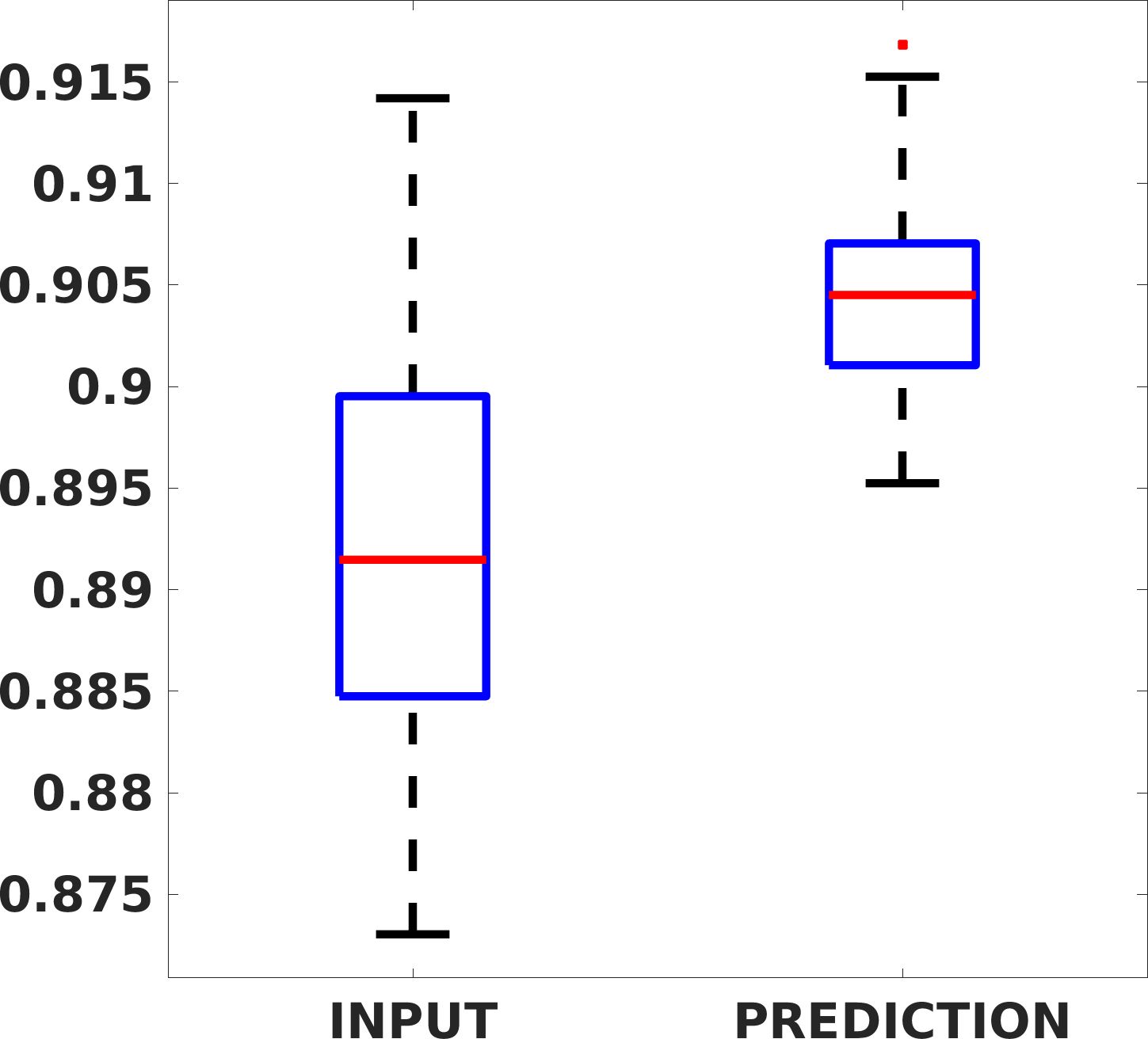} &
\includegraphics[width=0.28\columnwidth]{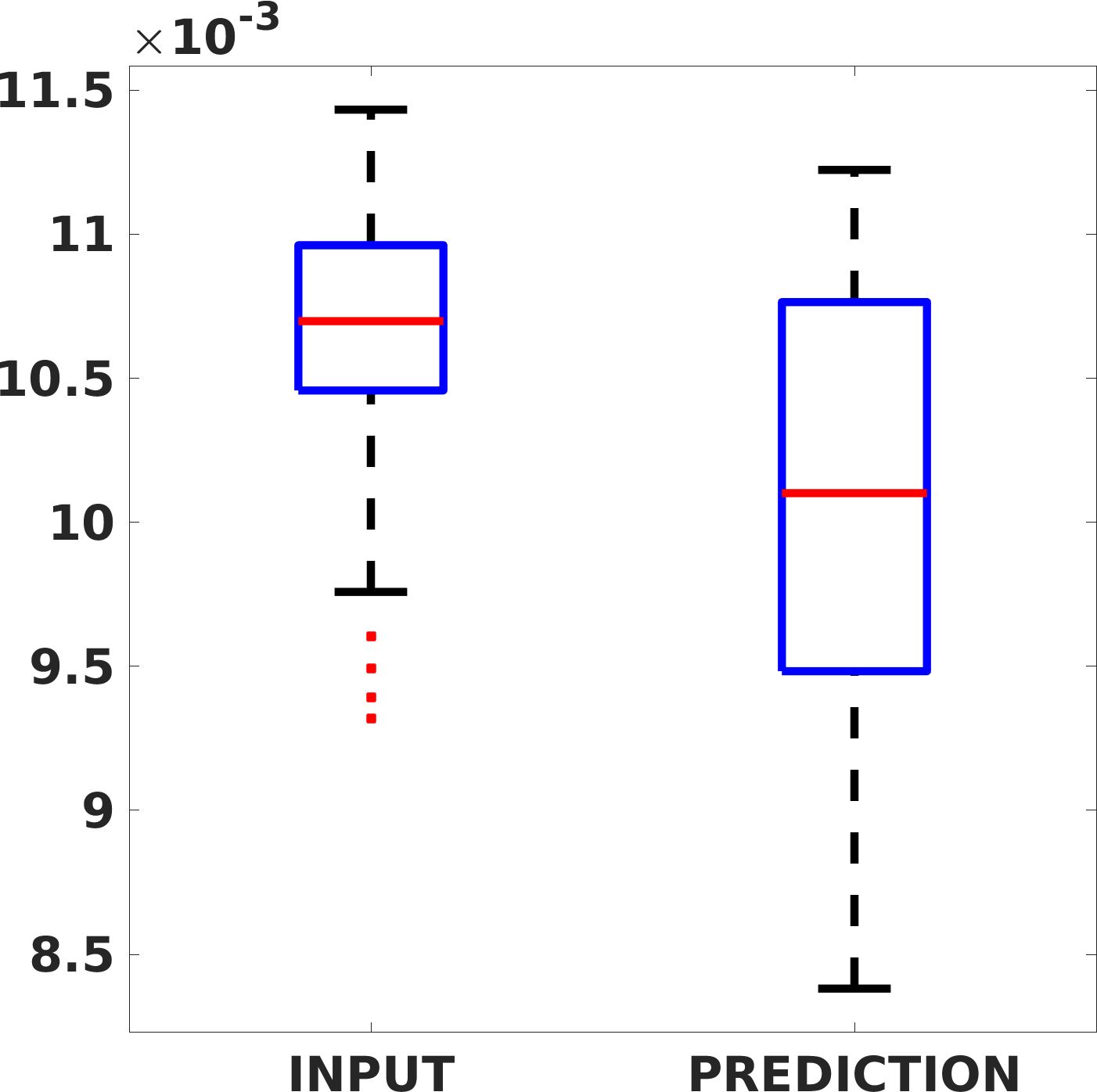} &
\includegraphics[width=0.28\columnwidth]{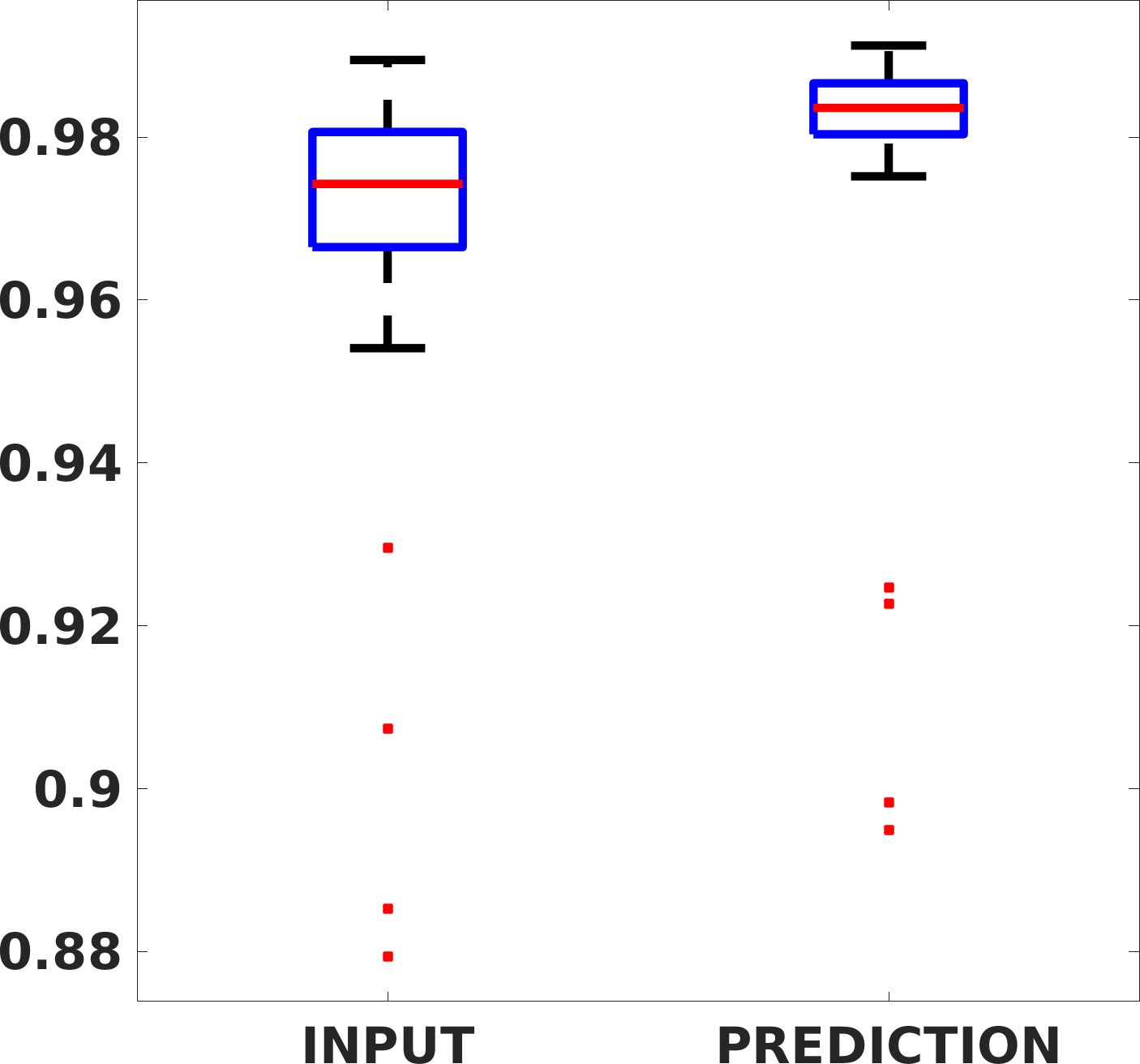} &
\includegraphics[width=0.28\columnwidth]{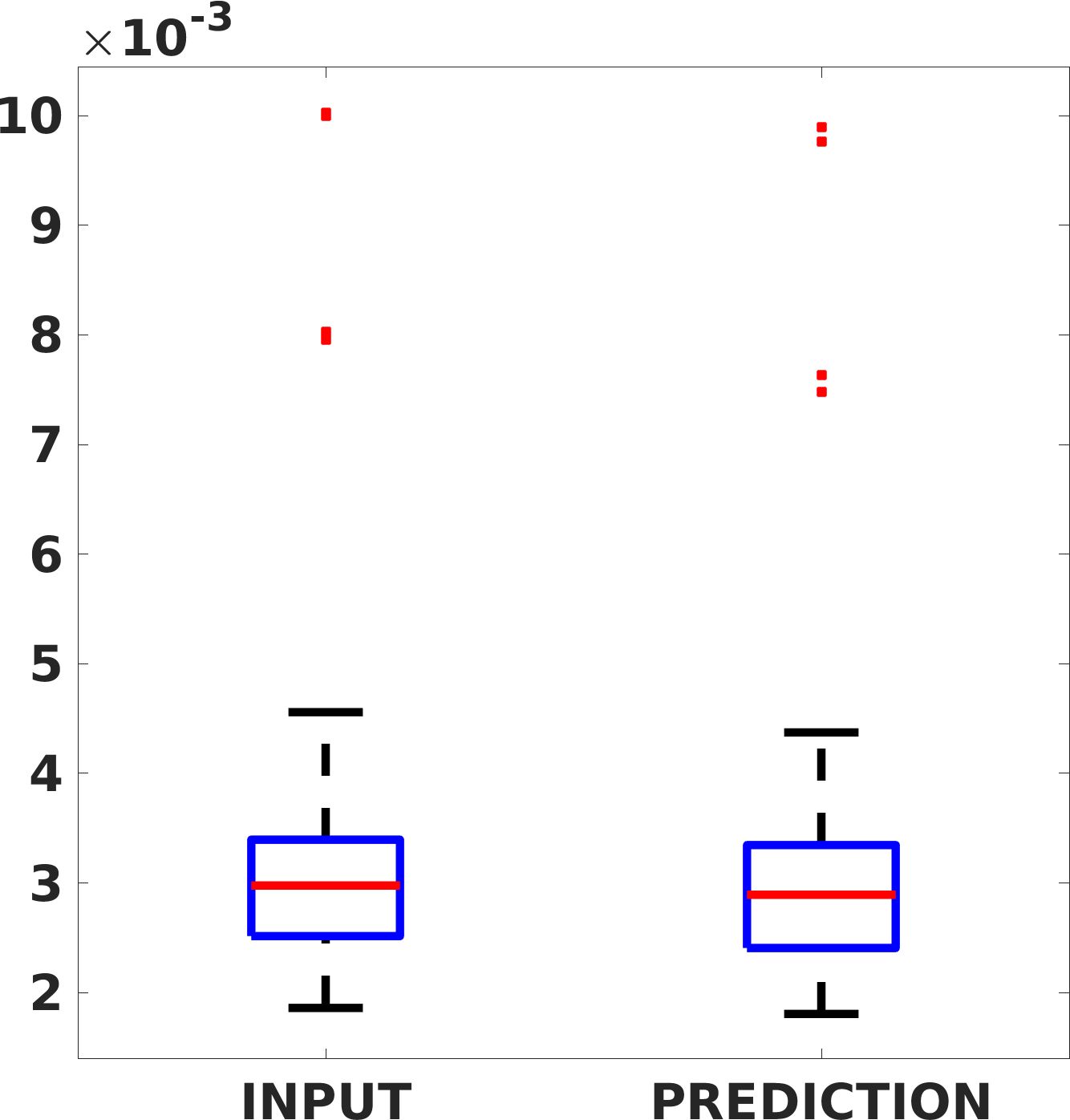} \\ 
\includegraphics[width=0.28\columnwidth]{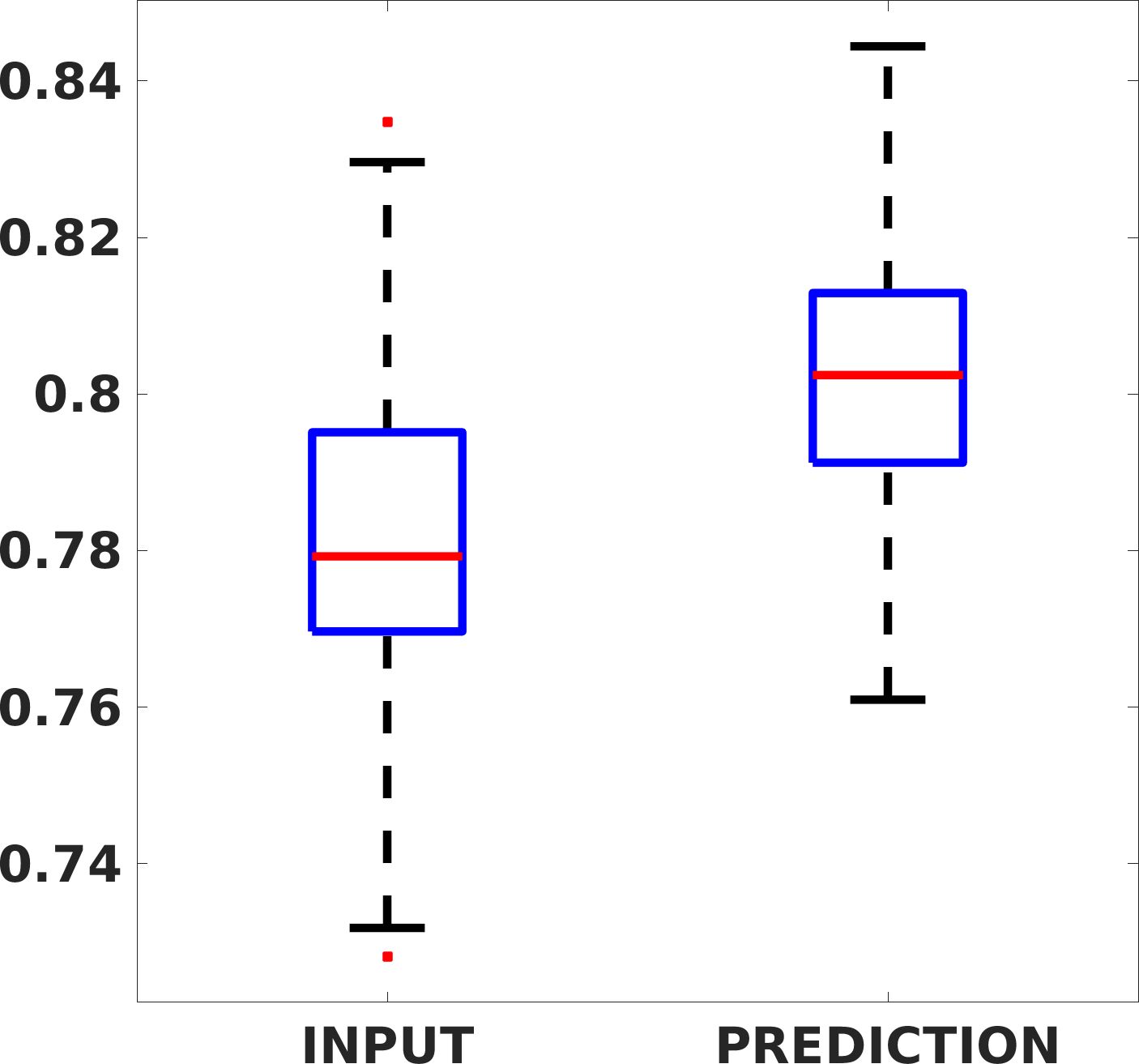} &
\includegraphics[width=0.28\columnwidth]{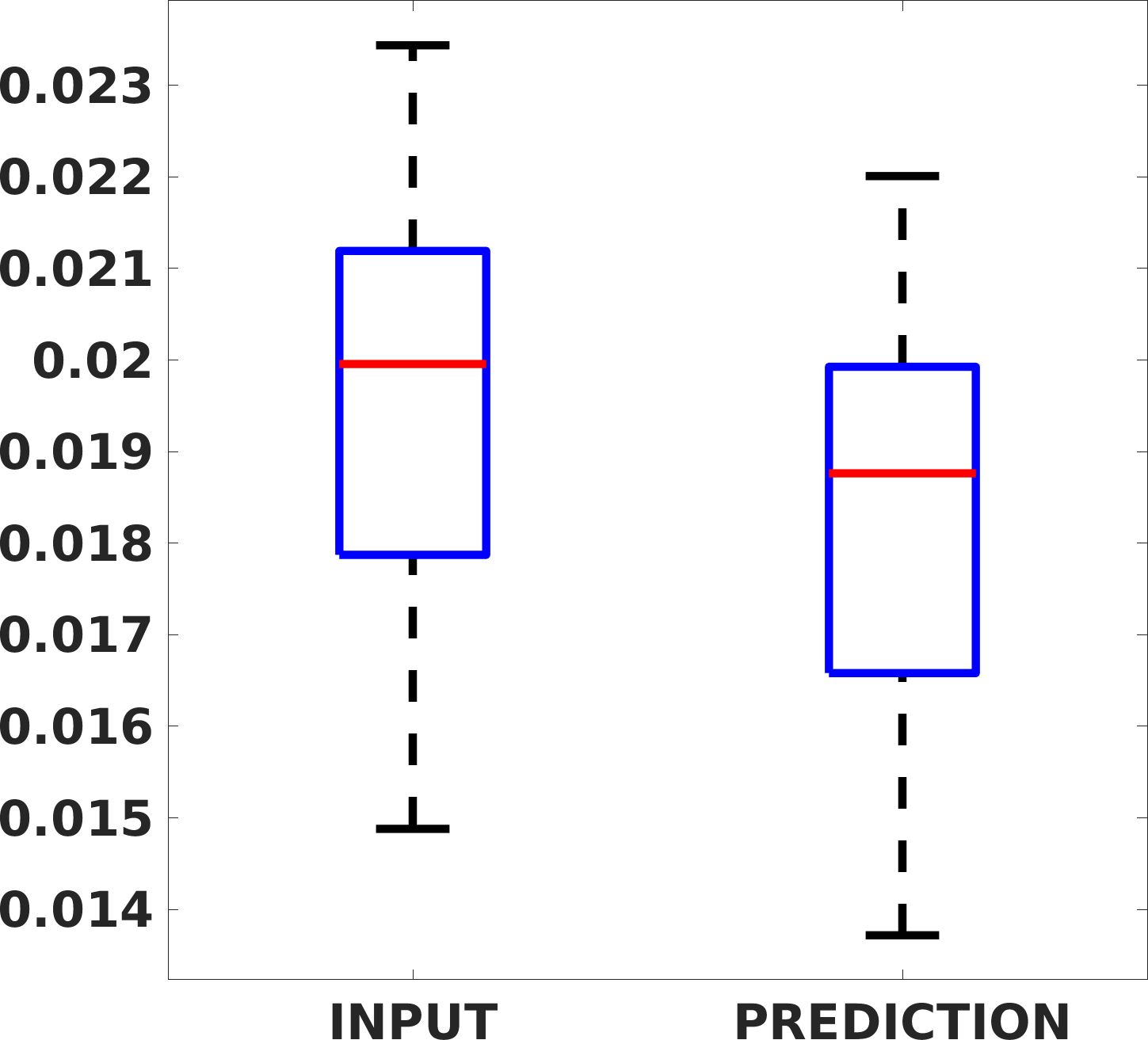} &
\includegraphics[width=0.28\columnwidth]{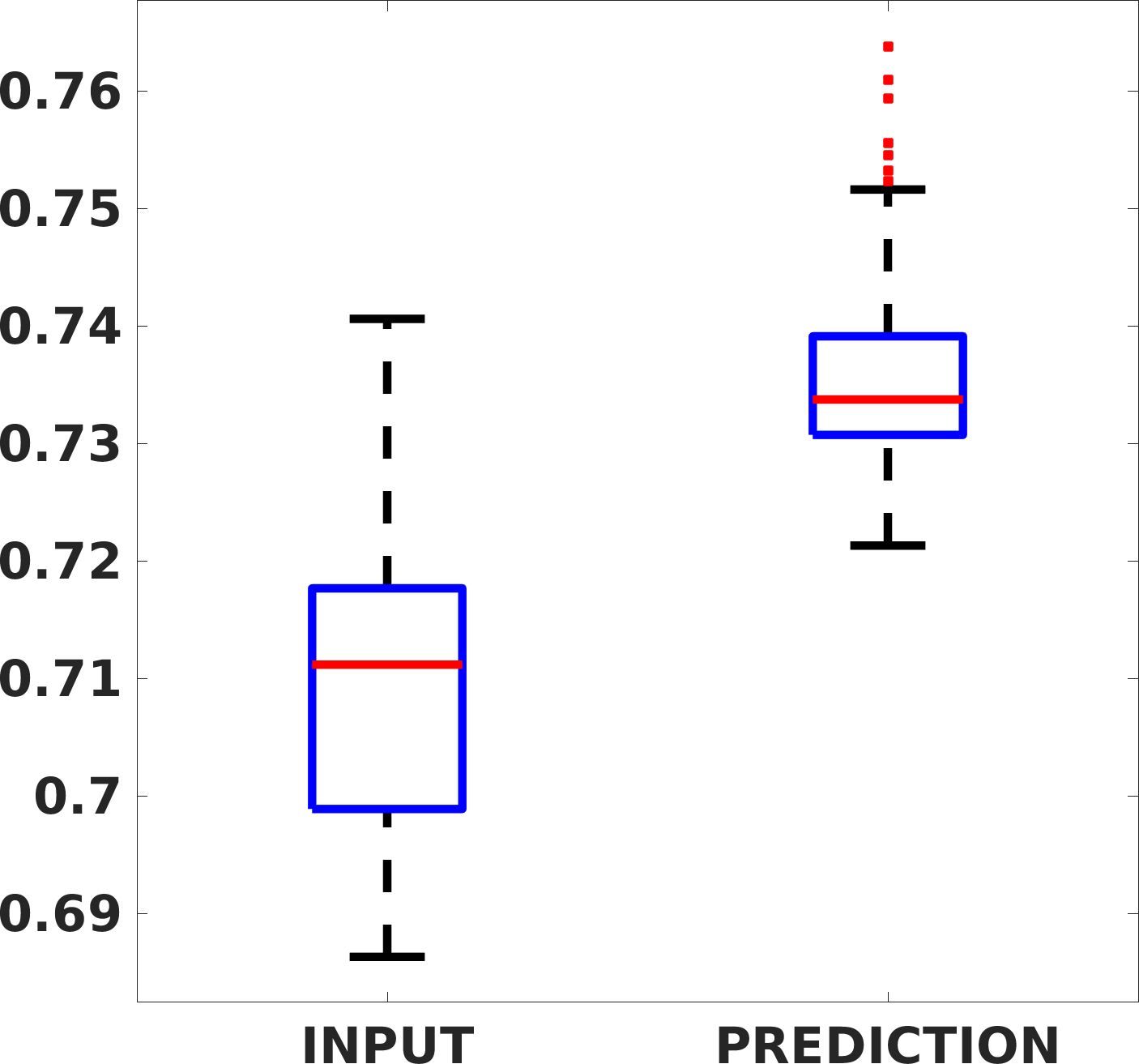} &
\includegraphics[width=0.28\columnwidth]{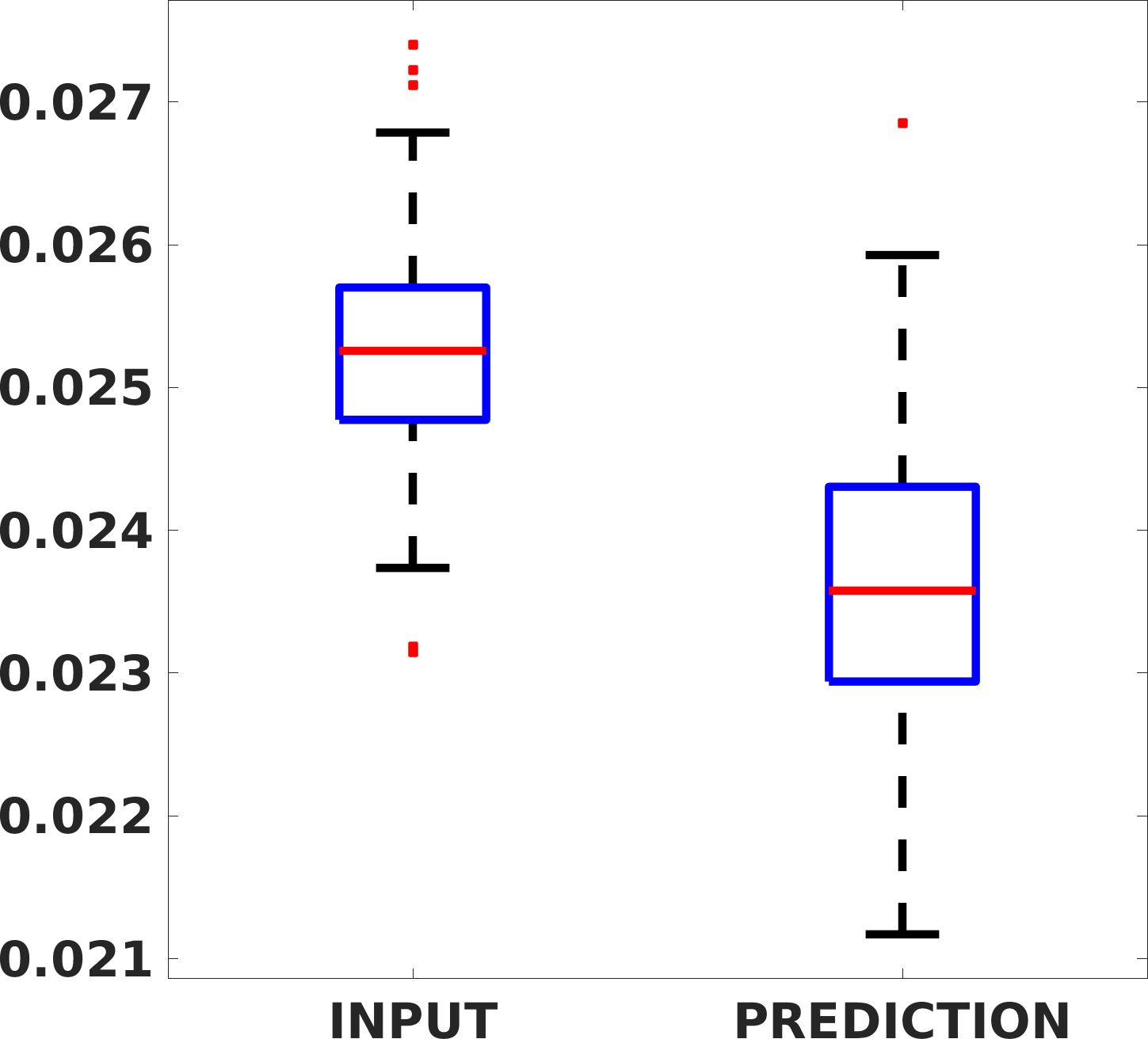} &
\includegraphics[width=0.28\columnwidth]{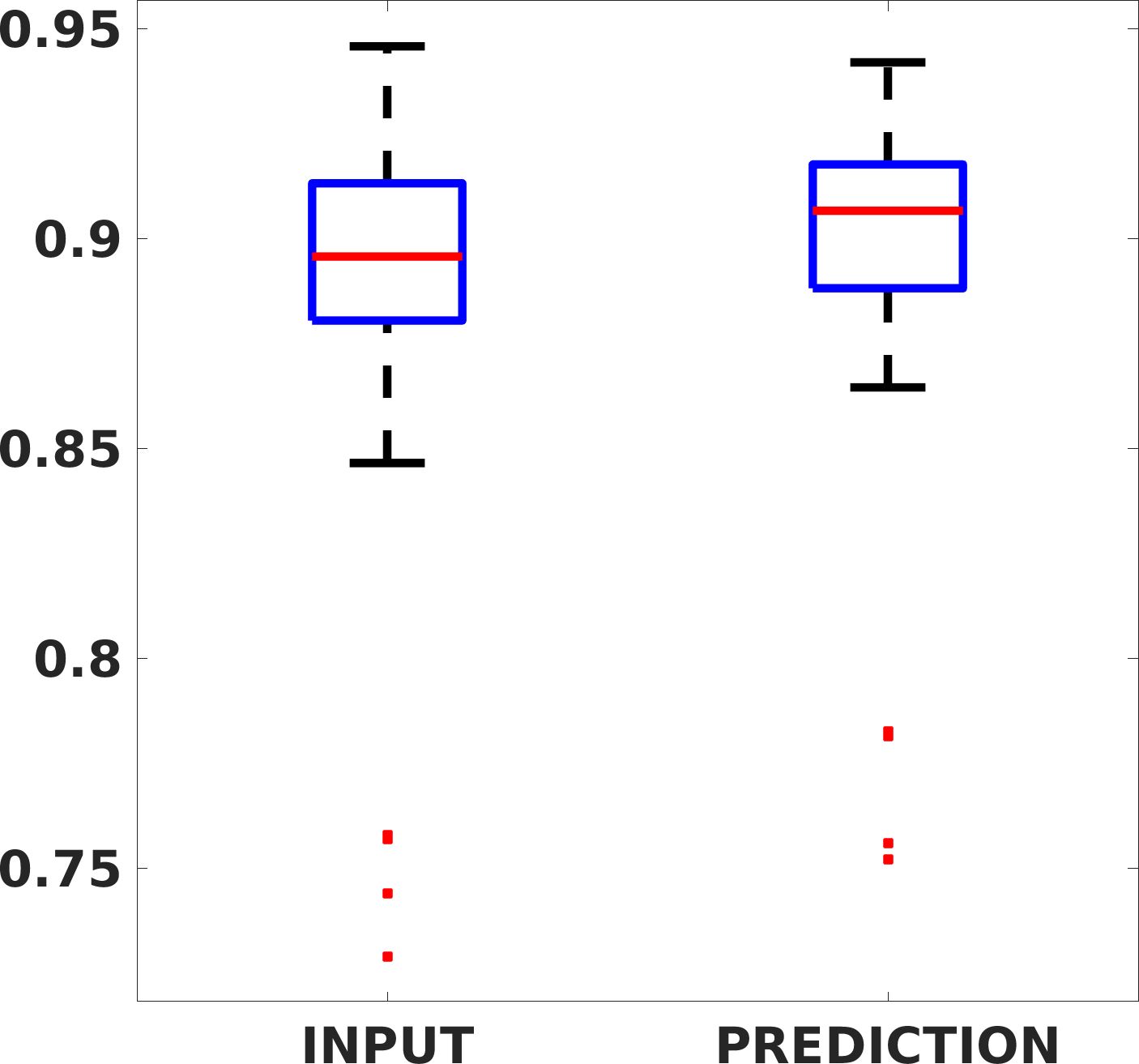} &
\includegraphics[width=0.28\columnwidth]{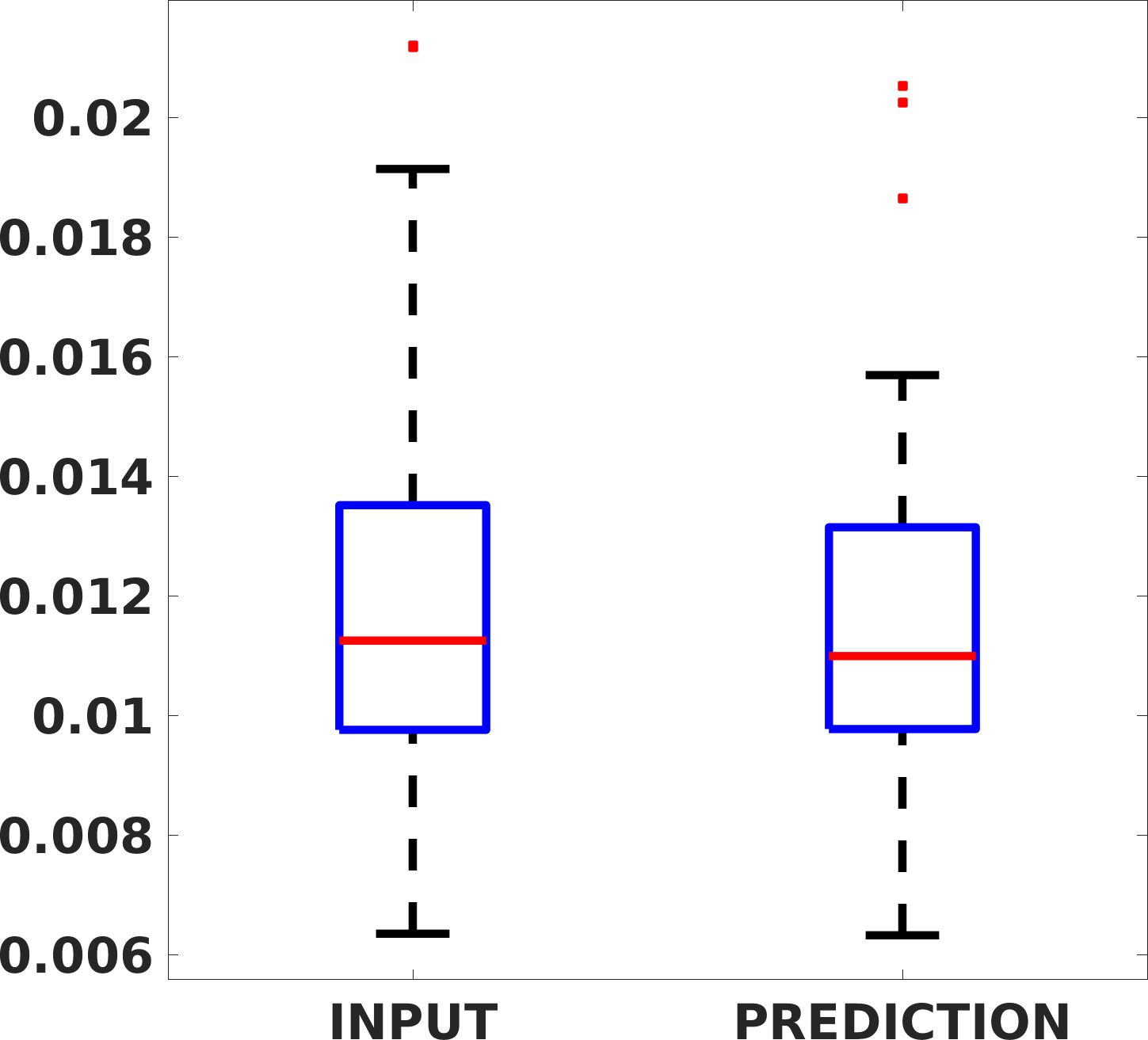} 
\end{tabular}
\caption{SSIM box-plot (left) and MAE box-plot (right) of the (a) obstetric, (b) cardiac, and (c) abdominal districts:  2X (first line) and 4X (second line) results. The median value of the SSIM has a maximum improvement of~$3\%$ (cardiac, 4X), while the MAE has a maximum improvement of~$6.5\%$ (obstetric, 2X). \label{FIG:OBQUANTITATIVE2}}
\end{figure*}

The echo signals are processed and combined to generate the underlying image, which has a resolution of~$l \times d$, where~$l$ is the number of beamlines (i.e., the lateral resolution), and~$d$ is the depth of the acquisition of each beam line (i.e., the axial resolution). Axial resolution refers to the ability to discern two separate objects that are longitudinally adjacent to each other in the ultrasound image; lateral resolution refers to the ability to discern two separate objects that are adjacent to each other; the lateral resolution is usually lower than the axial resolution in ultrasound.  The resolution of the image in terms of lateral direction (i.e., the direction perpendicular to the US propagation along the beam line) is primarily determined by the width of the ultrasound beam and the number of elements (i.e., the piezoelectric crystals) that are activated to generate the US waves. Current probes vary the number of beamlines acquired by activating/deactivating piezoelectric crystals thus reducing lateral resolution and image acquisition time.  The axial resolution can be varied by changing the length and frequency of the pulses, which affect the penetration of the ultrasound wave.

In this context, we focus on lateral low-resolution image acquisition, reduce the acquisition time, and subsequently reconstruct the high-resolution image without losing information in terms of data depth. This aspect is relevant to improve the quality of the US image, its visual interpretation by the physician, and post-processing steps, e.g., as classification~\cite{abdel2017breast}, segmentation~\cite{brown2020deep}, and morphological analysis~\cite{schoen2021morphological}.
\begin{figure*}
\centering
\begin{tabular}{ccc|ccc|c}
\multicolumn{3}{c|}{2X Upsampling}
&\multicolumn{3}{c}{4X Upsampling}
&\\
\hline\\ [-2mm]
Obstetric &Cardiac &Abdominal &Obstetric &Cardiac &Abdominal &\\
\hspace*{-0.375cm}
\includegraphics[width=0.15\textwidth]{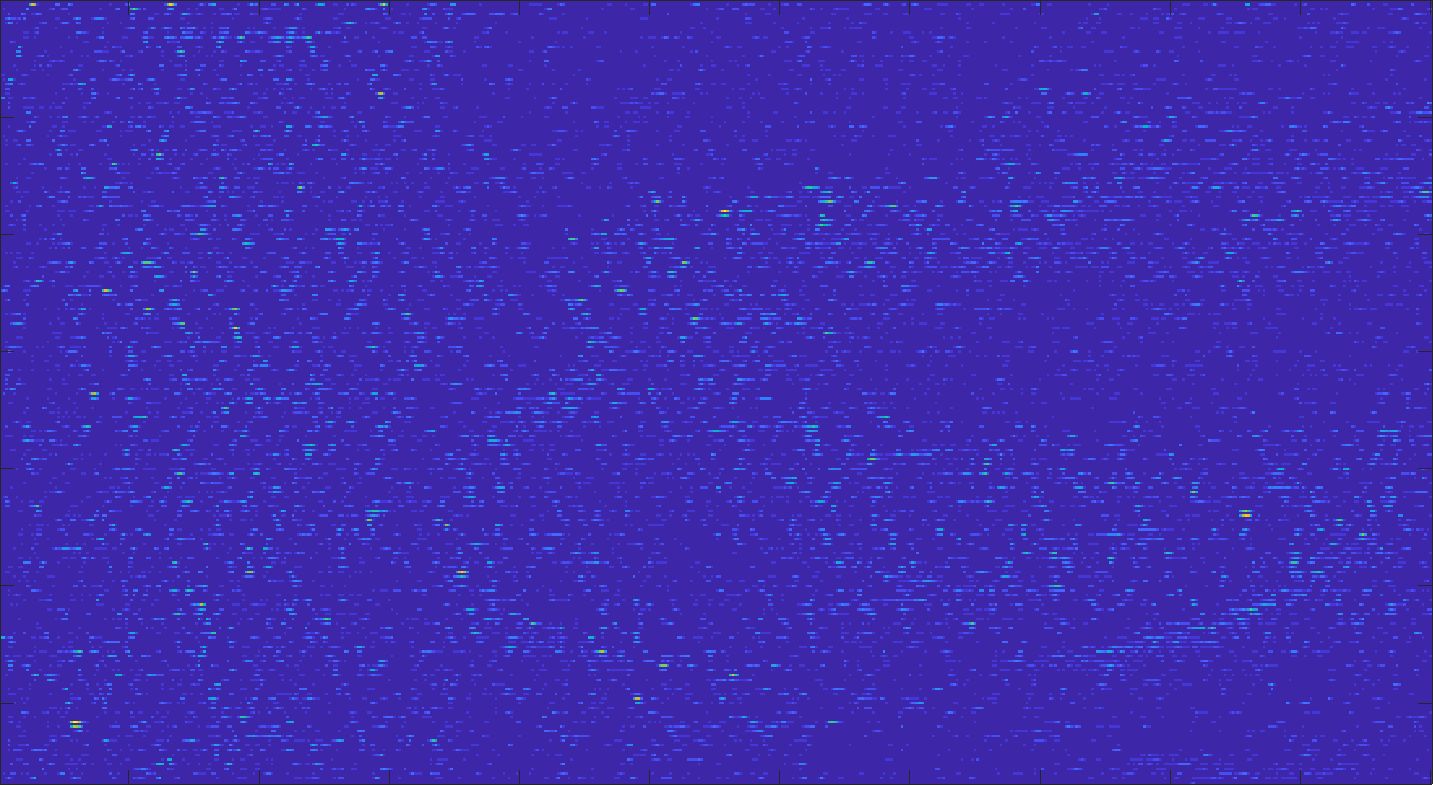} & \hspace*{-0.375cm}
\includegraphics[width=0.15\textwidth]{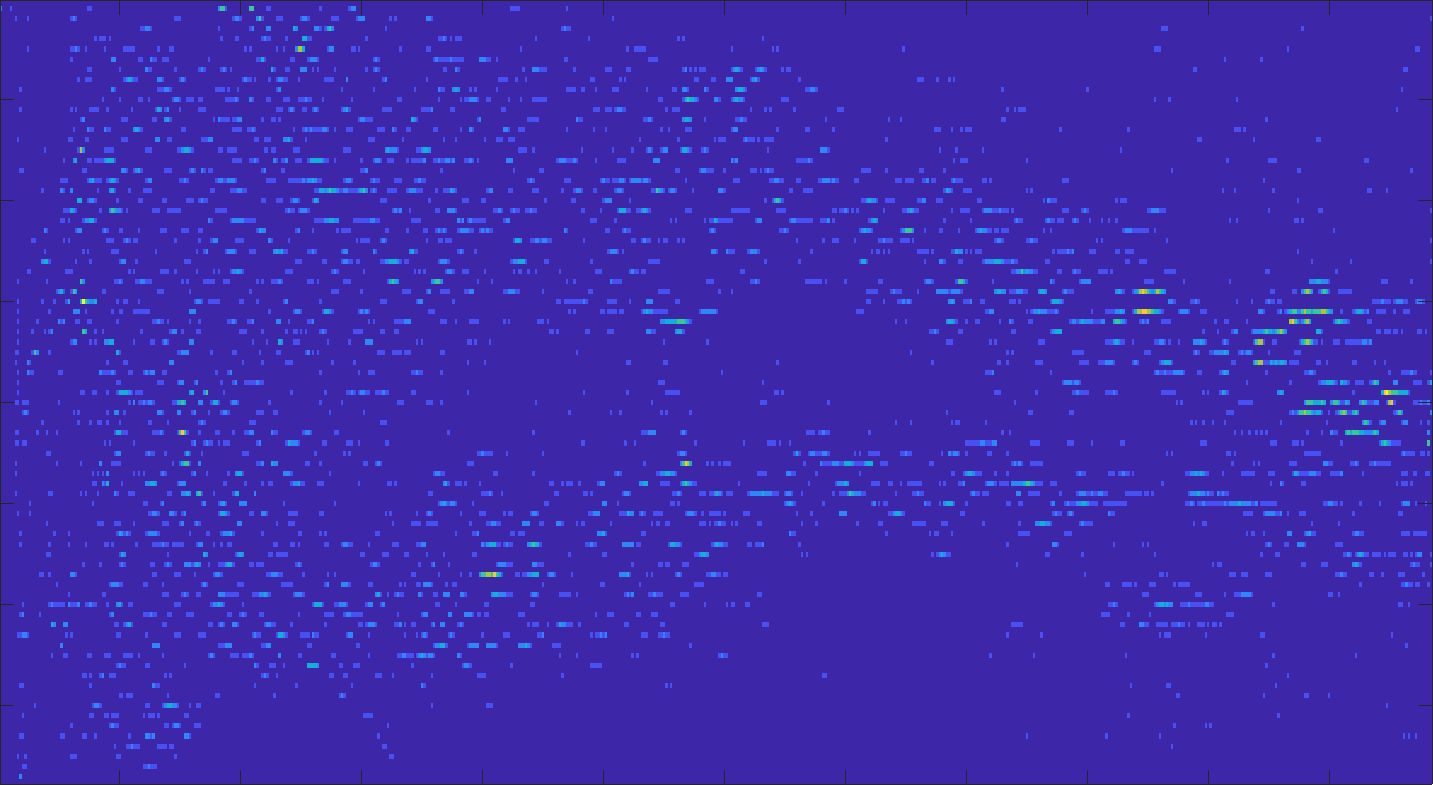} & \hspace*{-0.375cm}
\includegraphics[width=0.15\textwidth]{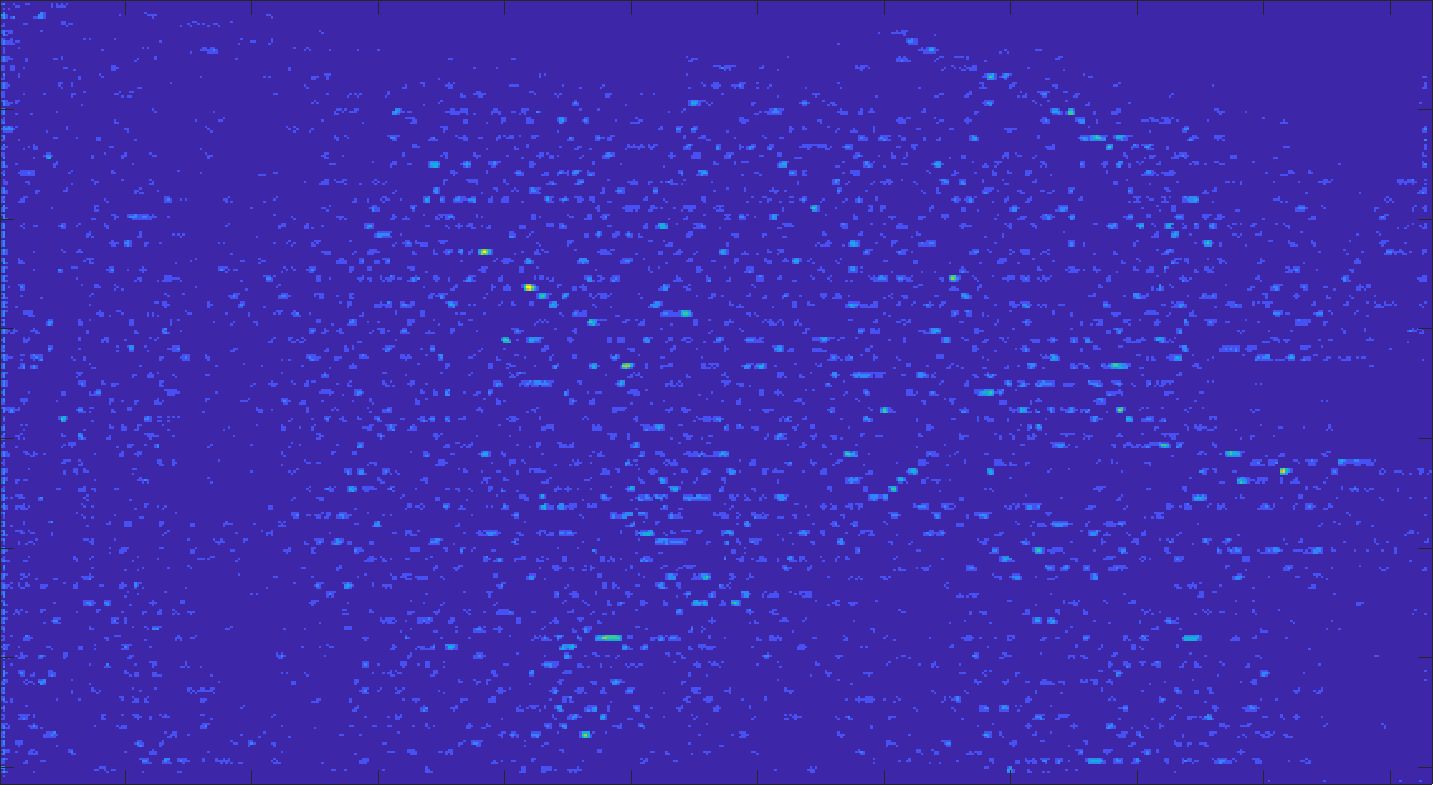} & 
\includegraphics[width=0.15\textwidth]{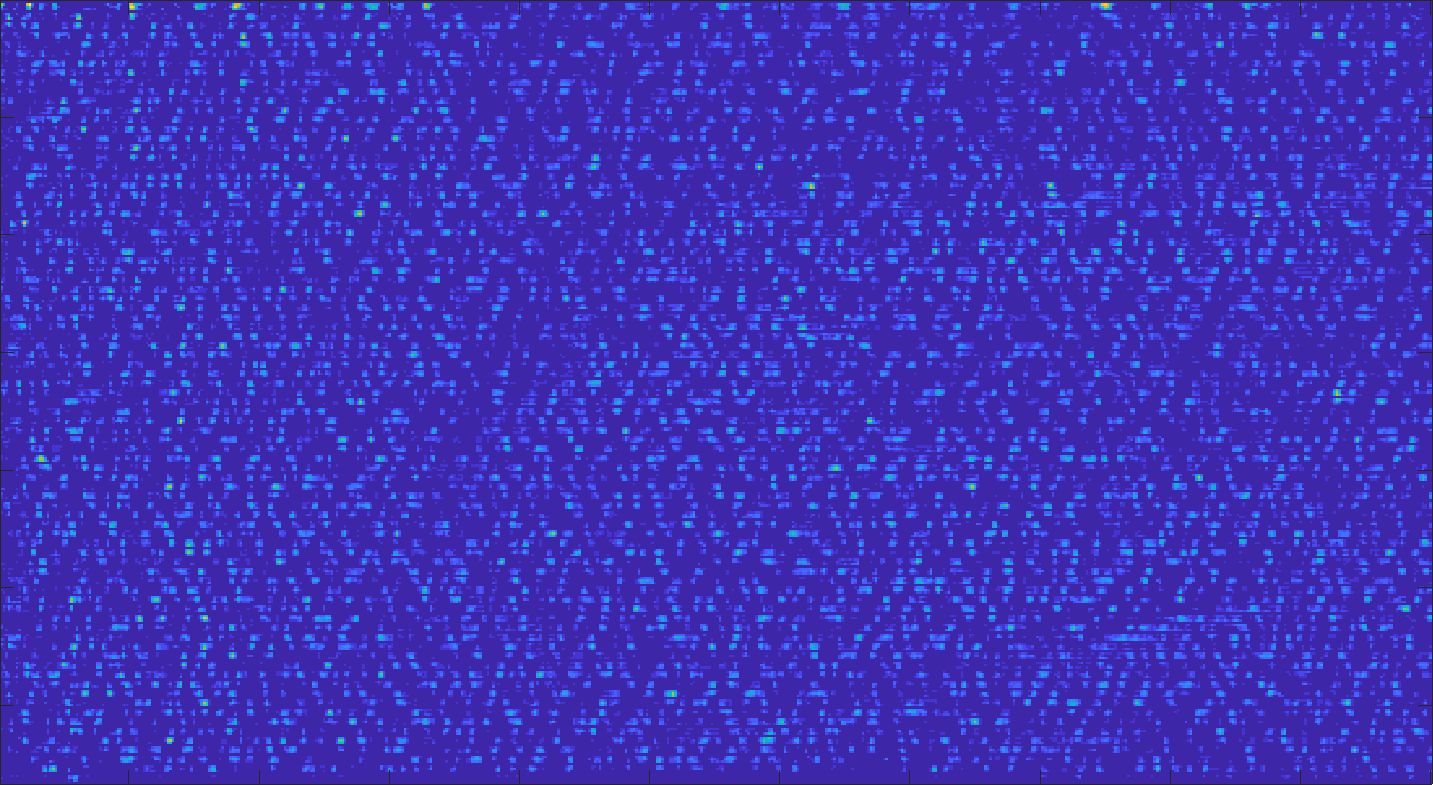} & \hspace*{-0.375cm}
\includegraphics[width=0.15\textwidth]{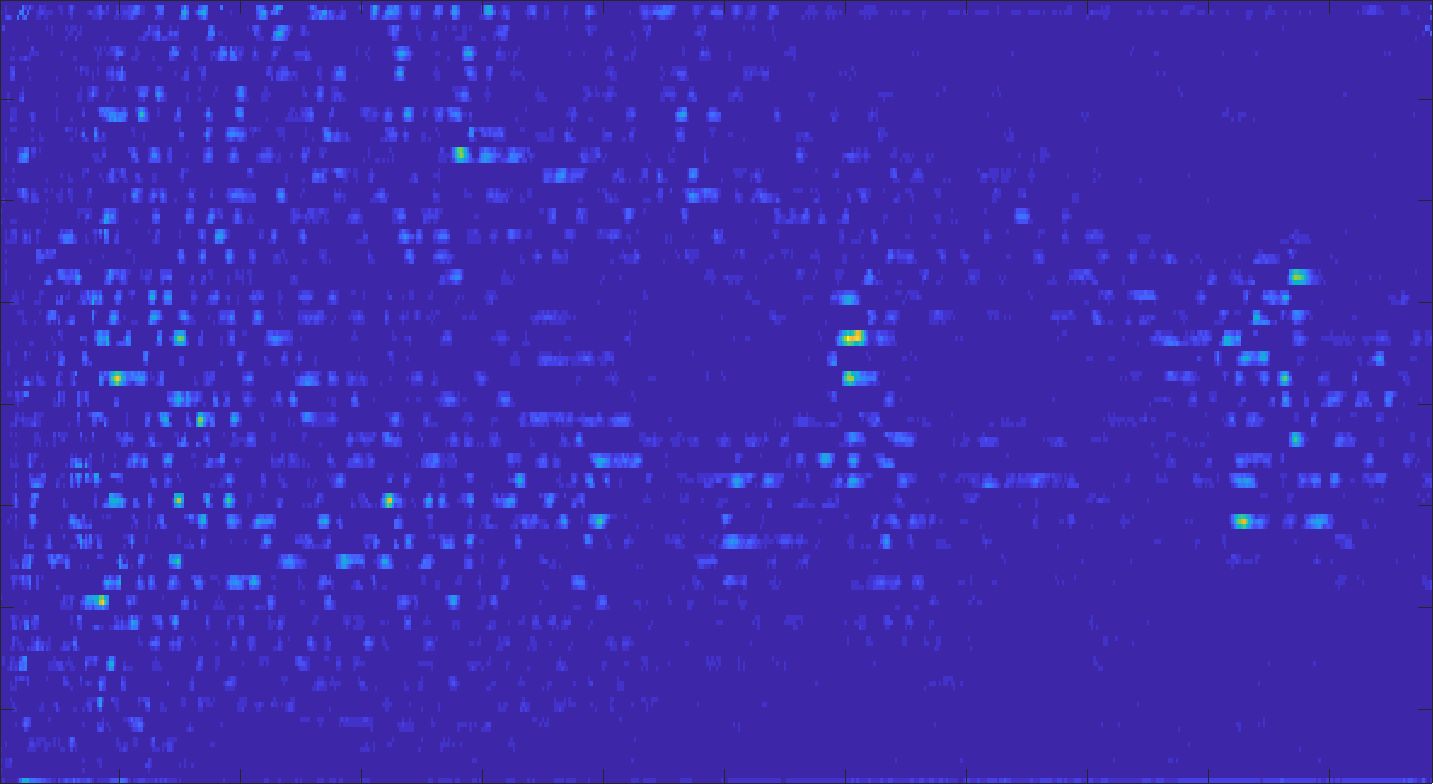} & \hspace*{-0.375cm}
\includegraphics[width=0.15\textwidth]{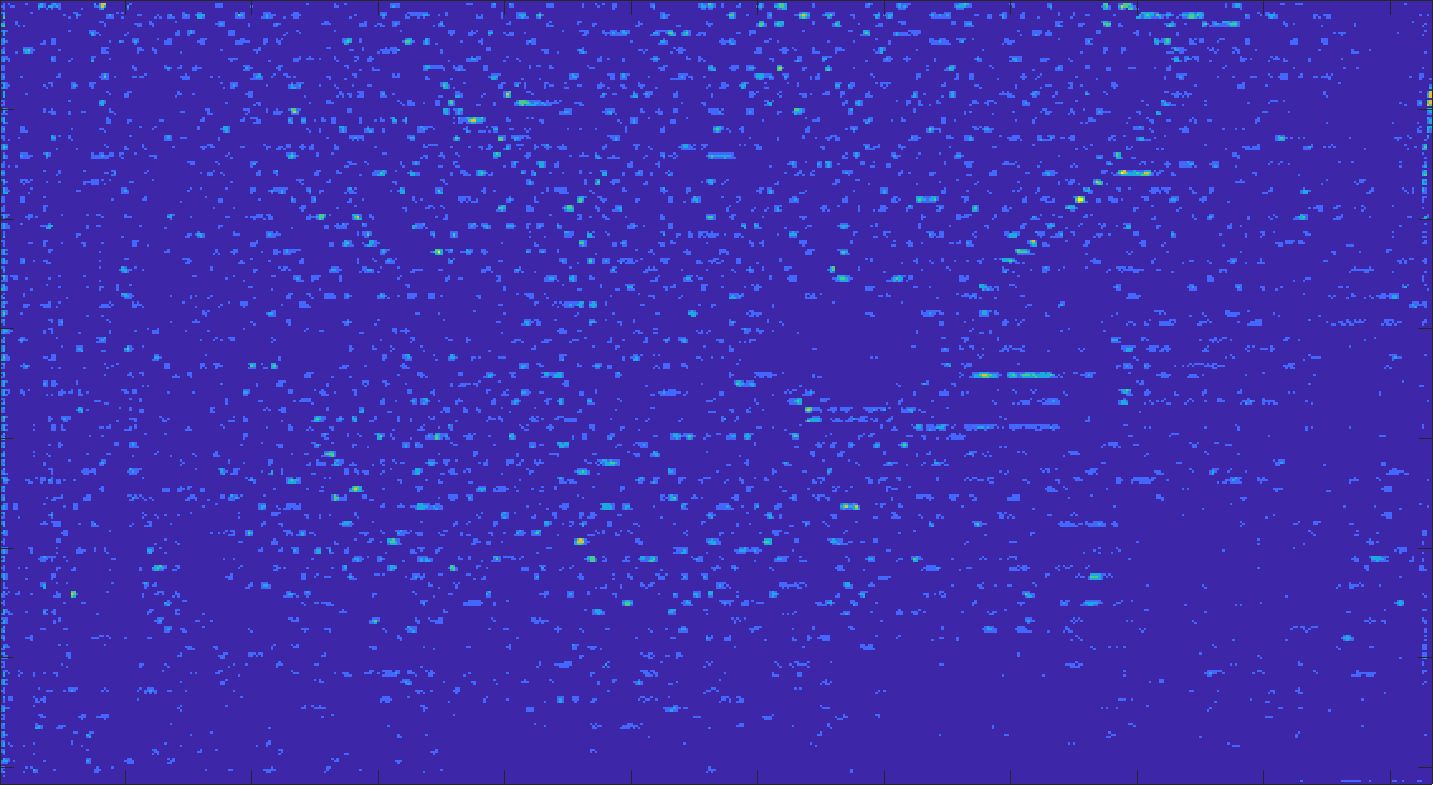} & \hspace*{-0.3cm}
\includegraphics[height=45pt]{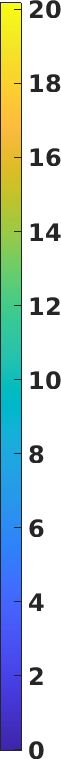}
\end{tabular}
\caption{Concerning Figs.~\ref{FIG:OBNETWORK},~\ref{FIG:CARDIONETWORK},~\ref{FIG:ABDONETWORK}, we report the absolute value of the distance between the input and the prediction, for both 2X (first row) and 4X (second row) up-sampling factors. The absolute value image shows the changes brought about by the prediction of the neural network, which are mainly located at the edges of anatomical structures, with a maximum value of 20 in the 0-255 grey intensity scale.\label{FIG:DELTA}}
\end{figure*}
\paragraph{Proposed super-resolution of US images}
Our framework is composed of two steps: first, we up-sample the low-resolution image through an interpolating method. After the comparison of state-of-the-art methods~(Sect.~\ref{SEC:EXUPSAMPLING}), we select \emph{Cubic Convolution} as the up-sampling algorithm. Then, we apply a learning-based network to improve the visual accuracy of the up-sampling.

For the experimental part, we consider the \emph{Esaote data set}, which contains more than 10K US images at different resolutions, and is acquired from different anatomical districts (e.g., obstetric, cardiac). Given a high-resolution image (i.e., the target) acquired by the probe, we build the corresponding low-resolution image by removing one line each 2 (0.5X) or 4 (0.25X). This approach is consistent with the acquisition of the US image, where the probe can acquire at the full, half, or a quarter of the maximum number of beamlines, depending on the activation of the piezoelectric crystals. We up-sample the low-resolution images through \emph{Cubic Convolution} at 2X (applied to 0.5X low-resolution) or 4X (applied to 0.25X low-resolution). Then, we use the couples of up-sampled and target high-resolution images to analyse the proposed framework, through the training and the prediction of the learning-based network, with a specialisation in anatomic districts. 
\begin{figure*}[t]
\centering
\begin{tabular}{c|cc}
Target & Input & Prediction \\
\includegraphics[width=0.3\textwidth]{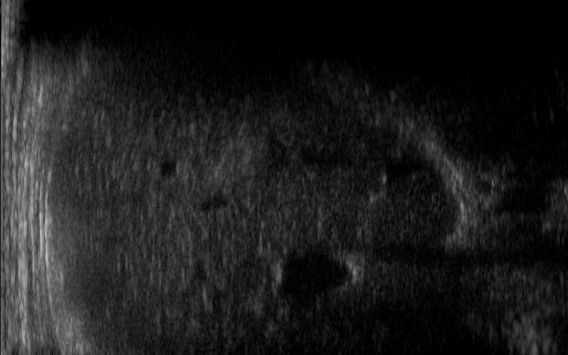} &
\includegraphics[width=0.3\textwidth]{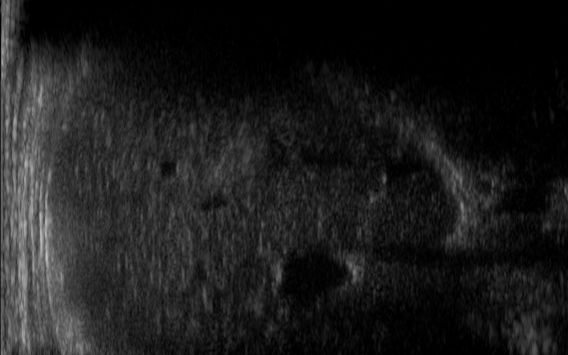} &
\includegraphics[width=0.3\textwidth]{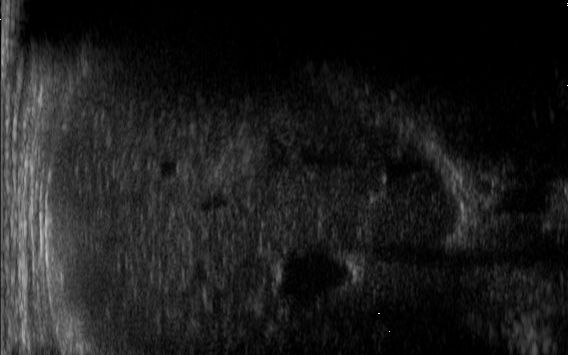} \\ [0mm]
Brightness: 28 & PSNR: 47.51 & PSNR: 47.5 \\ \hline
\includegraphics[width=0.3\textwidth]{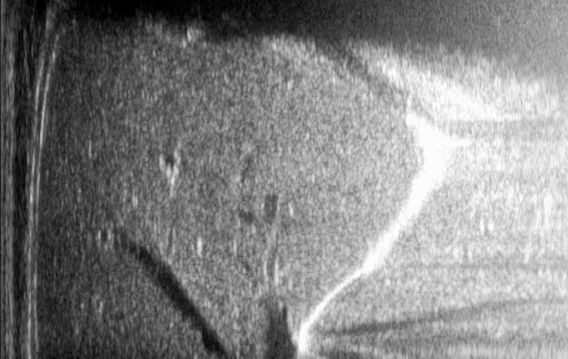} &
\includegraphics[width=0.3\textwidth]{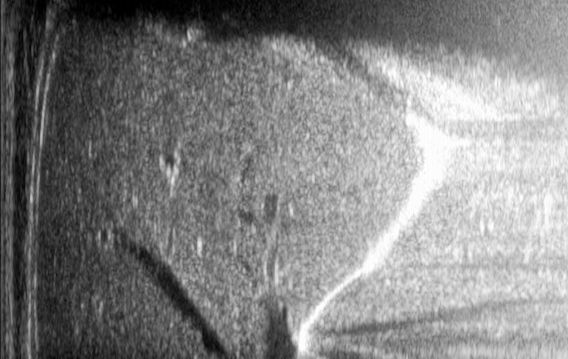} &
\includegraphics[width=0.3\textwidth]{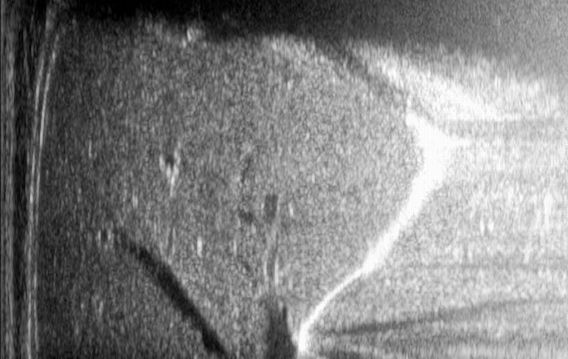} \\
Brightness: 131 & PSNR: 43.46 & PSNR: 43.55
\end{tabular}
\caption{Input and prediction of the raw images of the abdominal district 2X with different levels of brightness: low brightness (first row) and high brightness (second row).\label{FIG:BRIGHT2X}}
\end{figure*}

We generate a separate training data set of 1.5K images for each anatomical district and two different up-sampling resolutions of 2X and 4X. Then, the same images are denoised through a low-rank denoising algorithm~\cite{cammarasana2022learning} to build the training data set of 1.5K denoised images for each anatomical district. In total, we train 12 networks (i.e., 3 anatomical districts, 2 up-sampling factors, 2 (raw/denoised) images), each with 1.5K images as a training data set. In addition, for each anatomical district, up-sampling factor, and raw/denoised images we generate a validation data set of 400 images and a test data set of 200 images, using each image in only one of the three data sets.

Our approach requires the interpolation of the missing rows to the up-sampling method, while the learning model deals with the prediction of the target values from the interpolated values of the up-sampling method. Given two images~$\mathbf{A}$ and~$\mathbf{B}$ both of size~$m \times n$, as \emph{quantitative metrics} we consider the \emph{peak-signal-to-noise ratio} \mbox{$PSNR(\mathbf{A},\mathbf{B})=10\log_{10}\frac{ (\max(\mathbf{A}))^2 }{MSE(\mathbf{A},\mathbf{B})}$}, where we define the \emph{mean squared error} \mbox{$MSE(\mathbf{A},\mathbf{B})=\frac{1}{m\times n} \sum_{i=1}^{m} \sum_{j=1}^{n} (\mathbf{A}_{ij} - \mathbf{B}_{ij})^2$},  the \emph{structural similarity index measure}
\begin{equation*}
\begin{array}{l}
SSIM(\mathbf{A},\mathbf{B}) = l(\mathbf{A},\mathbf{B}) \times c(\mathbf{A},\mathbf{B}) \times s(\mathbf{A},\mathbf{B}),\\
l(\mathbf{A},\mathbf{B})=\frac{2\mu_{\mathbf{A}}\mu_{\mathbf{B}}+C_1}{\mu_{\mathbf{A}}^2 + \mu_{\mathbf{A}}^2 + C_1 },
\end{array}
\end{equation*}
\begin{equation*}
\begin{array}{l}
c(\mathbf{A},\mathbf{B}) = \frac{2\sigma_{\mathbf{A}}\sigma_{\mathbf{B}}+C_2}{\sigma_{\mathbf{A}}^2 + \sigma_{\mathbf{A}}^2 + C_2},\qquad
s(\mathbf{A},\mathbf{B}) = \frac{\sigma_{\mathbf{AB}} + C_3}{ \sigma_{\mathbf{A}} \sigma_{\mathbf{B}} + C_3},
\end{array}
\end{equation*}
where~$\mu(\cdot)$ is the mean of~$(\cdot)$,~$\sigma(\cdot)$  is the standard deviation of~$(\cdot)$,~$\sigma_{\mathbf{AB}}$  is the covariance between~$\mathbf{A}$ and~$\mathbf{B}$, the positive constants~$C_1$,~$C_2$ and~$C_3$ are used to avoid a null denominator. We also consider the \emph{mean absolute error} \mbox{$MAE(\mathbf{A},\mathbf{B})=\frac{1}{m\times n}\sum_{i=1}^{m} \sum_{j=1}^{n}|\mathbf{A}_{ij} - \mathbf{B}_{ij}|$}, and the \emph{pointwise absolute error image} \mbox{$| \mathbf{A}- \mathbf{B}|$} for the comparison of the high-resolution target with both the up-sampled image and the prediction of the network. We also compare the histogram of the absolute value of the prediction error to analyse the number of pixels whose error is lower than a certain threshold.

\paragraph{Deep learning network}
We select WDSR~\cite{yu2020wide}, an architecture that exploits residual blocks since it improves the prediction of images where the difference between the input and the target is small. We propose a customised version of this network: \emph{custom-WDSR}. In particular, our network architecture is a variant of WDSR-A, where the expansion of the features before the rectified linear unit (ReLU) activation allows more information to pass through while preserving the non-linearity of the network. After the normalisation of the data, we apply a 2D convolution and a weighted normalisation that improves the conditioning of the optimisation problem and thus the convergence. Then, we apply 8 residual blocks with wide activation where each residual block is composed of two convolution layers with ReLU activation and a final 2D convolution with a weighted normalisation layer; finally, we combine residual blocks and convolution layers and apply the denormalisation. The kernel filter size depends on the up-sampling factor:~$(3 \times 3)$ in 2X up-sampling, and~$(5 \times 5)$ in 4X up-sampling. The convolution layer does not need to perform the interpolation of the missing values, since this operation has already been performed by the up-sampling algorithm. For this reason, we did not implement the WDSR-B network which adds a linear low-rank convolution and neither pixel shuffling for the deconvolution operation.
With this setting, the total number of trained parameters is 889K for 2X network and 253K for 4X network.
\begin{figure*}[t]
\centering
\begin{tabular}{c|cc}
Target & Input & Prediction \\
\includegraphics[width=0.3\textwidth]{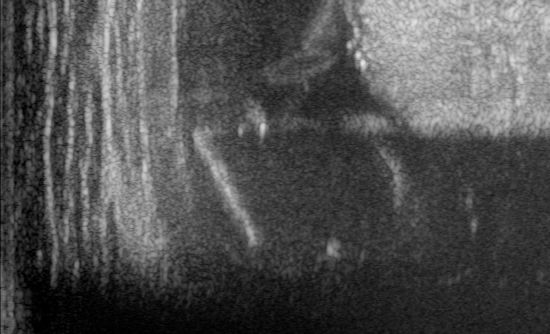} &
\includegraphics[width=0.3\textwidth]{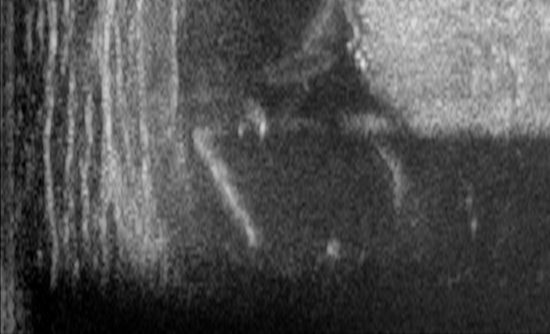} &
\includegraphics[width=0.3\textwidth]{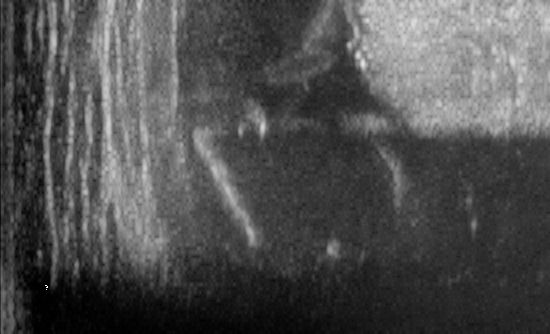} \\ [0mm]
Brightness: 54 & PSNR: 31.01 & PSNR: 31.48 \\ \hline
\includegraphics[width=0.3\textwidth]{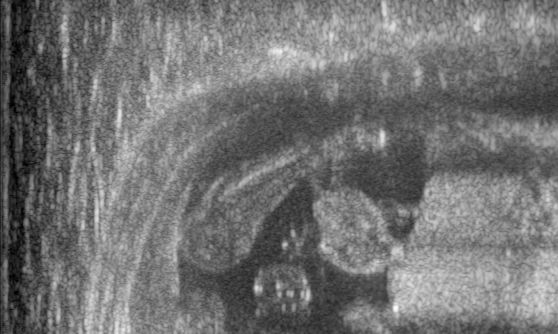} &
\includegraphics[width=0.3\textwidth]{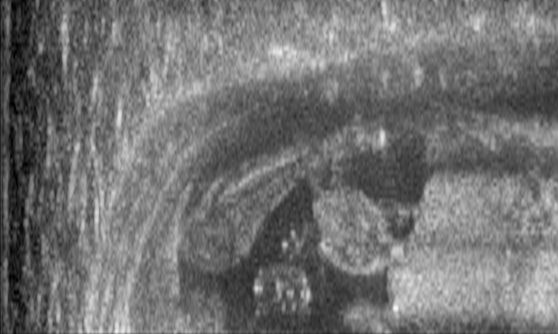} &
\includegraphics[width=0.3\textwidth]{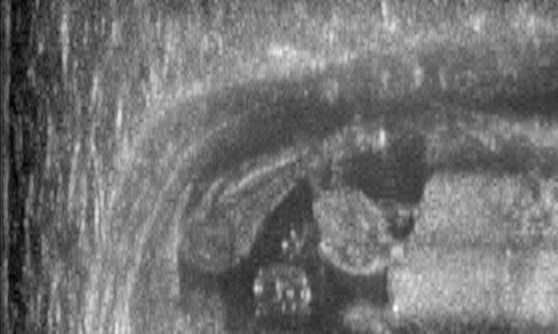} \\
Brightness: 108 & PSNR: 30.02 & PSNR: 30.45
\end{tabular}
\caption{Input and prediction of the raw images of the obstetric district 4X with different levels of brightness: low brightness (first row) and high brightness (second row).\label{FIG:BRIGHT4X}}
\end{figure*}

Given an~$y=L \times D$ target image, and its approximation~$\hat{y}$, our loss function is defined as
\begin{equation*}
Loss(y,\hat{y}) = \begin{dcases}
\sum_{l,d=1}^{L,D}\log\frac{\vert y_{ld}-\hat{y}_{ld} \vert + \epsilon}{k}, \quad \bmod(l,s)=0, \\ 
0,  \quad \textup{otherwise,}
\end{dcases}
\end{equation*}
where~$s$ controls the number of lines acquired by the sensor (e.g.,~$s=4$ when 4X up-sampling is applied) and neglecting their contribution to the training loss; \mbox{$\epsilon=10^{-4}$} avoids a null error for the logarithmic loss, and~$k=5$ determines the curvature of the logarithmic loss function. We enhance the pixels where the loss is less than 5 on the 0-255 range, to improve the visual similarity between the prediction and the target image. We underline that data are normalised in the range 0-1, and consequently the~$k$ value is set to~$5/255 \approx 0.019$. The value of~$\epsilon$ is selected sufficiently smaller than~$1/255\approx 4\cdot10^{-3}$, which is the normalisation of the smallest possible difference value between two pixels; we have experimentally set~$\epsilon=1\cdot10^{-4}$. The size of the kernel of the convolution filter depends on the up-sampling factor; in the case of a 2X up-sampling, we apply a~$3\times3$ filter; for a 4X up-sampling, we apply a~$5\times5$ filter. This choice allows us to include at least two lines acquired by the probe, in the convolution operator. Finally, we set the number of layers to 16 and the number of kernels to 10. The learning rate iteratively decreases, up to~$10^{-6}$, and the number of epochs is set to 200. The input and output layers of the network are~$\#batch \times L \times D$ size.
\begin{figure*}[t]
\centering
\begin{tabular}{ccc|c }
Target & Down-sampling & Cubic Convolution & Our  \\
\includegraphics[width=0.23\textwidth]{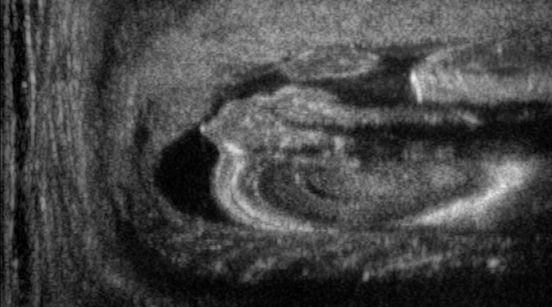} &
\includegraphics[width=0.23\textwidth]{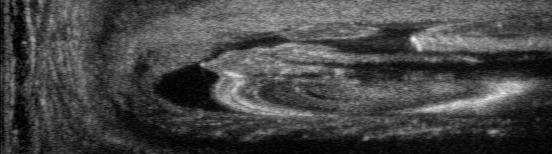} &
\includegraphics[width=0.23\textwidth]{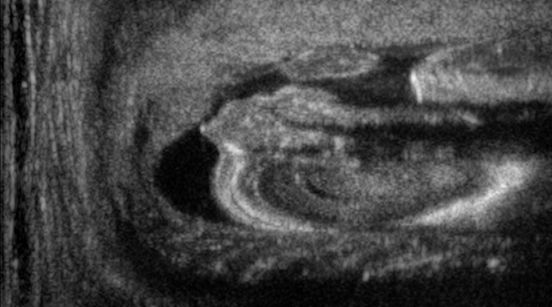} &
\includegraphics[width=0.23\textwidth]{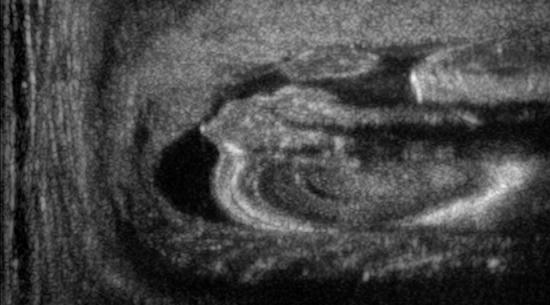} \\
EDSR & SRGAN & SISR & \\ 
\includegraphics[width=0.23\textwidth]{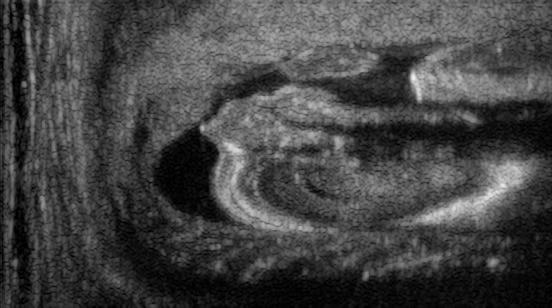}&
\includegraphics[width=0.23\textwidth]{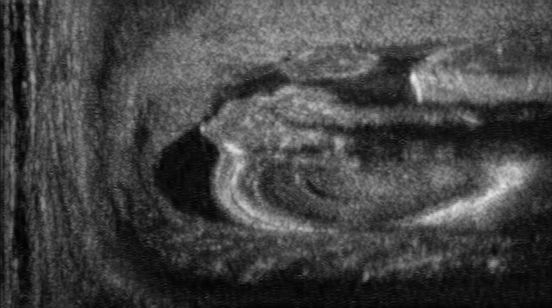} &
\includegraphics[width=0.23\textwidth]{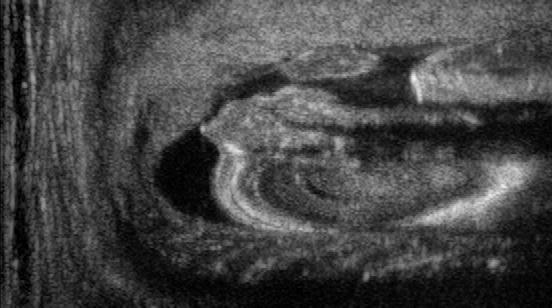} &
\end{tabular}
\caption{Comparison of up-sampling methods vs. our method on the obstetric district: 0.5X low-resolution and 2X up-sampling. See also Table~\ref{TAB:PSNRTEST}.\label{FIG:05COMPARISON}}
\end{figure*}
\section{Experimental results\label{SEC:RESULTS}}
We discuss the results of the proposed super-resolution of 2D US images (Sect.~\ref{sec:US-IMAGE-SYPERES}) and compare our results with previous work (Sect.~\ref{SEC:EXUPSAMPLING}); we present the results with 2D US videos (Sect.~\ref{sec:US-VIDEO-SYPERES}) and noisy images (Sect.~\ref{SEC:DENOISING}), and discuss the execution time (Sect.~\ref{SEC:EXTIME}).
\begin{figure*}[t]
\centering
\begin{tabular}{ccc|c}
Target & Down-sampling & Cubic Convolution & Our  \\
\includegraphics[width=0.22\textwidth]{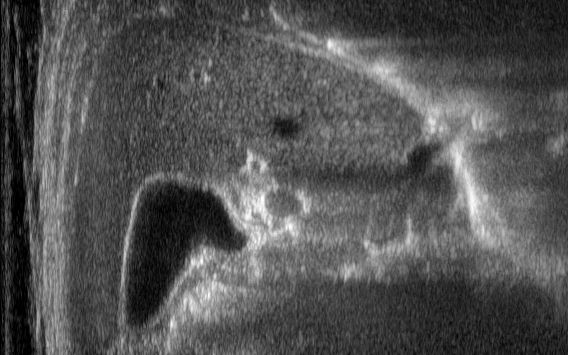} &
\includegraphics[width=0.22\textwidth]{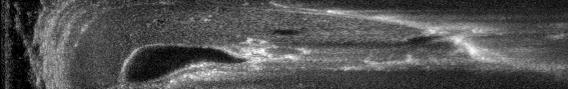} &
\includegraphics[width=0.22\textwidth]{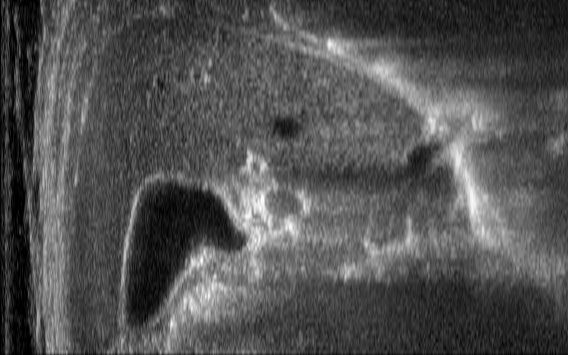} &
\includegraphics[width=0.22\textwidth]{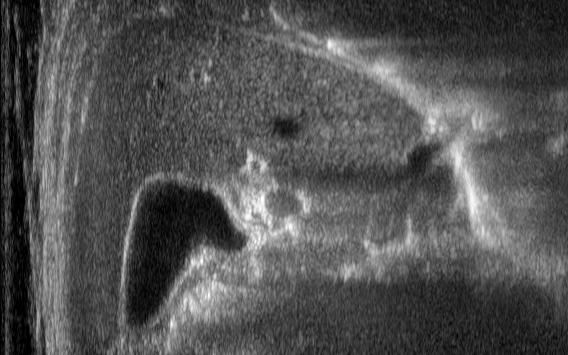} \\
EDSR & SRGAN & SISR & \\ 
\includegraphics[width=0.22\textwidth]{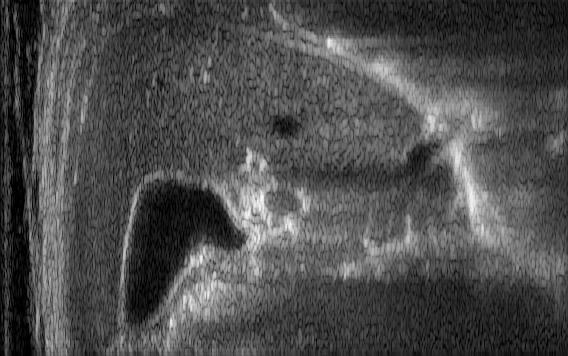}&
\includegraphics[width=0.22\textwidth]{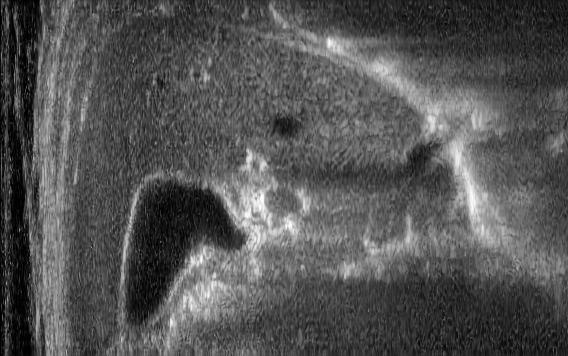} &
\includegraphics[width=0.22\textwidth]{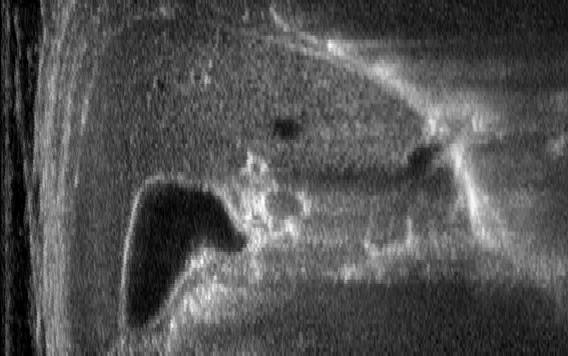} &
\end{tabular}
\caption{Comparison of up-sampling methods vs. our method on the abdominal district: 0.25X low-resolution and 4X up-sampling. See also Table~\ref{TAB:PSNRTEST}.\label{FIG:025COMPARISON}}
\end{figure*}
\subsection{Proposed super-resolution of US images\label{sec:US-IMAGE-SYPERES}}
We train each learning-based network (\emph{custom-WDSR}) with 1.5K images, where the input is the outcome of the selected up-sampling method (i.e., \emph{Cubic convolution}), and the target is the original high-resolution image. Indeed, input and target images have the same resolution, as the reconstruction of the missing lines has been already performed by \emph{Cubic convolution}. Figs.~\ref{FIG:OBNETWORK},~\ref{FIG:CARDIONETWORK}, and~\ref{FIG:ABDONETWORK} show the results of the network prediction, compared with the input and the target images. Target images correspond to spatial high-resolution images; input images are the outcome of the up-sampling interpolation, which is applied to spatial low-resolution images (i.e., the down-sampling along the lateral direction of high-resolution images); prediction images represent the output of the neural network.

Our framework visually improves the results, in terms of blurring and artefacts. This result is more evident in the magnification of the ear of the foetus (Fig.~\ref{FIG:OBNETWORK}), the mitral valve (Fig.~\ref{FIG:CARDIONETWORK}), and the mass edges (Fig.~\ref{FIG:ABDONETWORK}). Fig.~\ref{FIG:ERRORIMAGE} shows the error image of the three anatomical districts with both 2X and 4X up-sampling factors, with the maximum error in the scale~$0-255$. The error is more evident in the contours of the anatomical structures; moreover, the abdominal district shows a smaller error than the cardiac and obstetric ones. We underline that the view for each image is scaled to its maximum, to improve the visualisation of the error.

Fig.~\ref{FIG:OBQUANTITATIVE}(a-b-c, left) shows the box plot of the statistics of the PSNR on three different anatomical districts, comparing the target images with the prediction and the cubic convolution, respectively. The metrics are computed on a data set of 200 images of the same district and with the same up-sampling factor. We report that the PSNR median value improves of~$1.7\%$ on obstetric 2X raw images,~$6.1\%$ on cardiac 2X raw images, and~$4.4\%$ on abdominal raw 4X images. 
\begin{figure*}[t]
\centering
\begin{tabular}{cc}
2X & 4X \\ \hline
Max. error: 18 & Max. error: 35 \\
\includegraphics[width=0.25\textwidth]{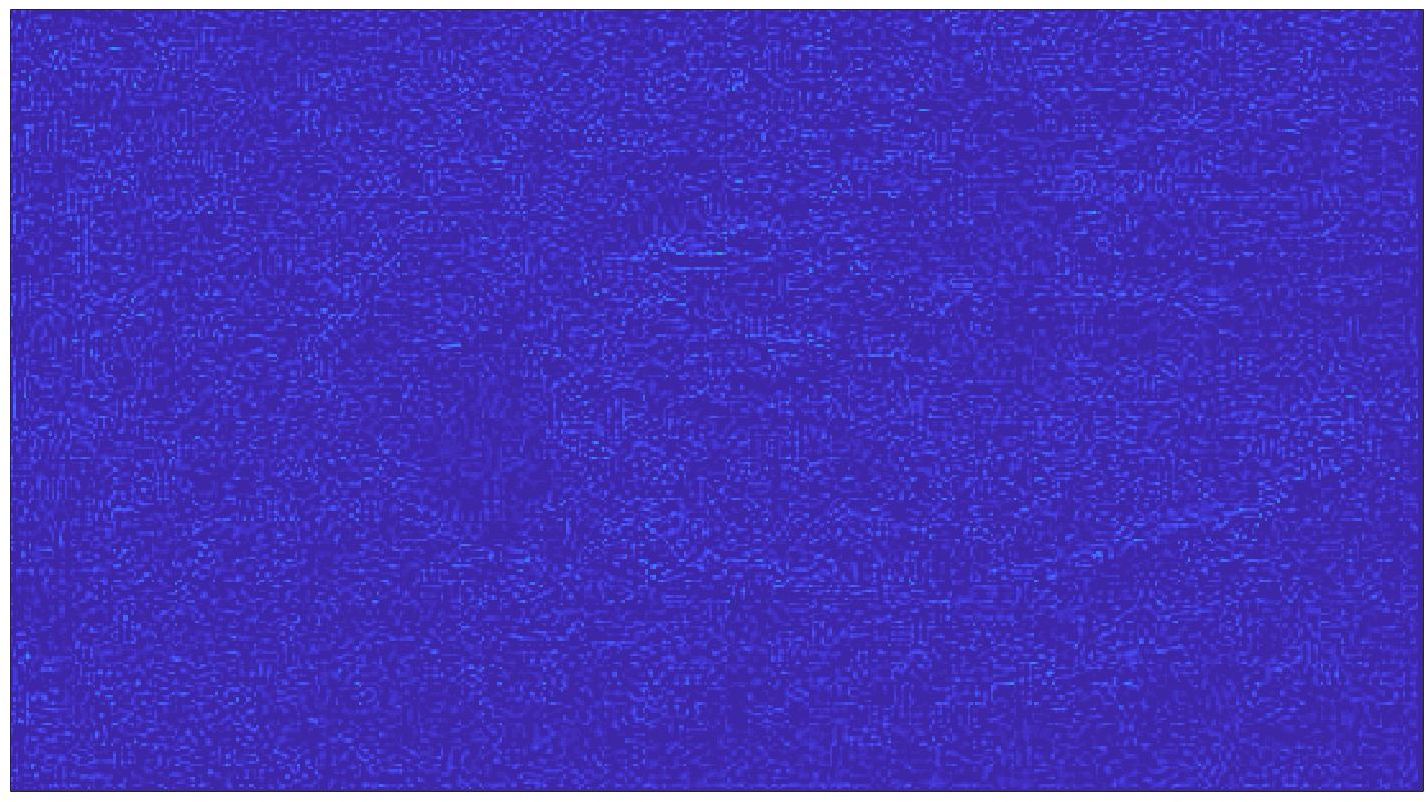} &
\includegraphics[width=0.25\textwidth]{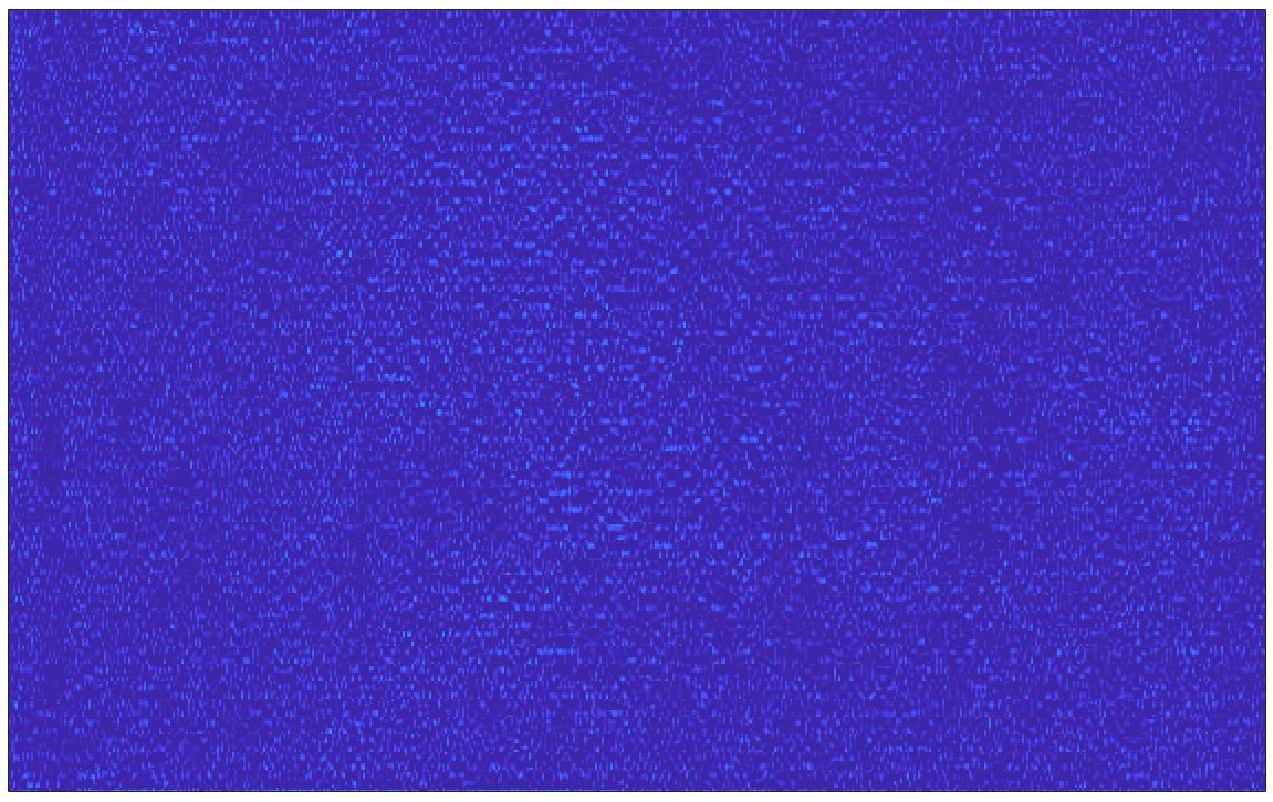} \\
Max. error: 33 & Max. error: 62 \\
\includegraphics[width=0.25\textwidth]{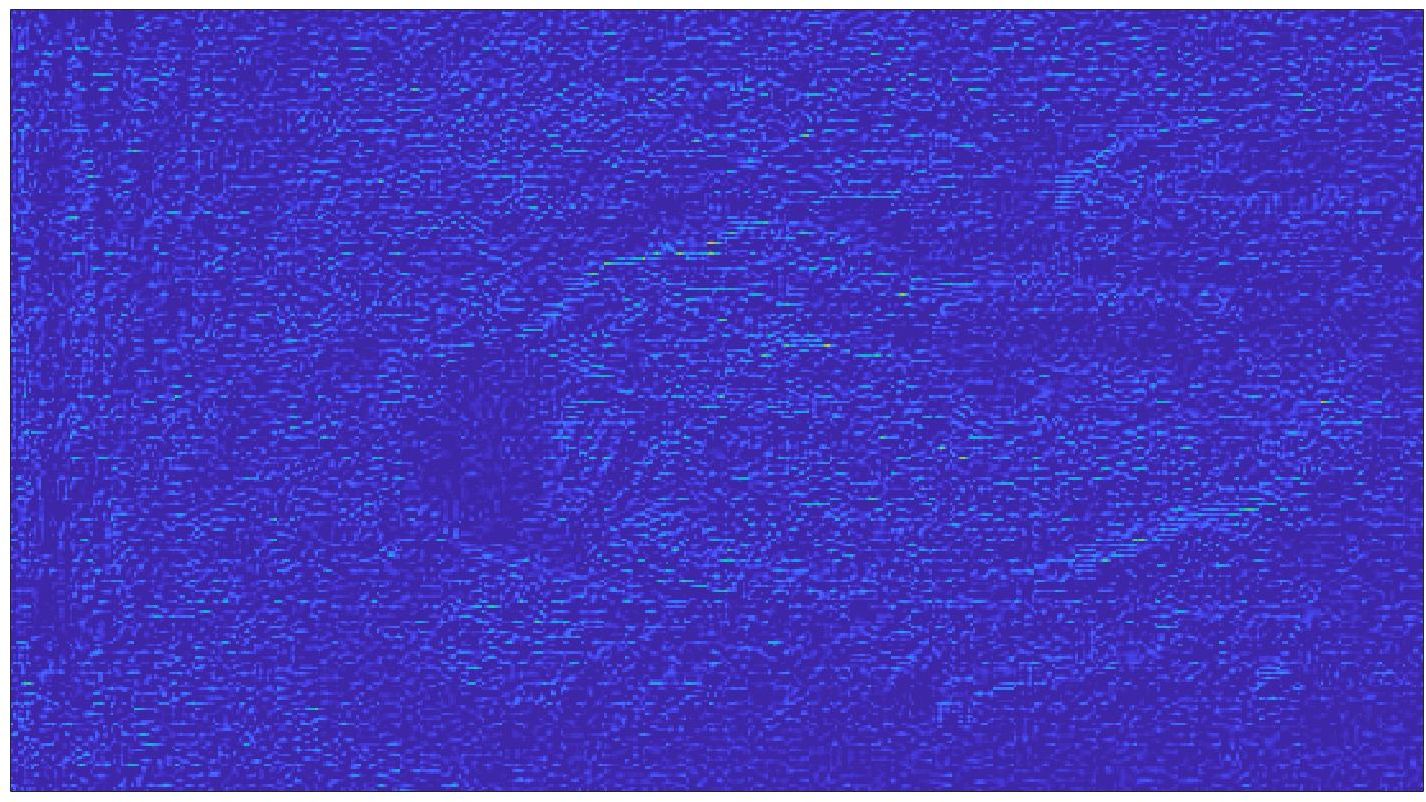} &
\includegraphics[width=0.25\textwidth]{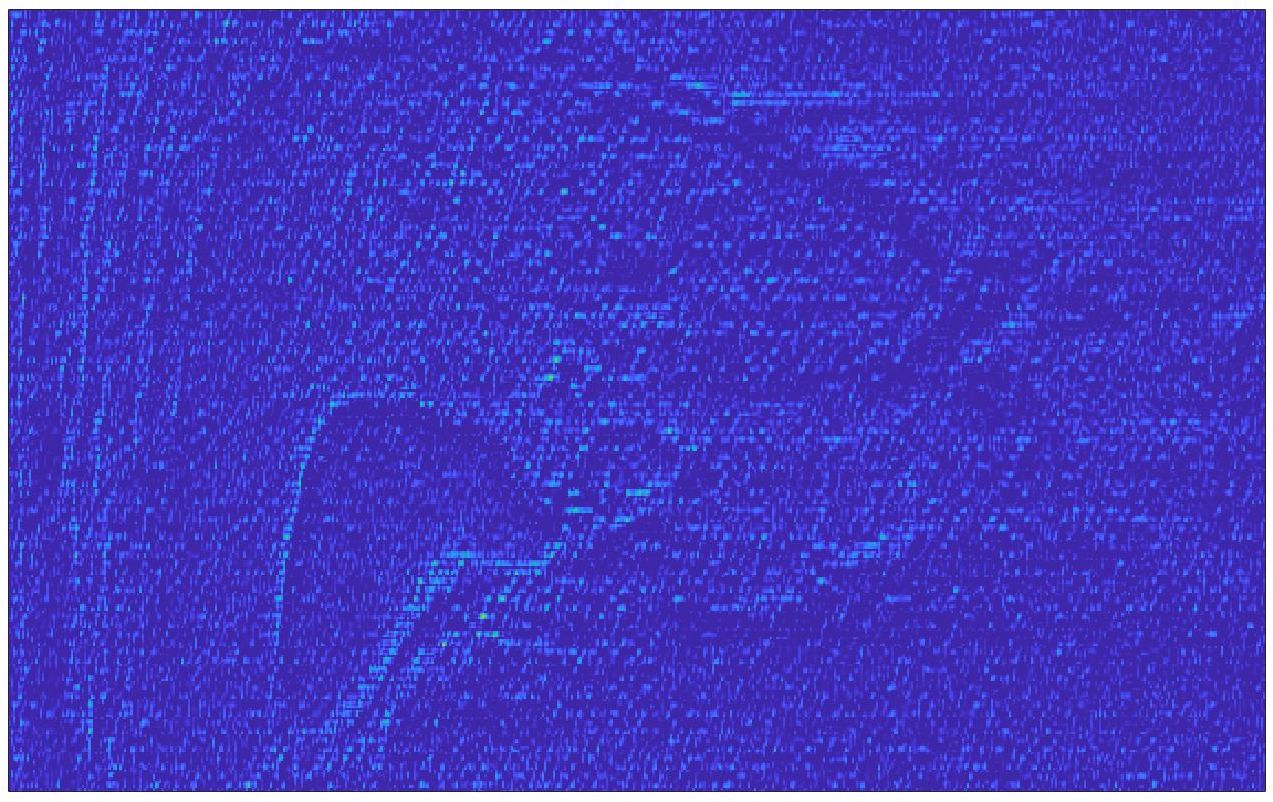} \\
Max. error: 40 & Max. error: 80 \\
\includegraphics[width=0.25\textwidth]{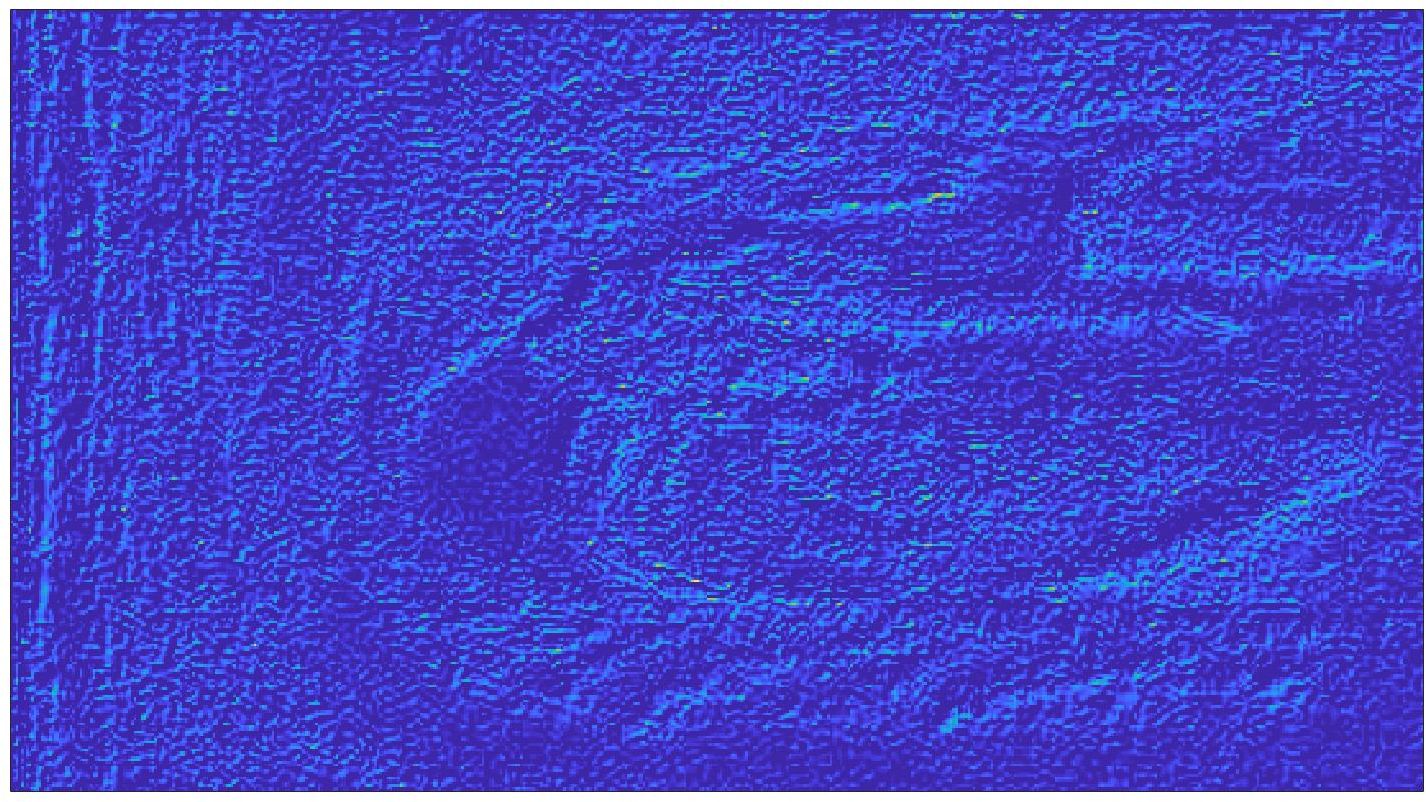} &
\includegraphics[width=0.25\textwidth]{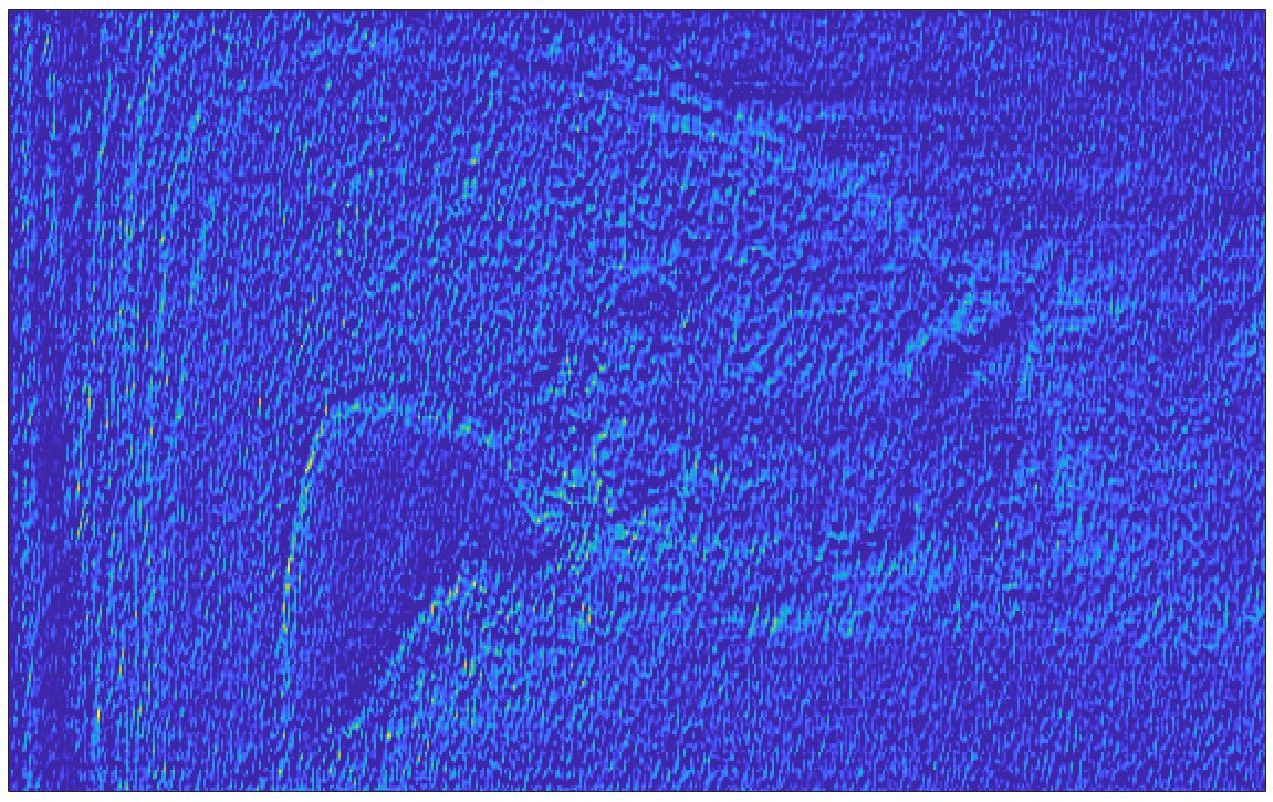} \\
Max. error: 41 & Max. error: 75 \\
\includegraphics[width=0.25\textwidth]{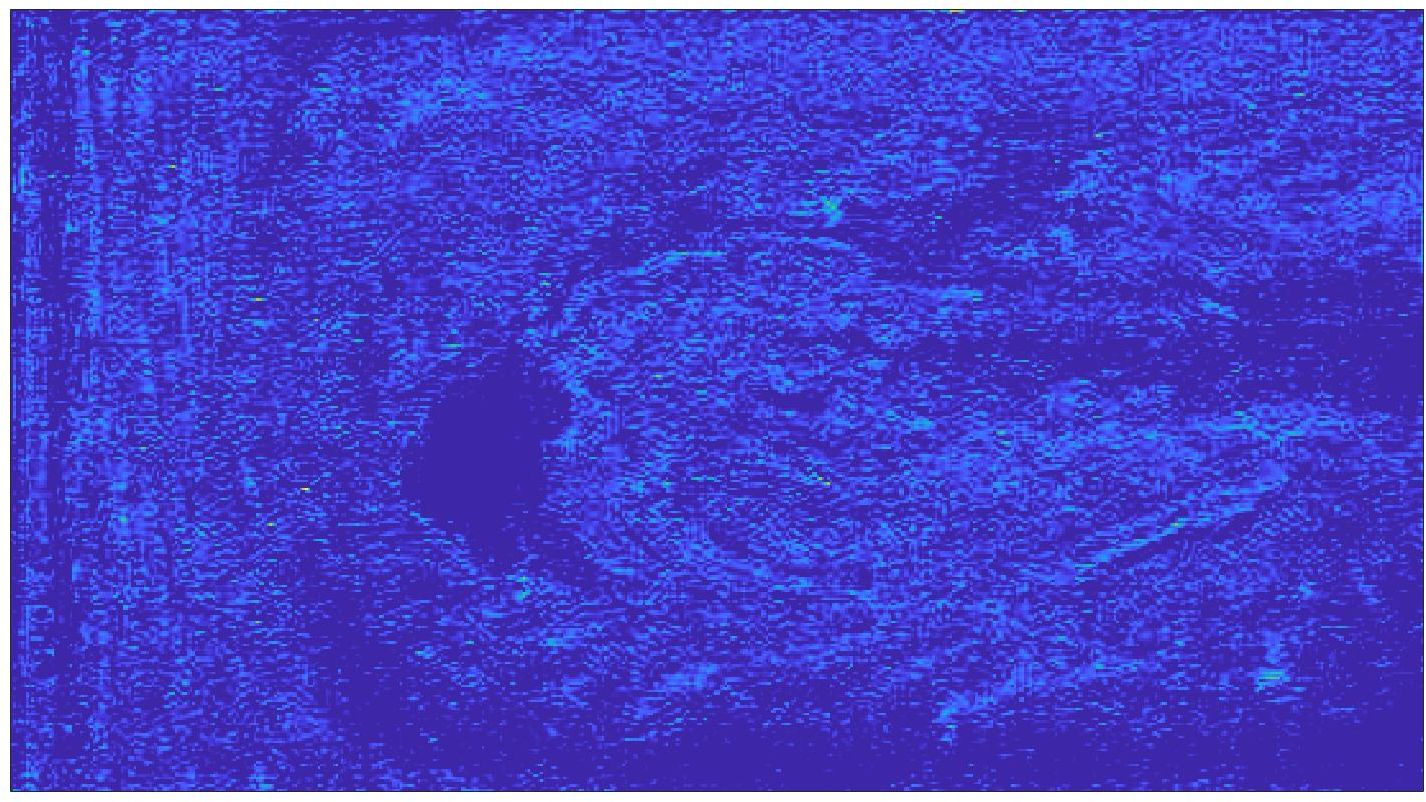} &
\includegraphics[width=0.25\textwidth]{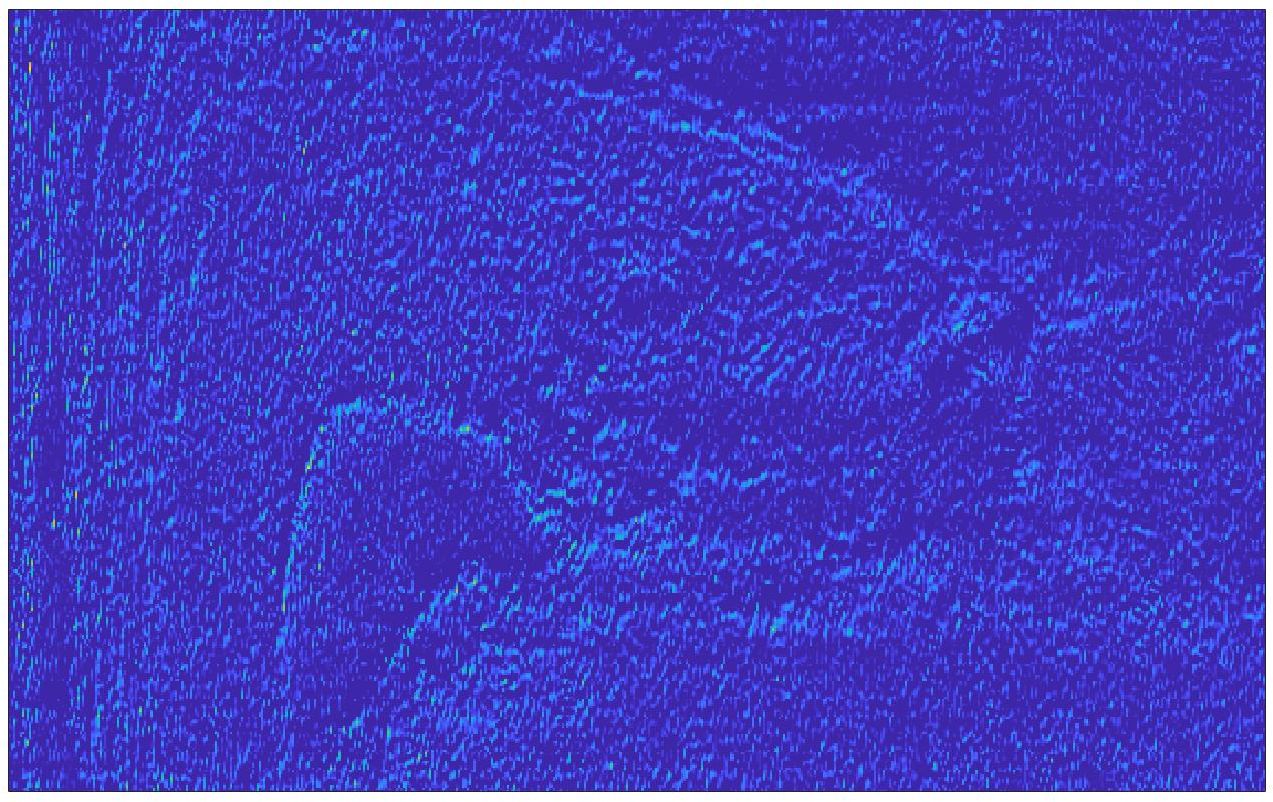} \\ \hline
Max. error: 12 & Max. error: 22 \\
\includegraphics[width=0.25\textwidth]{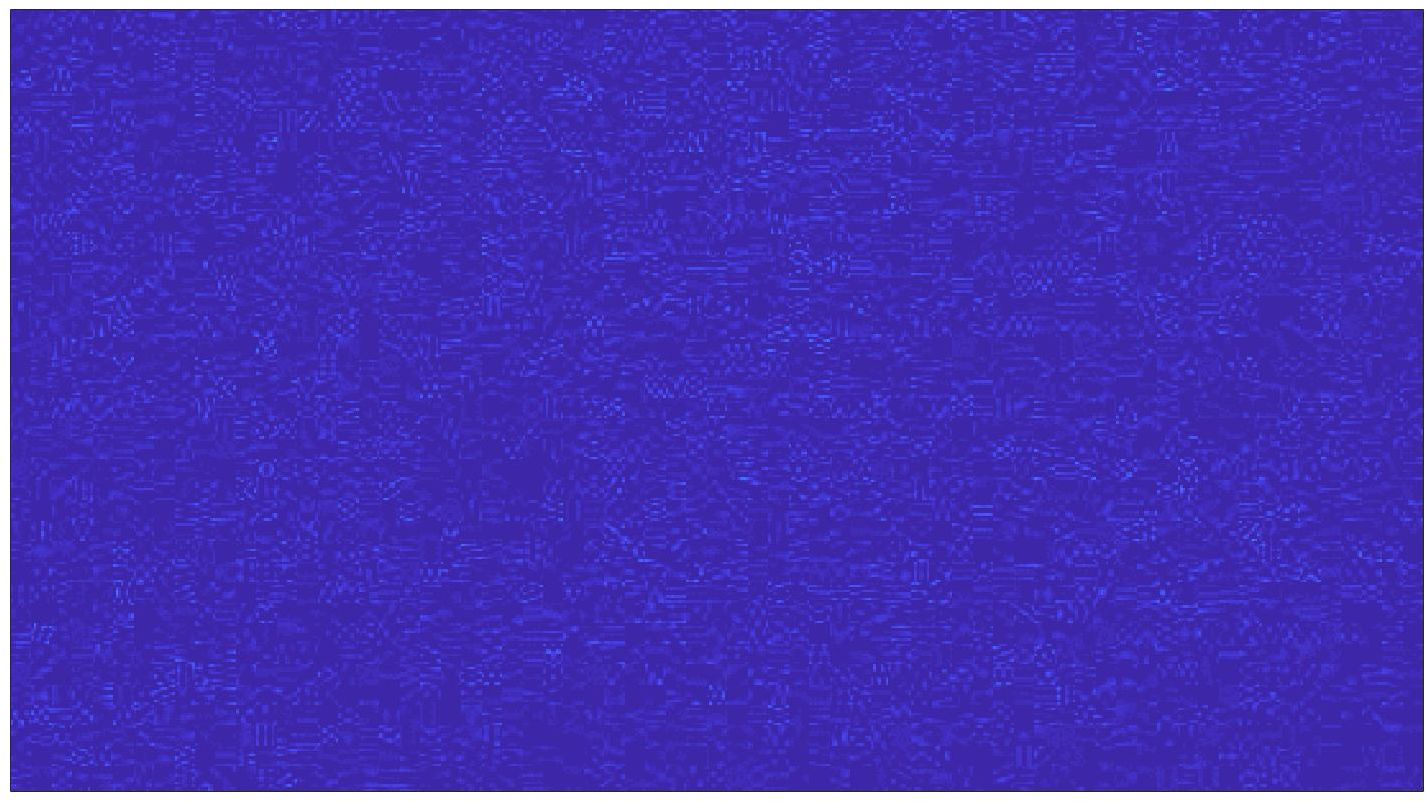} &
\includegraphics[width=0.25\textwidth]{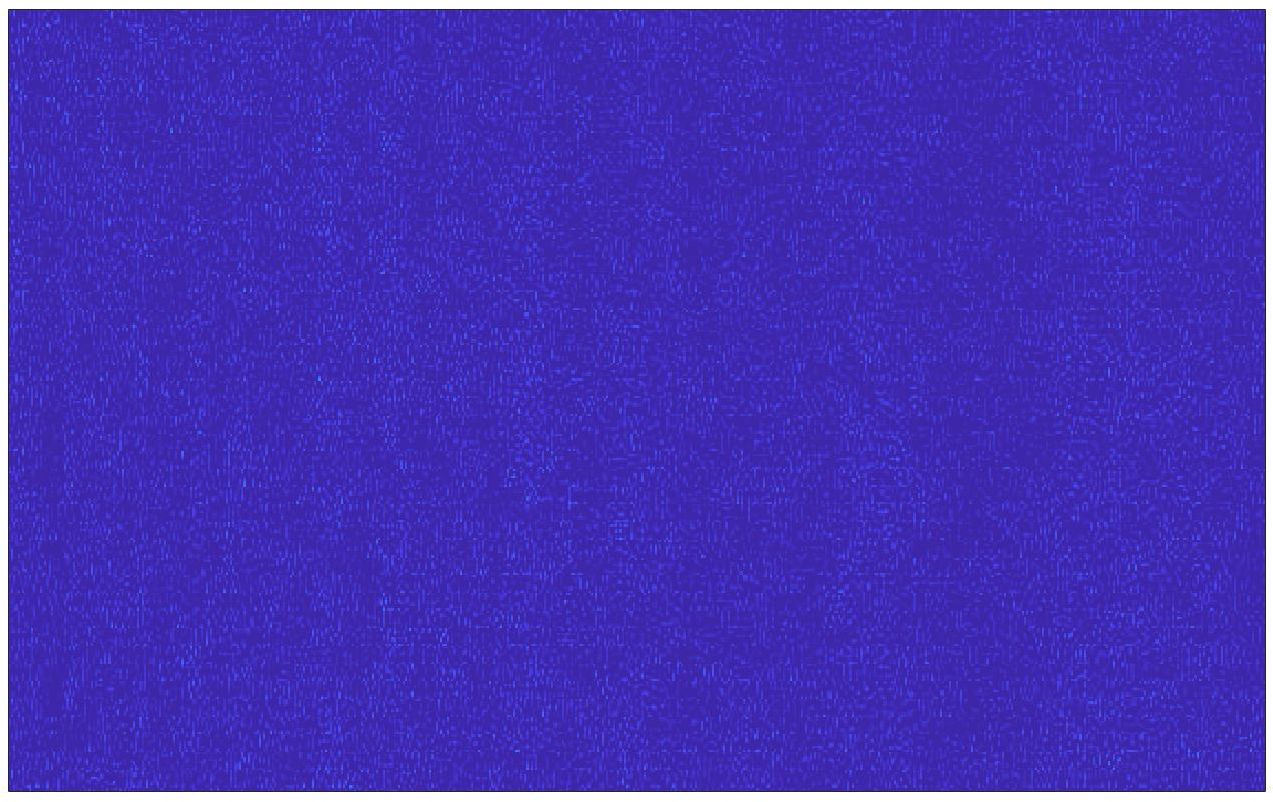}
\end{tabular}
\caption{Error image of SOTA up-sampling methods vs. our method on the obstetric (0.25X low-resolution) and abdominal (4X up-sampling) anatomical district: \emph{cubic convolution} (first row); \emph{SISR} (second row); \emph{EDSR} (third row); \emph{SRGAN} (fourth row); \emph{our} (fifth row). For each image, we report the maximum error value in the range~$0-255$. All the images of the same up-sampling factor (i.e., 2X and 4X) are represented with the same colour scale.\label{FIG:ERRORIMAGECOMPARISON}}
\end{figure*}

Fig.~\ref{FIG:OBQUANTITATIVE}(a-b-c, right) shows the histogram of the absolute value of the error with respect to the target image, of the prediction and \emph{Cubic convolution} results, respectively. The histograms show the number of pixels where the prediction error is lower than 5 (i.e., the first bin of the histogram), which means very similar to the target when visually analysing the images. From the \emph{Cubic convolution} to the predicted images, this value increases of~$9.0\%$ on obstetric 4X raw images,~$5.2\%$ on cardiac 4X raw images, and~$6.2\%$ on abdominal 4X raw images.

Fig.~\ref{FIG:OBQUANTITATIVE2} shows the box plot of the SSIM (a-b-c, left) and MAE (a-b-c, right) quantitative metrics, as performed for PSNR metric. Also, these metrics show that our method improves the results of \emph{Cubic convolution} both in terms of average value and variability. For example, the SSIM median value improves of~$2.5\%$ on obstetric 4X images and the MAE median value improves of~$4.7\%$ on cardiac 2X images.

The analysis of the absolute value of the difference between the input and the prediction of the network~(Fig.~\ref{FIG:DELTA}) shows that the alteration of our prediction to the pixel values ranges from 0 to a maximum absolute value of 20, mainly located on the edges of the anatomical structures; furthermore, the black uniform areas are less affected by the prediction. In terms of the distance between the input and the prediction, we do not observe a significant difference among anatomical districts and between 2X and 4X up-sampling.
\begin{table}[t]
\centering
\caption{Concerning Figs.~\ref{FIG:05COMPARISON},~\ref{FIG:025COMPARISON}, we report the PSNR metric computed between target and up-sampling methods, as the mean value among the 200 test images. \label{TAB:PSNRTEST}}
\begin{tabular}{c|c|c}
Test & Obstetric 2X & Abdominal 4X \\ \hline
Cubic Convolution & 36.52  & 42.17  \\
EDSR & 32.08  & 34.91 \\
SRGAN & 33.70 & 36.35 \\
SISR & 34.75 & 38.58 \\ \hline
OURS &~$\mathbf{37.00}$  &~$\mathbf{44.35}$ 
\end{tabular}
\end{table}

We also verify the robustness of our method on images at different brightness. Characterising the brightness of an image as the average value of all pixels, we test images with high and low brightness on different anatomical districts and up-sampling factors. Figs.~\ref{FIG:BRIGHT2X},~\ref{FIG:BRIGHT4X} show that the prediction performed with our trained network is robust to different values of image brightness, never lowering the output accuracy or generating artefacts. Comparing the input and the prediction of our network with the target image, we improve the PSNR value from 43.46 to 43.55 with high brightness images from the abdominal district 2X up-sampling, and from 31.01 to 31.48 with low brightness images from the obstetric district 4X up-sampling.
\begin{table}[t]
\centering
\caption{Concerning Figs.~\ref{FIG:05COMPARISON},~\ref{FIG:025COMPARISON}, we report the SSIM metric computed between target and up-sampling methods, as the mean value among the 200 test images. \label{TAB:SSIMTEST}}
\begin{tabular}{c|c|c}
Test & Obstetric 2X & Abdominal 4X \\ \hline
Cubic Convolution & 0.935  & 0.904  \\
EDSR & 0.878  & 0.61 \\
SRGAN & 0.902 & 0.632 \\
SISR & 0.927 & 0.773 \\ \hline
OURS &~$\mathbf{0.941}$  &~$\mathbf{0.906}$ 
\end{tabular}
\end{table}
\subsection{Comparison with previous work\label{SEC:EXUPSAMPLING}}
We address both the comparison among state-of-the-art algorithms that are used for the selection of the up-sampling method of our framework and the comparison of our results with previous work. Among up-sampling STAR methods, we test four methods belonging to different classes: \emph{Cubic Convolution}~\cite{keys1981cubic}, a kernel-based interpolating method; \emph{Enhanced Deep Residual Networks - EDSR}~\cite{lim2017enhanced}, a learning-based method trained on generic images; \emph{Enhanced Super-Resolution Generative Adversarial Network Plus - ESRGAN+}~\cite{rakotonirina2020esrgan+}, a learning-based GAN method, specialised on US images with a dedicated training; \emph{Single Image Super Resolution - SISR}~\cite{peleg2014statistical}, an up-sampling method which exploits sparse representations. We evaluate the up-sampling results of the selected methods on different anatomical districts and resolutions: obstetric district with 0.5X down-sampling (Fig.~\ref{FIG:05COMPARISON}); abdominal district with 0.25X down-sampling (Fig.~\ref{FIG:025COMPARISON}). Fig.~\ref{FIG:ERRORIMAGECOMPARISON} shows the error image between target and SOTA super-resolution on both 2X and 4X up-sampling, with the maximum error value in the range~$0-255$: \emph{Cubic convolution} has visually the best results in terms of approximation error. Furthermore, \emph{our method} improves the error image results with respect to \emph{Cubic convolution}, improving the approximation of the target image, including the maximum error. All the error images of each up-sampling factor are represented with the same colour scale to better visualise the differences among the methods.
\begin{table}[t]
\centering
\caption{Concerning Figs.~\ref{FIG:05COMPARISON},~\ref{FIG:025COMPARISON}, we report the MAE metric~$[\cdot 10^{-2}]$ computed between target and up-sampling methods, as the mean value among the 200 test images. \label{TAB:MAETEST}}
\begin{tabular}{c|c|c}
Test & Obstetric 2X & Abdominal 4X \\ \hline
Cubic Convolution & 0.8  & 1.19  \\
EDSR & 1.08  & 7.55 \\
SRGAN & 1.21 & 4.13 \\
SISR & 0.92 & 3.31 \\ \hline
OURS &~$\mathbf{0.75}$  &~$\mathbf{1.16}$ 
\end{tabular}
\end{table}

Tables~\ref{TAB:PSNRTEST}, ~\ref{TAB:SSIMTEST}, ~\ref{TAB:MAETEST} summarise the comparison with the PSNR, SSIM, and MAE metrics on a test data set of 200 images. \emph{Cubic convolution} has a mean PSNR value of 36.52 and 42.17 for 2X and 4X upsampling, respectively. According to these results, we select \emph{Cubic convolution} as the best method for the up-sampling of US images. This method interpolates the missing lines, without generating artefacts. In comparison, our method improves the results of previous work (Fig.~\ref{FIG:05COMPARISON}, Fig.~\ref{FIG:025COMPARISON}, Table~\ref{TAB:PSNRTEST}), with a mean PSNR value of 37.00 and 44.35 for 2X and 4X super-resolution, respectively. Finally, we underline that 4X super-resolution on the abdominal district has better results than 2X super-resolution on the obstetric district, due to the complexity and variety of each anatomic district data set.
\begin{figure*}[t]
\centering
\begin{tikzpicture}[spy using outlines={rectangle,magnification=2,size=1cm,width=1cm,height=1cm,every spy on node/.append style={thick}}] 
\node {
\begin{tabular}{c|cc}
Target & Input & Prediction \\
\includegraphics[align=c,width=0.3\textwidth]{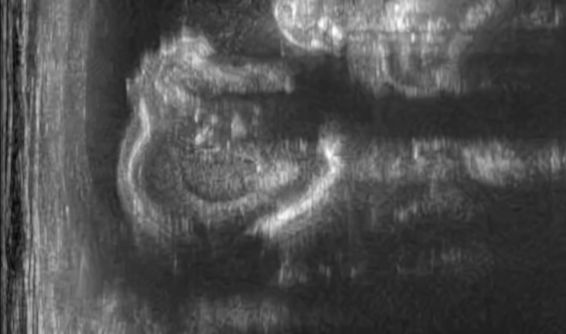} &
\includegraphics[width=0.3\textwidth]{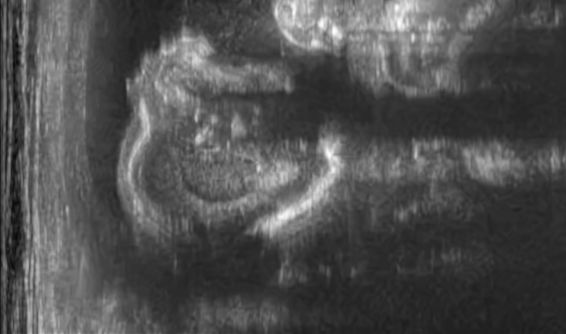} &
\includegraphics[width=0.3\textwidth]{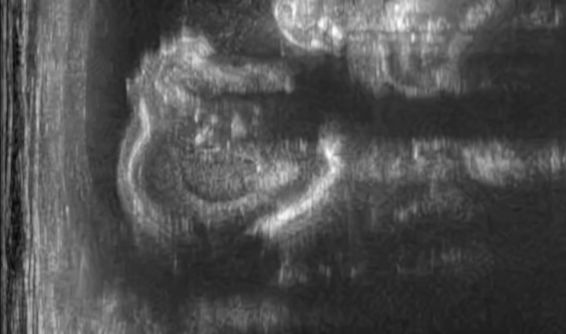} \\ [-13mm]
&
\includegraphics[width=0.3\textwidth]{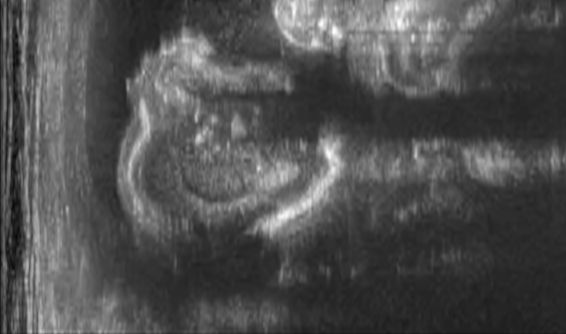} &
\includegraphics[width=0.3\textwidth]{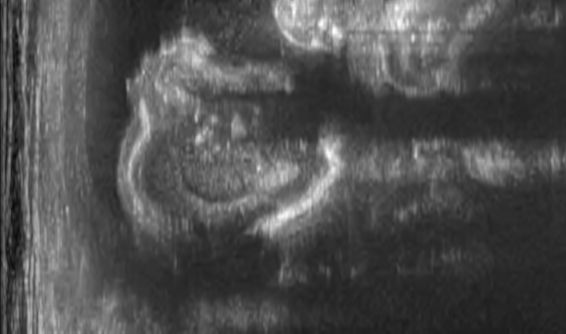} 
\end{tabular}
};
\spy [red] on (-6.3,0.4) in node [left] at (-5.8,0.4); 
\spy [red] on (-0.65,-1.40) in node [left] at (-0.15,-1.40); 
\spy [red] on (4.97,-1.40) in node [left] at (5.47,-1.40); 
\spy [red] on (-0.65,1.85) in node [left] at (-0.15,1.85); 
\spy [red] on (4.97,1.85) in node [left] at (5.47,1.85); 
\end{tikzpicture}
\caption{Prediction on the denoised images of the obstetric district: 2X up-sampling (first line); 4X up-sampling (second line).\label{FIG:OBDENNETWORK}}
\end{figure*}
\subsection{Proposed super-resolution of US videos\label{sec:US-VIDEO-SYPERES}}
Applying our approach to US videos with a low spatial resolution and a high frequency (e.g., for the cardiac district), we can generate high-frequency 2D US video with an increased spatial resolution of each frame, thus overcoming the main limits of current US probes, whose spatial resolution decreases as the acquisition frequency increases. The relationship between image resolution and video frequency~$f$ is given by \mbox{$f=c/(2\cdot d \cdot l)$}, where~$c$ is the speed of sound. The acquisition of low-resolution US images allows the physician to increase the acquisition frequency. The probe acquires a reduced number of lines: we refer to 0.5X and 0.25X low-resolution images, as~$l/2 \times d$ and~$l/4 \times d$ resolution, respectively. We refer the reader to the uploaded video for the experimental tests on the spatial super-resolution of 2D US videos (see URL below). In the video, the input signal is a 2D US video at full resolution~$L \times D \times T$ with~$L$ lines,~$D$ depth, and~$T$ frames. We down-sample each image at~$L/2$ or~$L/4$, and apply our framework for the spatial super-resolution, to reconstruct the full-resolution 2D video. Video URL: \url{https://www.dropbox.com/s/p42pzxxvgf9gacl/SuperResolution-US.mp4?dl=0}.
\begin{figure*}[t]
\centering
\begin{tabular}{cc}
\includegraphics[width=0.45\textwidth]{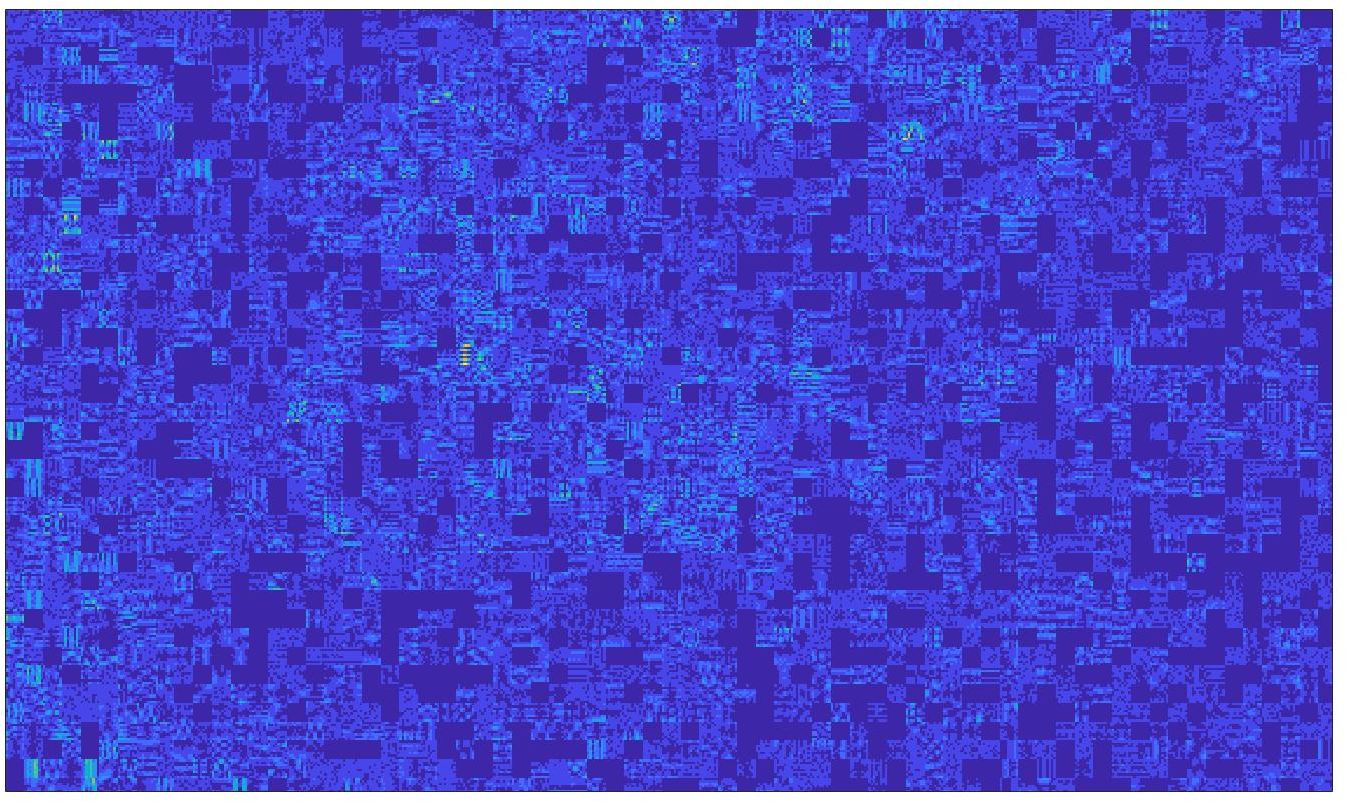} &
\includegraphics[width=0.45\textwidth]{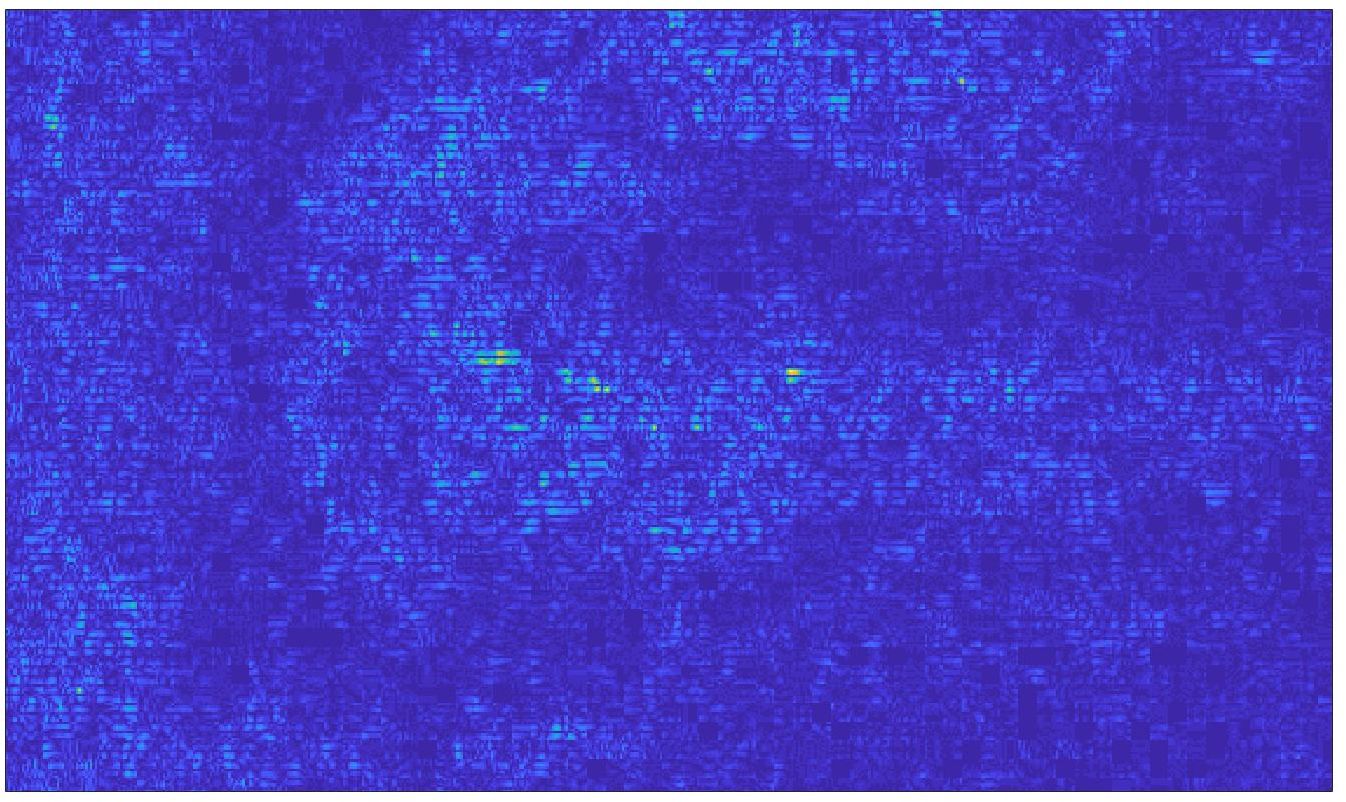} \\
Max. error: 6 & Max. error: 20
\end{tabular}
\caption{Concerning Fig.~\ref{FIG:OBDENNETWORK}, we show the error image of our method with respect to the target image with both 2X and 4X up-sampling factors on obstetric denoised images. For each image, we report the maximum error in the scale~$0-255$.\label{FIG:ERRORIMAGEDEN}}
\end{figure*}
\subsection{Denoising and super-resolution\label{SEC:DENOISING}}
To evaluate the effect of denoising for the super-resolution of US images, we apply to input raw images a learning-based low-rank denoising which allows us to select a soft intensity of the smoothing. This approach generates denoised images that are visually similar to raw images, and simultaneously more uniform. Then, we generate down-sampled images (0.5X and 0.25X) and apply the \emph{Cubic convolution} up-sampling. These couples of images (i.e., denoised at full resolution and up-sampled) are used to train the learning-based network~(Sect.~\ref{SEC:METHOD}). With this approach, we verify the performance of both the up-sampling algorithm and our learning-based prediction when applied to input denoised images.
\begin{figure*}[t]
\centering
\begin{tabular}{cc|cc|cc}
\multicolumn{2}{c|}{(a) Obstetric district}
&\multicolumn{2}{c|}{(b) Cardiac district}
&\multicolumn{2}{c}{(c) Abdominal district}\\
\hline
\includegraphics[width=0.28\columnwidth]{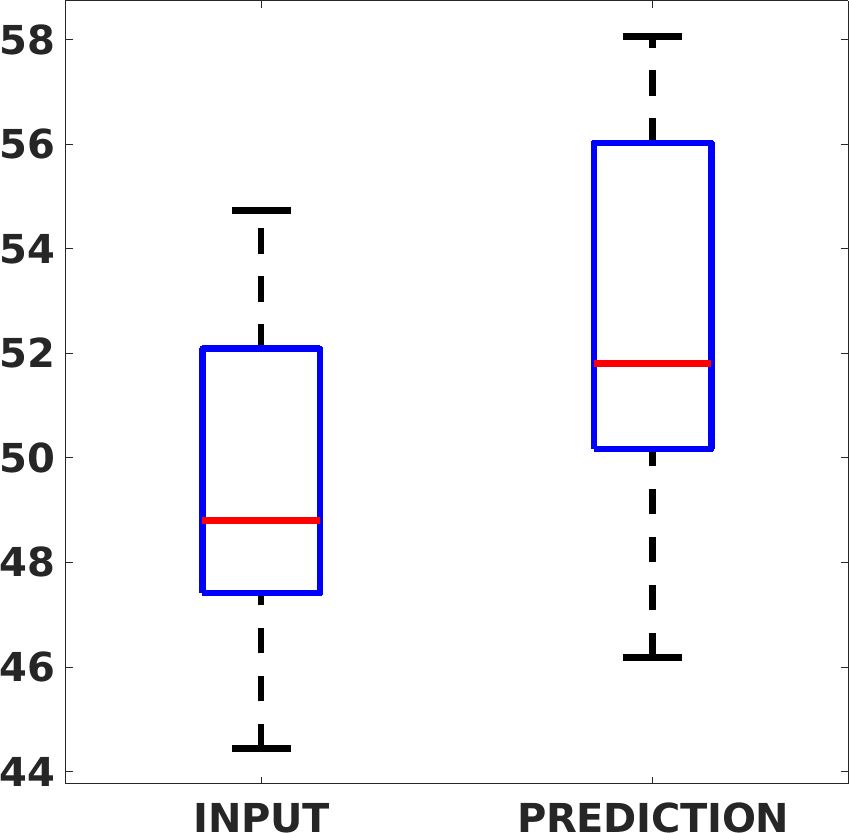} &
\includegraphics[width=0.28\columnwidth]{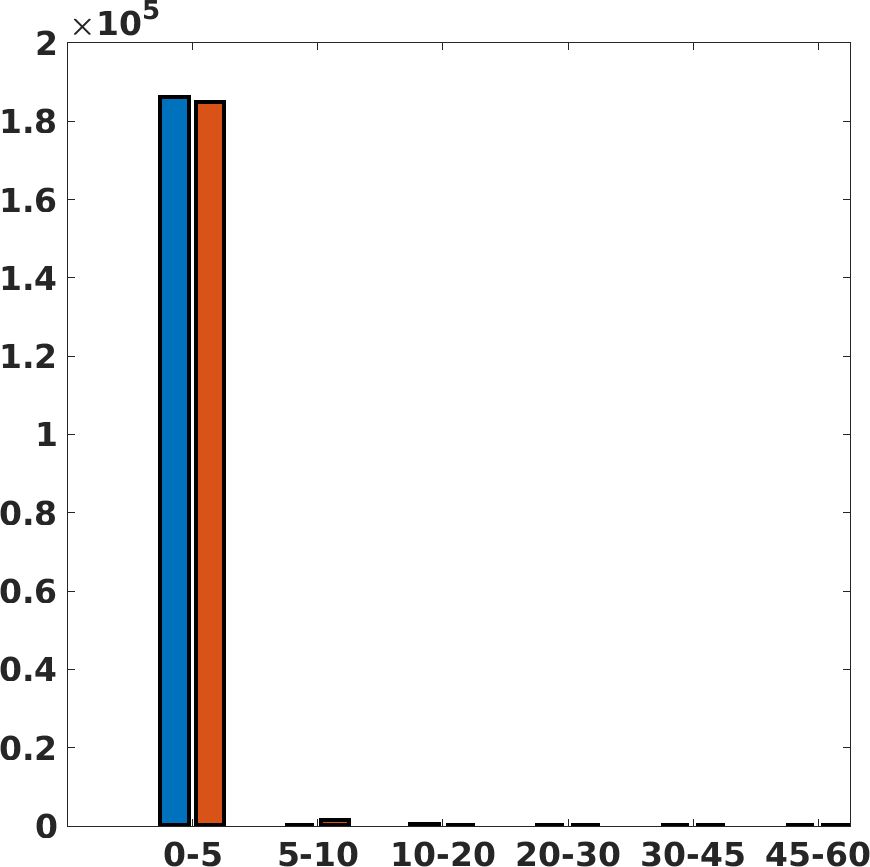} &
\includegraphics[width=0.28\columnwidth]{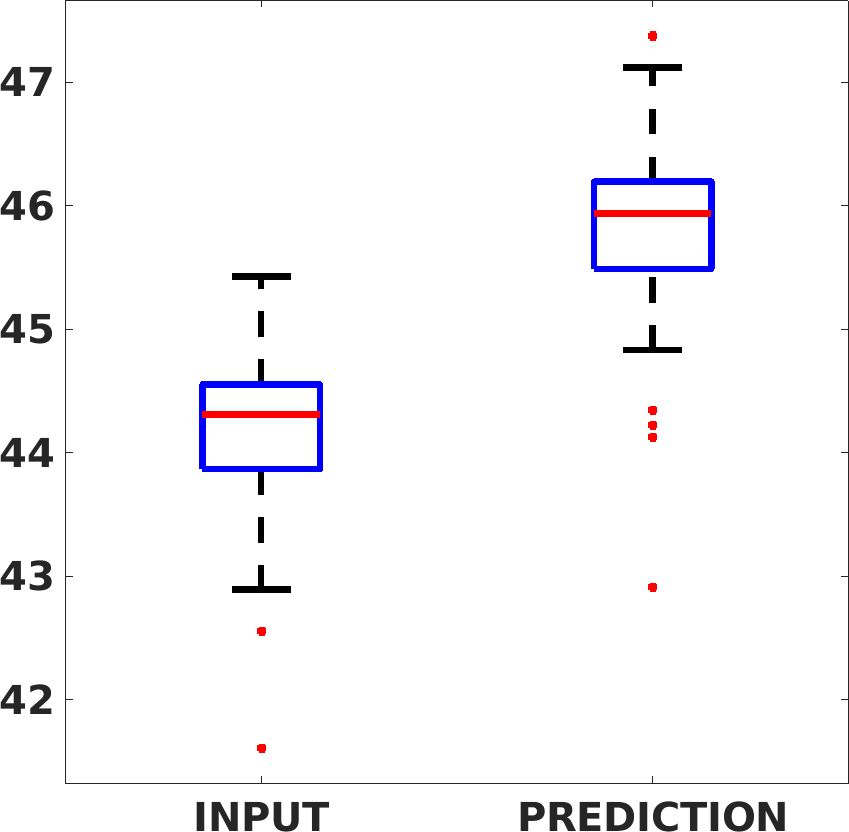} &
\includegraphics[width=0.28\columnwidth]{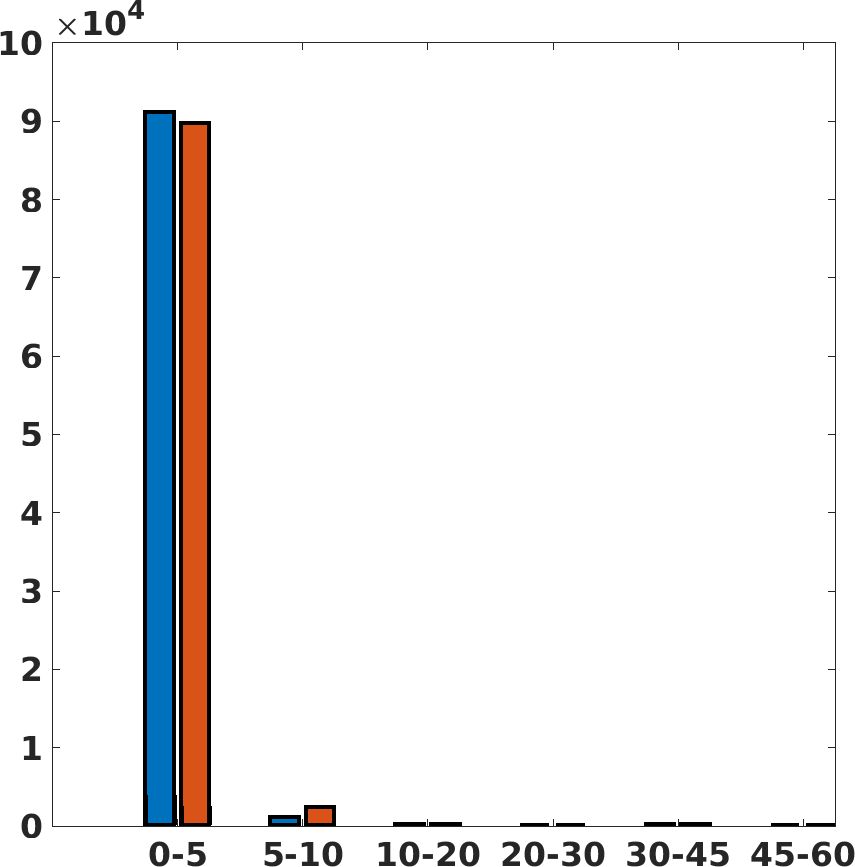} &
\includegraphics[width=0.28\columnwidth]{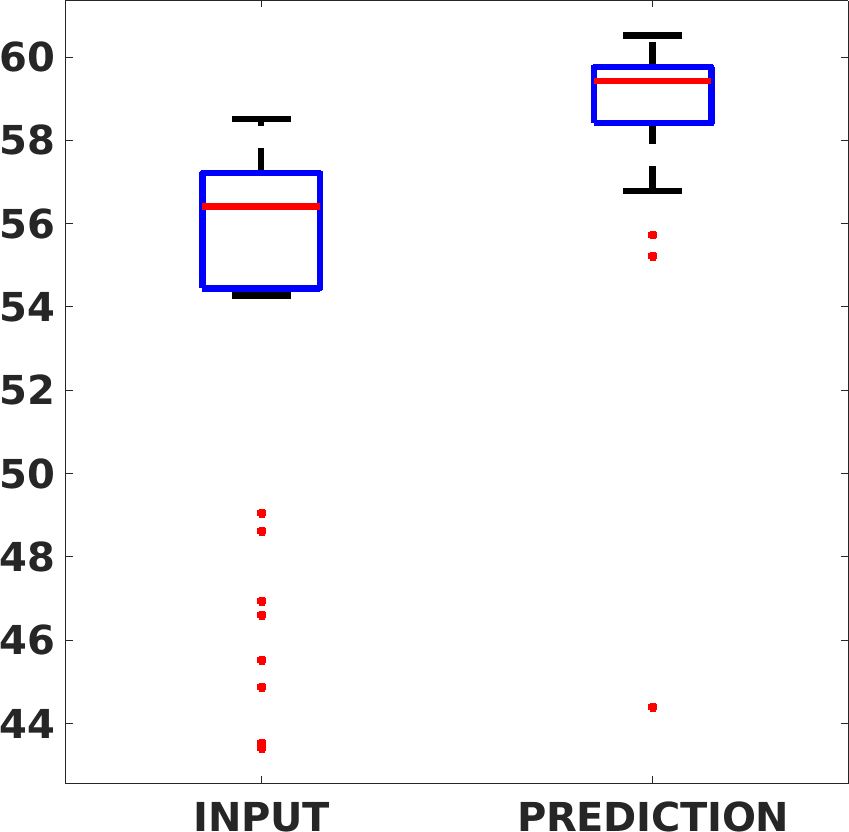} &
\includegraphics[width=0.28\columnwidth]{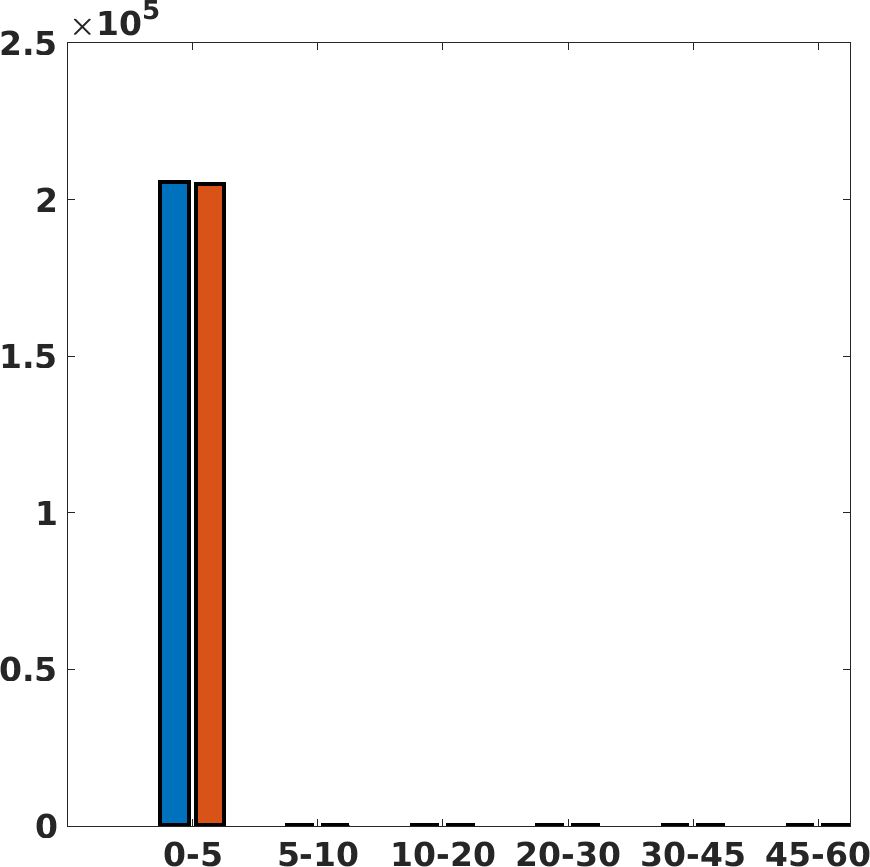} \\ 
\includegraphics[width=0.28\columnwidth]{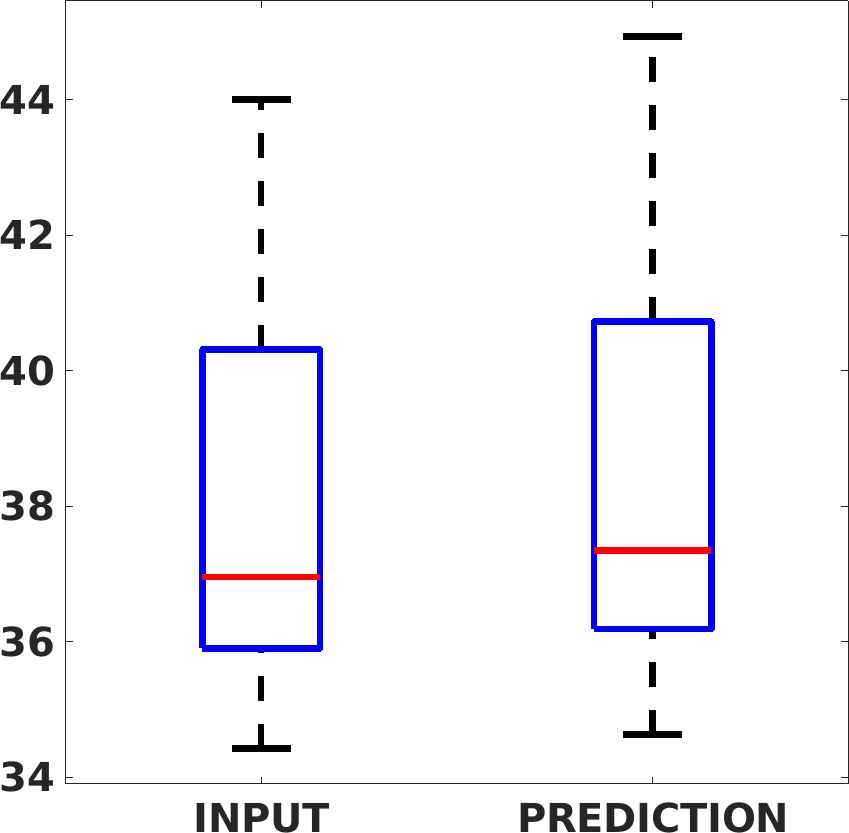} &
\includegraphics[width=0.28\columnwidth]{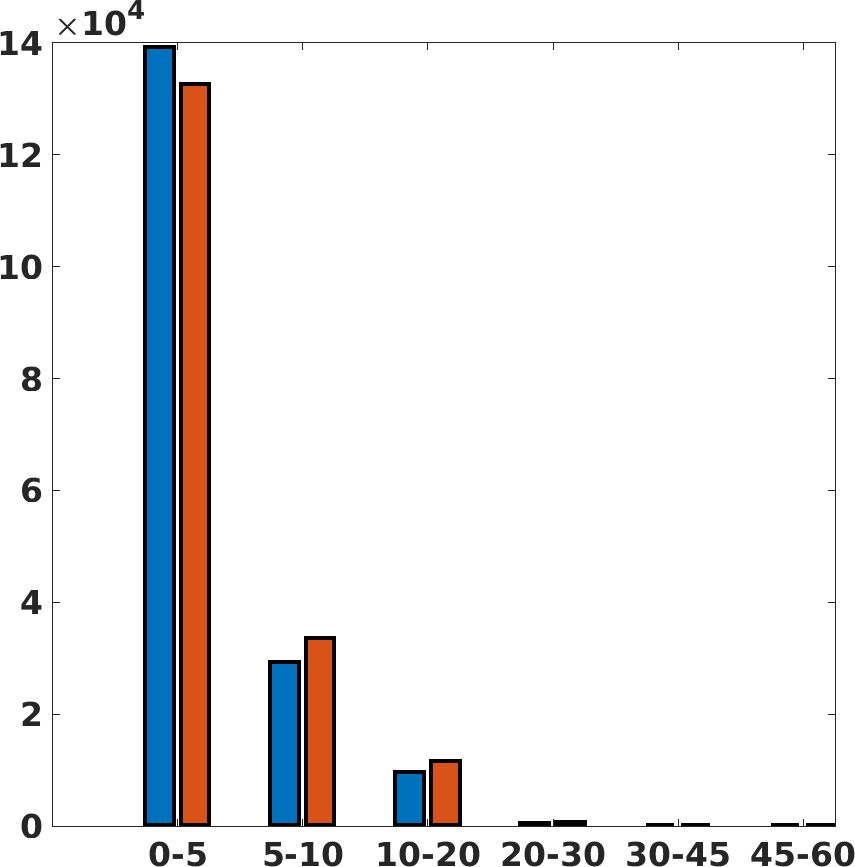} &
\includegraphics[width=0.28\columnwidth]{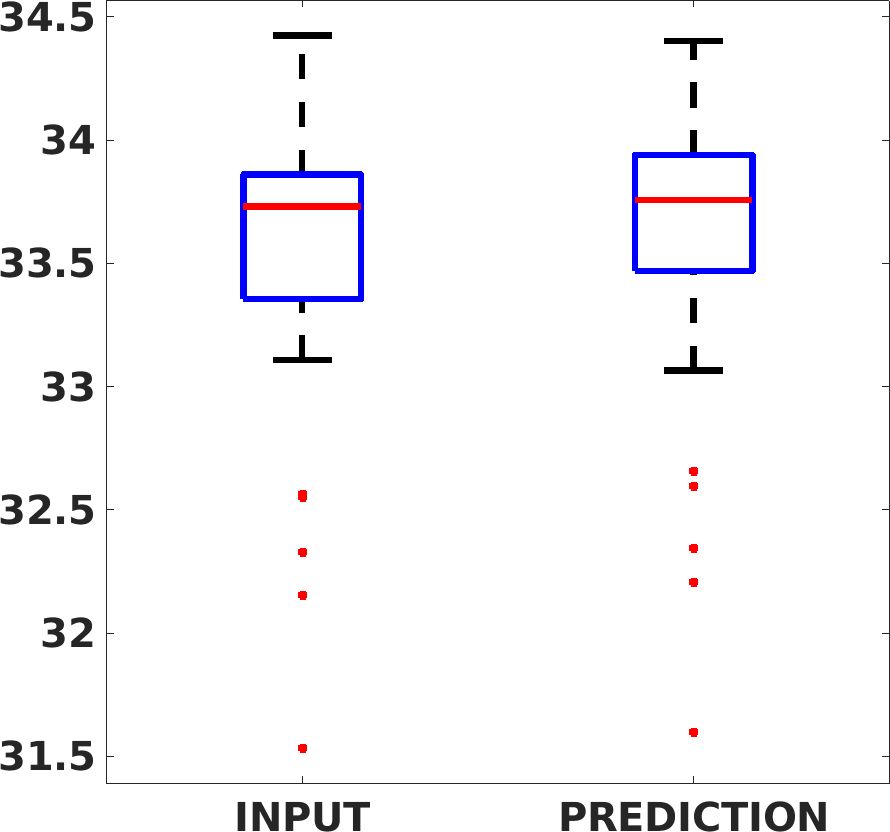} &
\includegraphics[width=0.28\columnwidth]{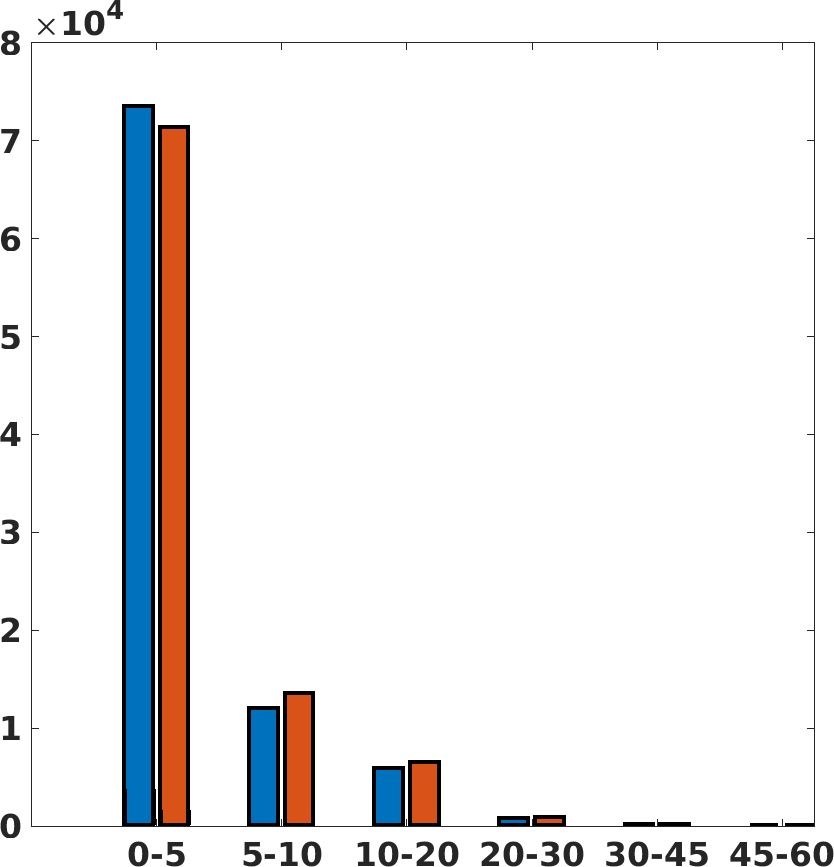} &
\includegraphics[width=0.28\columnwidth]{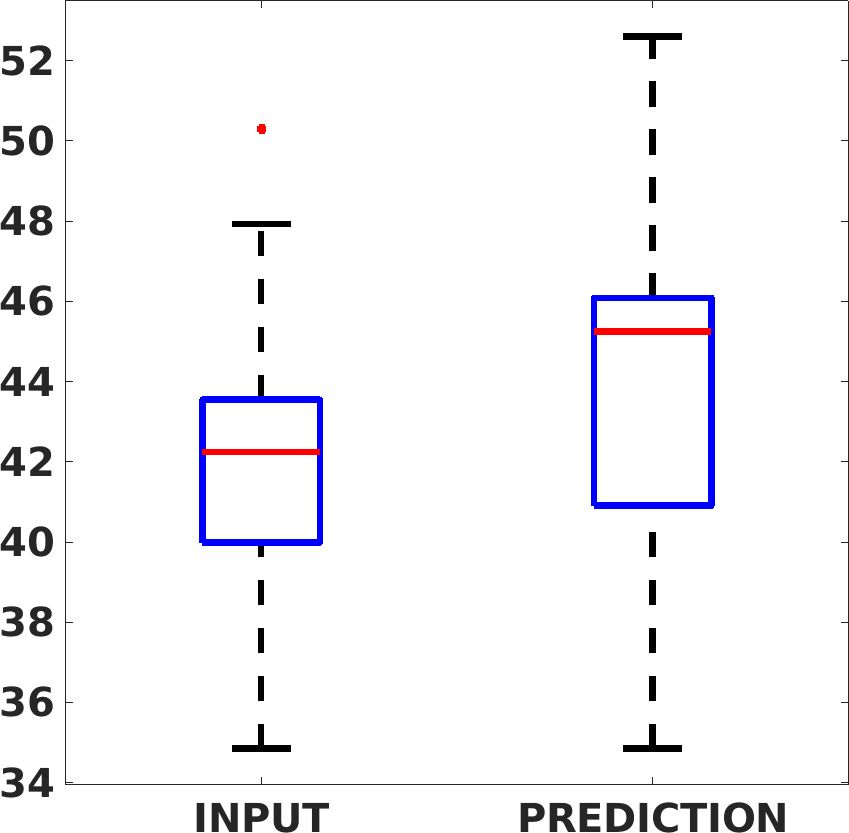} &
\includegraphics[width=0.28\columnwidth]{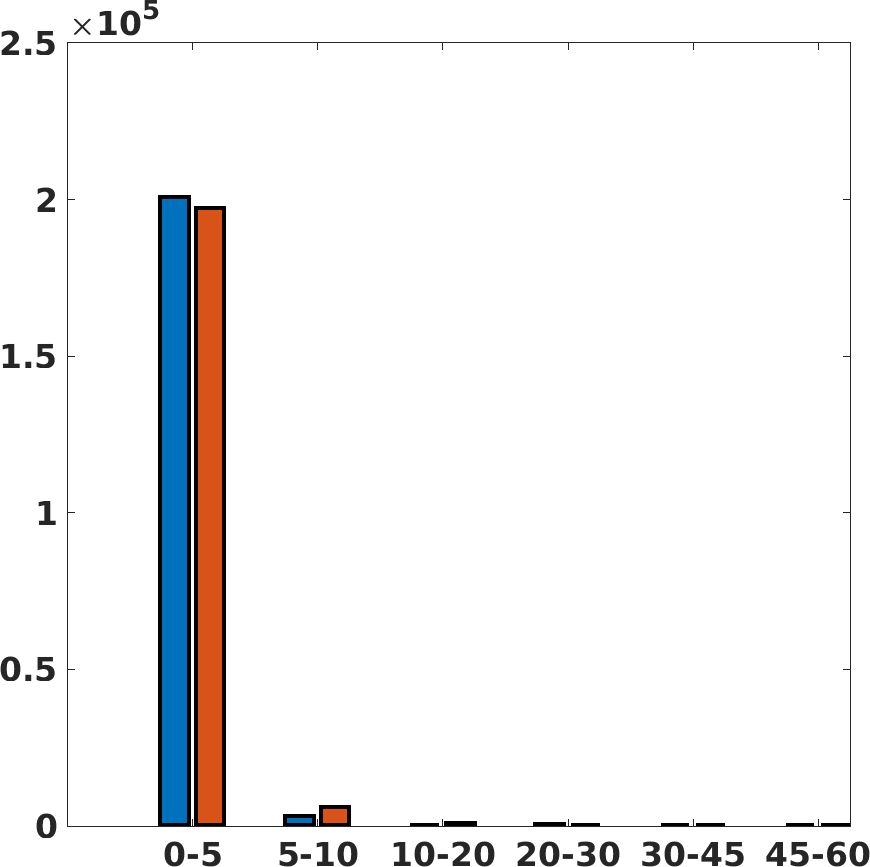} 
\end{tabular}
\caption{PSNR box-plot (left) with denoised images of the (a) obstetric, (b) cardiac, and (c) abdominal districts, and error histogram (right): prediction (blue) vs. input (red): 2X (first line) and 4X (second line) results.\label{FIG:OBDENQUANTITATIVE}}
\end{figure*}

Fig.~\ref{FIG:OBDENNETWORK} shows the results of the prediction of the network, compared with the input and the target denoised images of the obstetric district. Our framework visually improves the results, in terms of blurring and artefacts. Fig.~\ref{FIG:ERRORIMAGEDEN} shows the error image of our prediction with respect to the target denoised image, for both 2X and 4X up-sampling. The error is mainly distributed on the edges of the anatomical structure. Furthermore, the maximum error of the 2X up-sampling is 6 in the range of~$0-255$, showing us that our method accurately predicts the target if soft denoising is applied before up-sampling.

Fig.~\ref{FIG:OBDENQUANTITATIVE} (left) shows the box plot of the quantitative metrics, comparing the target images with the prediction and the \emph{Cubic convolution}, respectively. The PSNR metric is computed on a data set of 200 images, belonging to the same district, and with the same up-sampling factor. Analysing the obstetric anatomical district and concerning the corresponding raw images (Fig.~\ref{FIG:OBQUANTITATIVE} (a, left)), the denoising allows the network to significantly improve the results of the up-sampling and the prediction. In particular, comparing the target images with the predicted images, the median PSNR value of obstetric 2X denoised images is 51.8, compared to the median PSNR value of obstetric 2X raw images which is 36.9.
\begin{figure*}[t]
\centering
\begin{tabular}{cc|cc|cc}
\multicolumn{2}{c|}{(a) Obstetric district}
&\multicolumn{2}{c|}{(b) Cardiac district}
&\multicolumn{2}{c}{(c) Abdominal district}\\
\hline
\includegraphics[width=0.28\columnwidth]{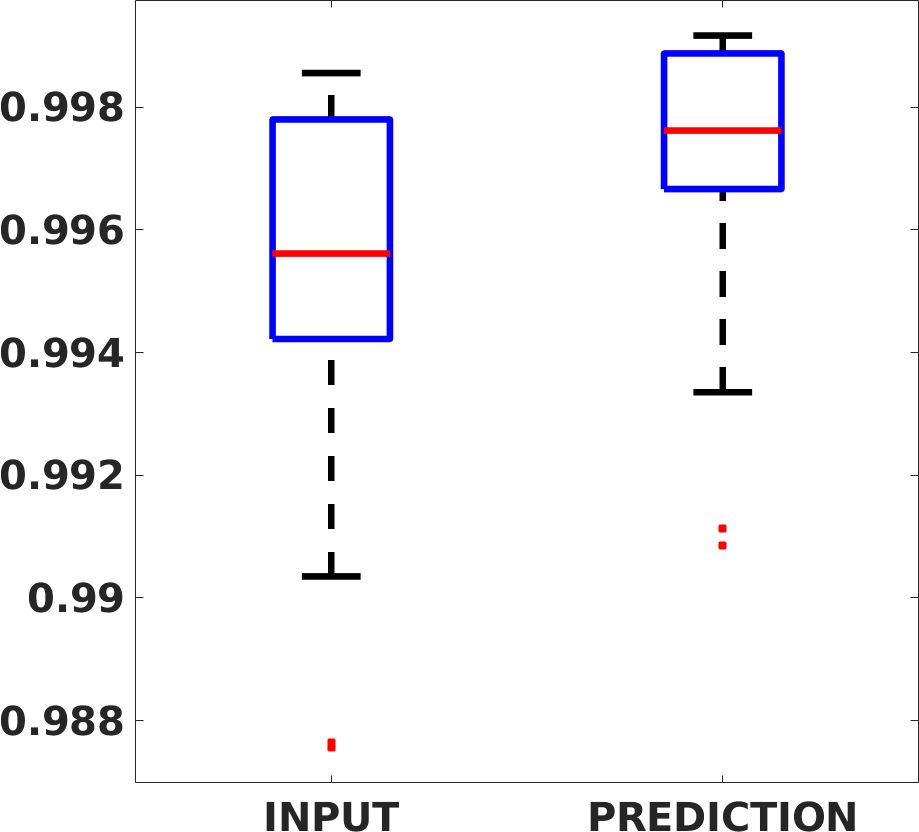} &
\includegraphics[width=0.28\columnwidth]{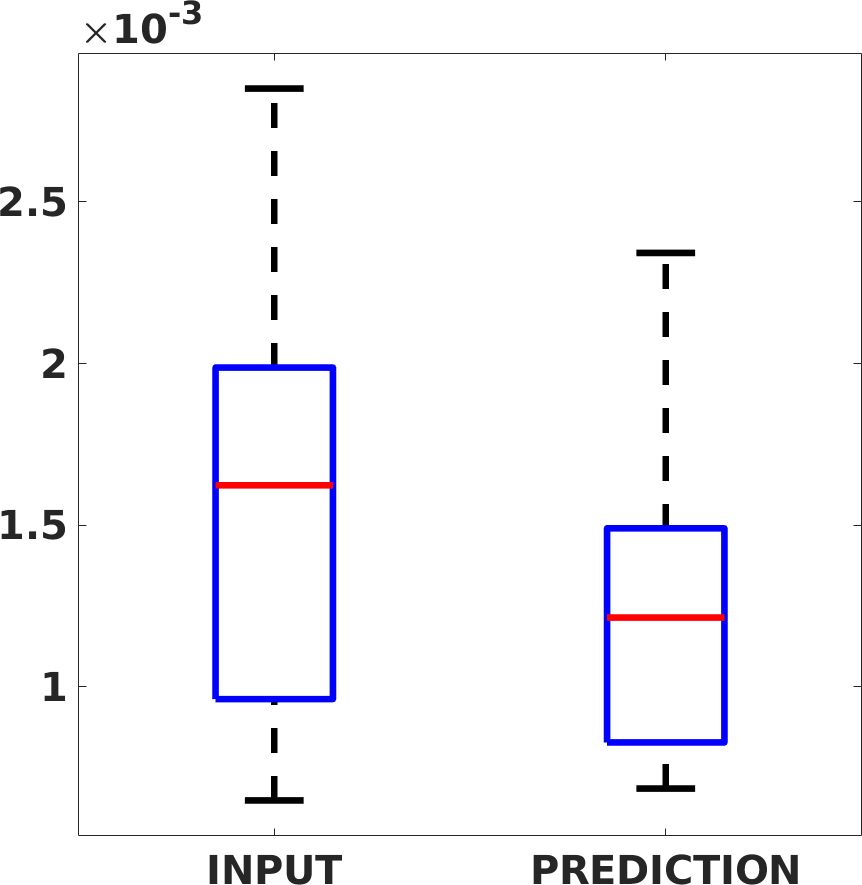} &
\includegraphics[width=0.28\columnwidth]{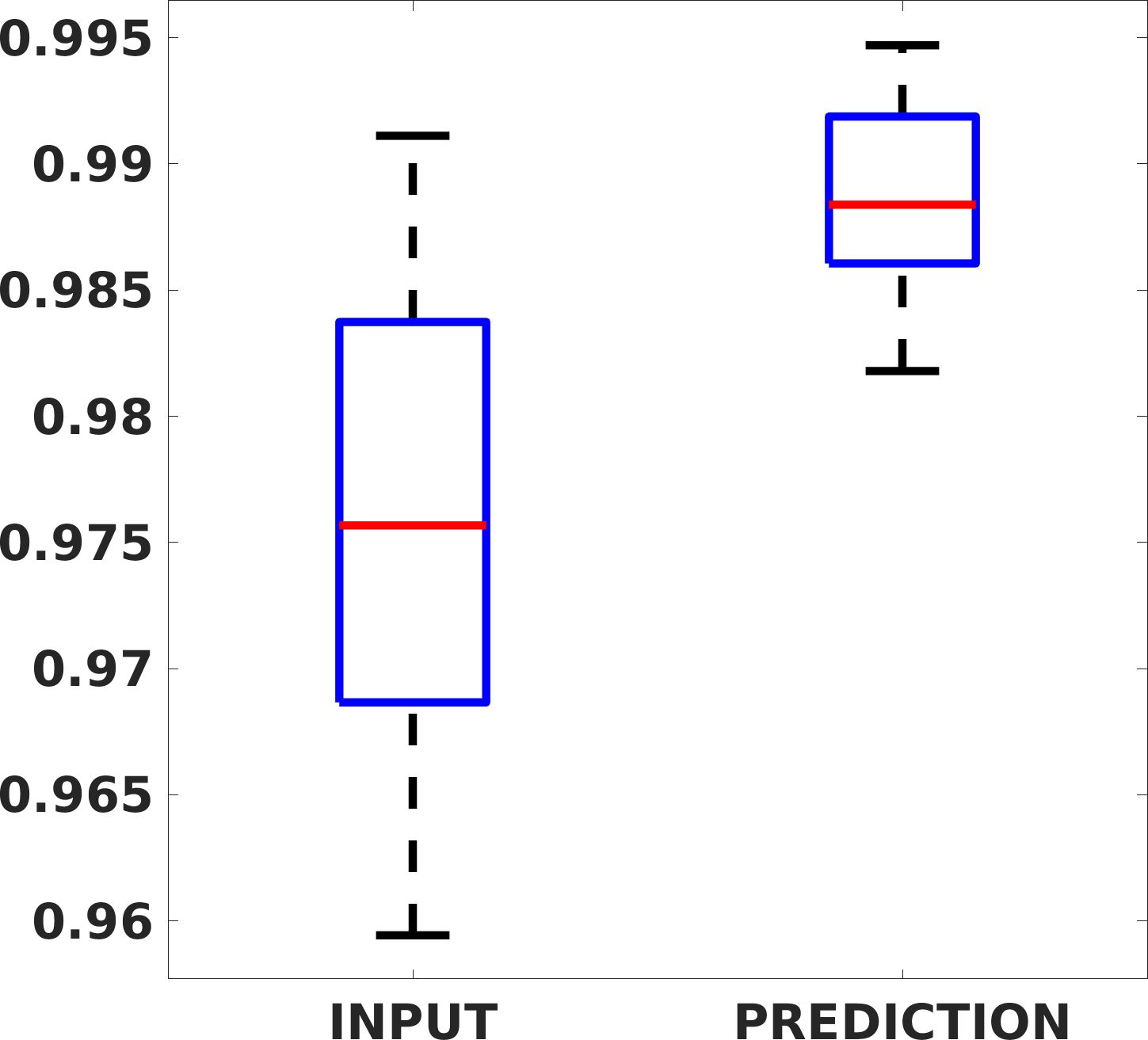} &
\includegraphics[width=0.28\columnwidth]{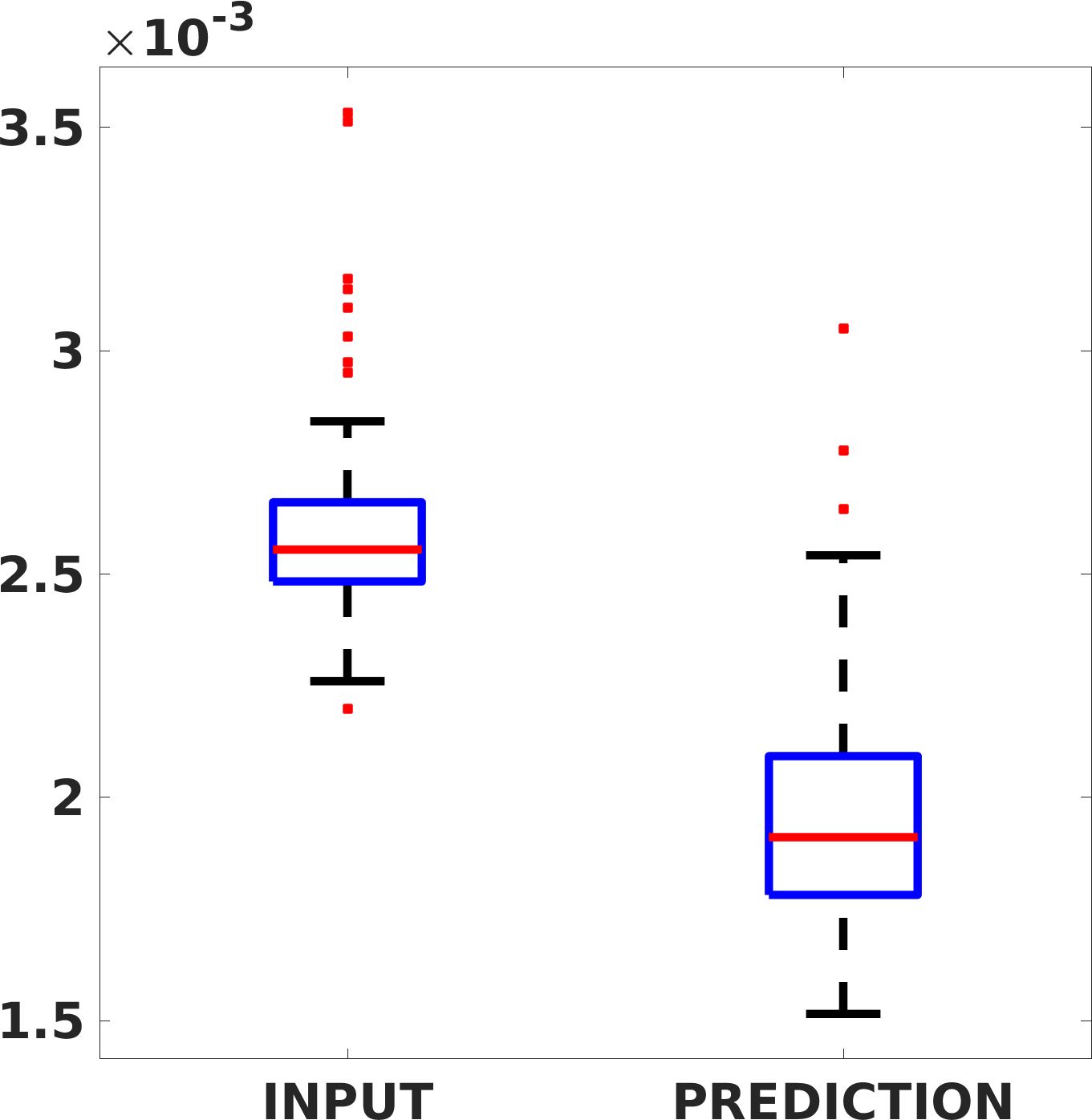} &
\includegraphics[width=0.28\columnwidth]{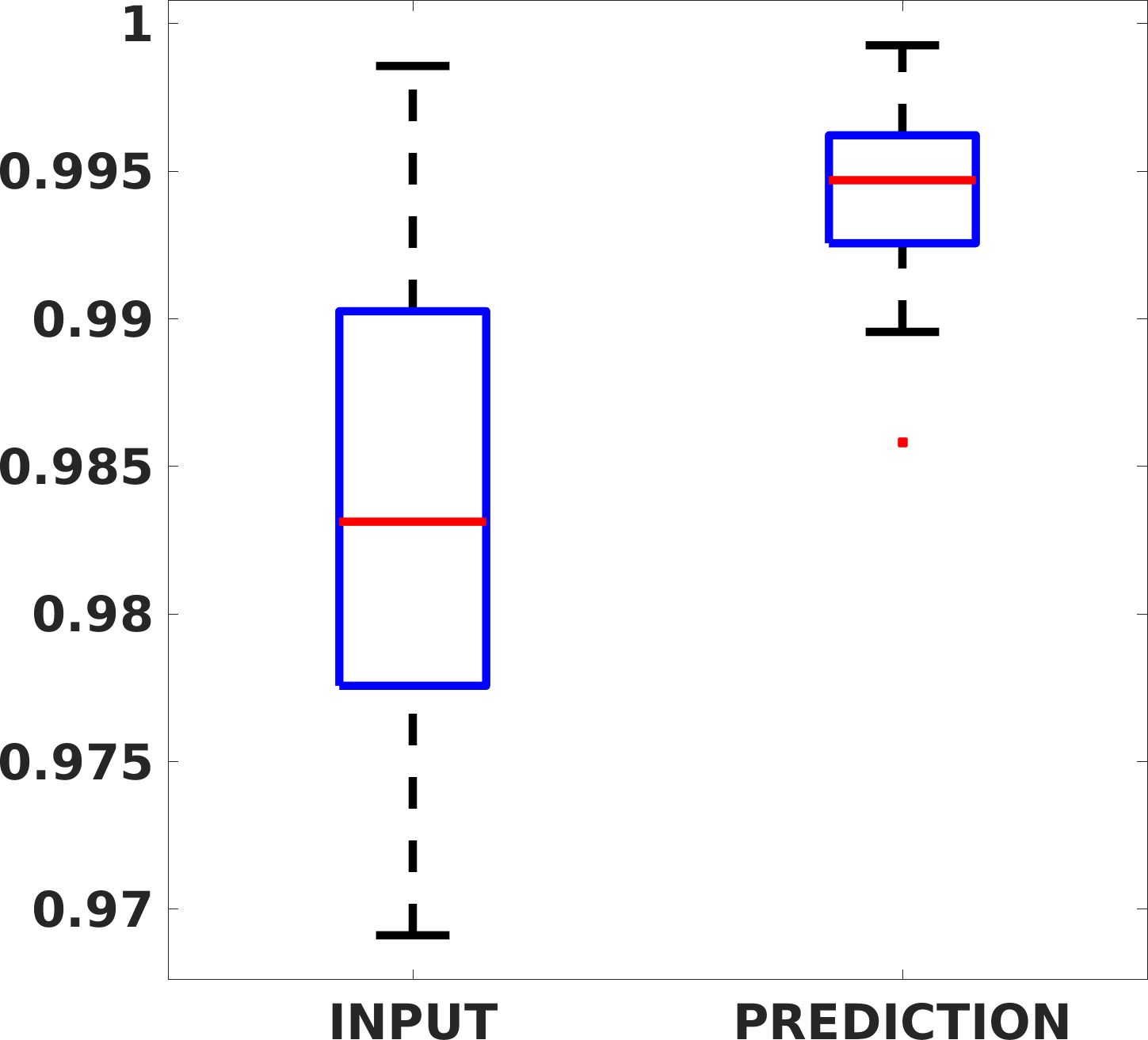} &
\includegraphics[width=0.28\columnwidth]{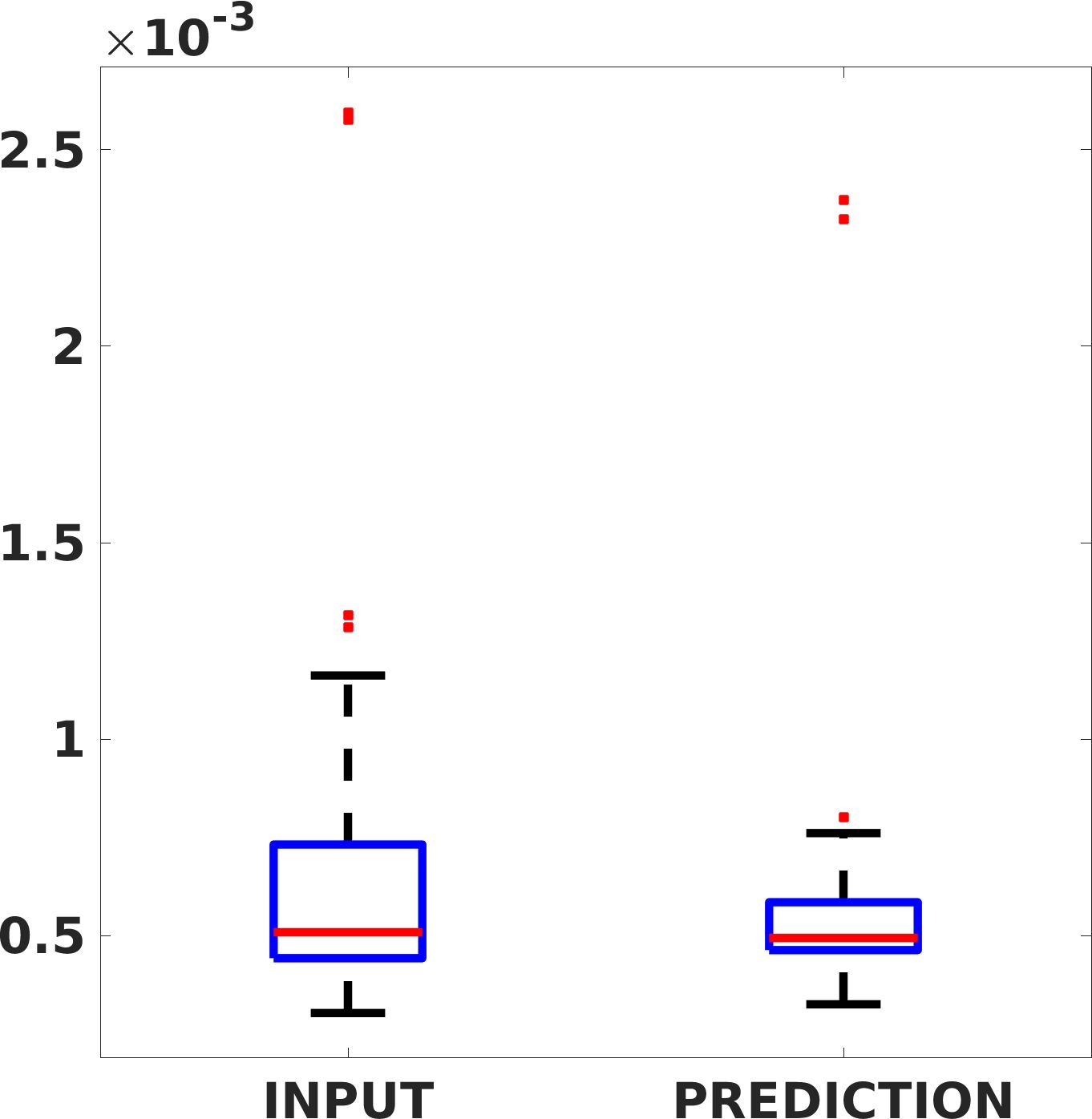} \\ 
\includegraphics[width=0.28\columnwidth]{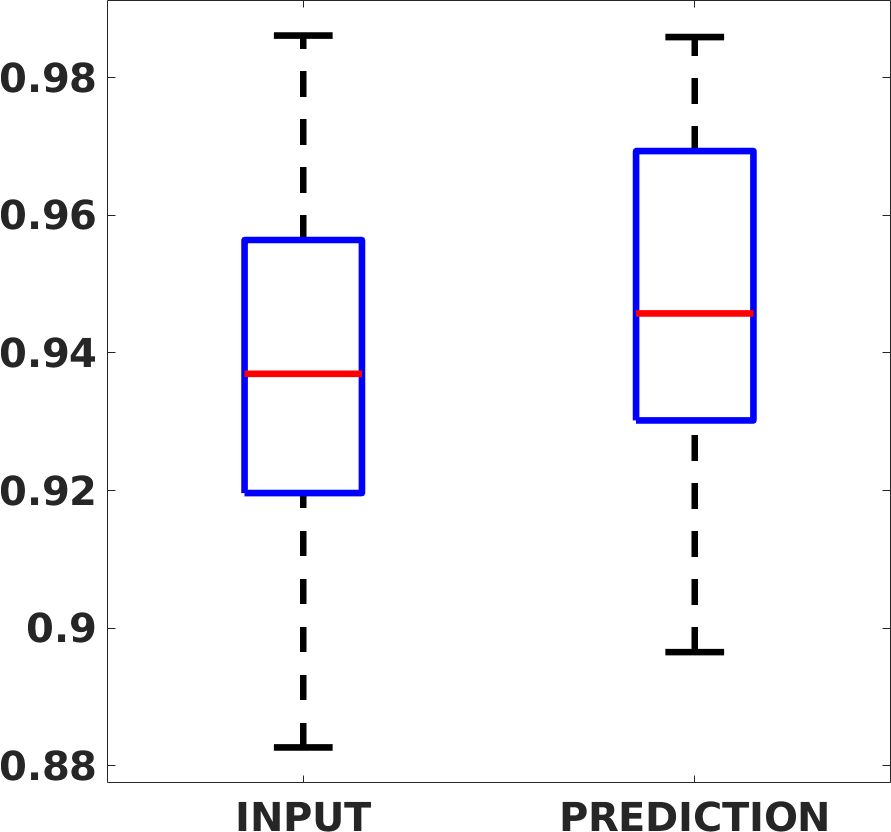} &
\includegraphics[width=0.28\columnwidth]{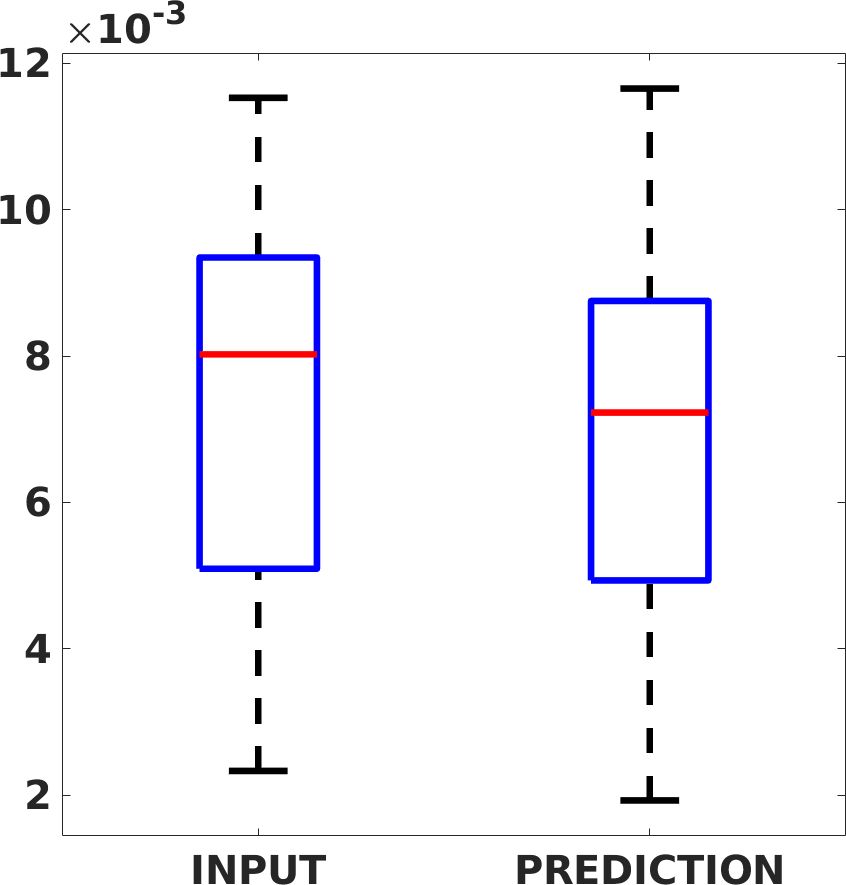} &
\includegraphics[width=0.28\columnwidth]{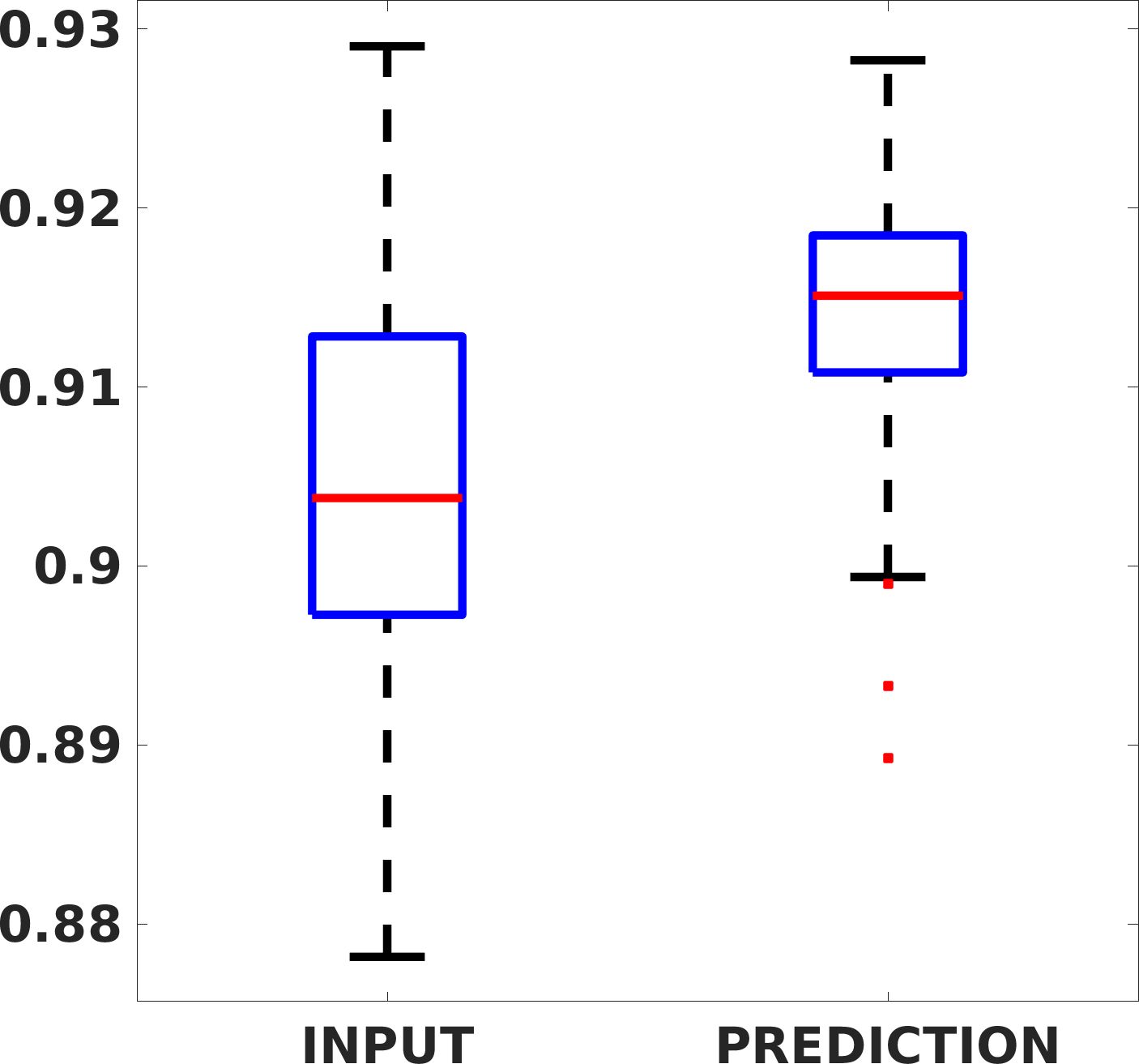} &
\includegraphics[width=0.28\columnwidth]{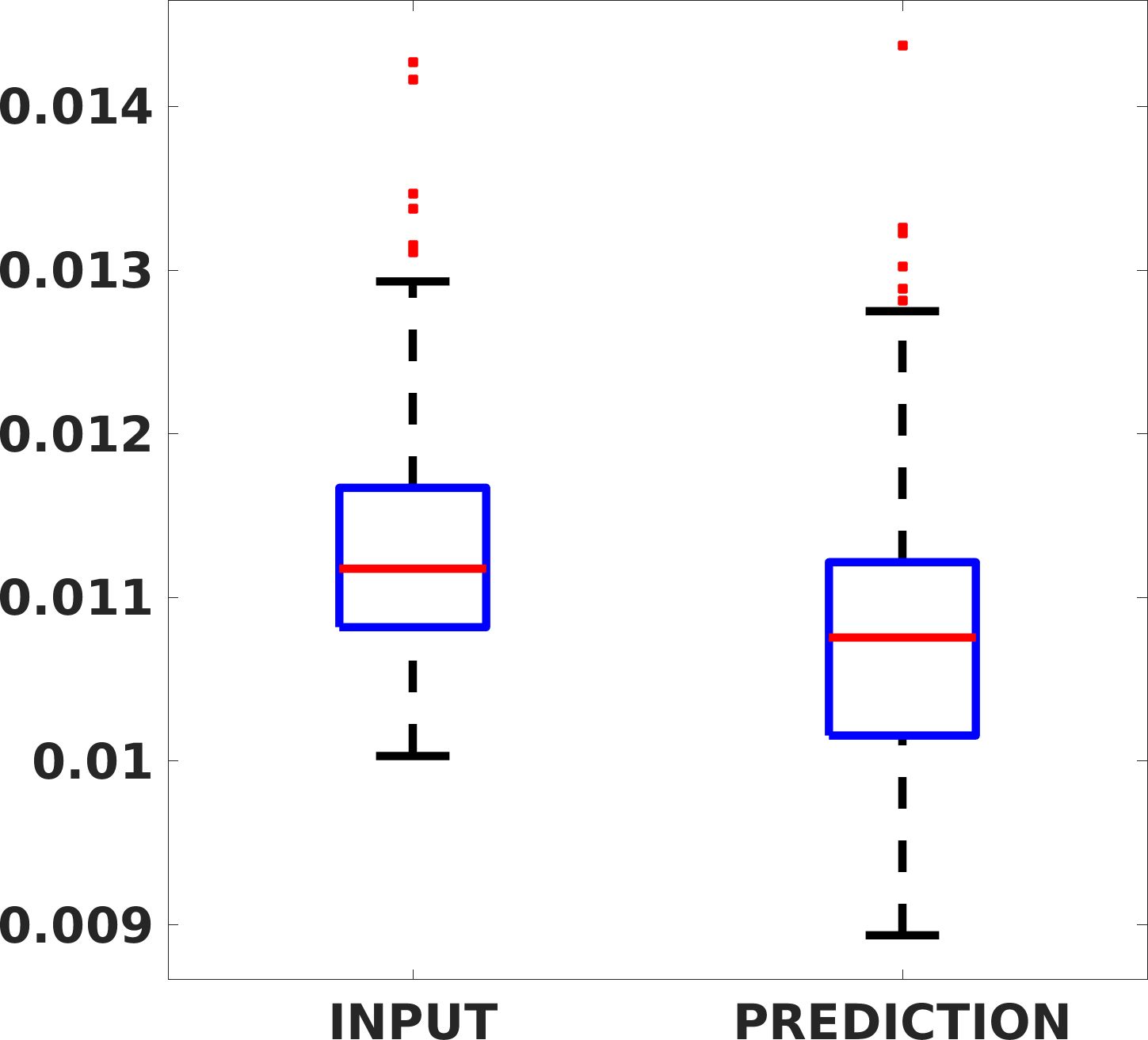} &
\includegraphics[width=0.28\columnwidth]{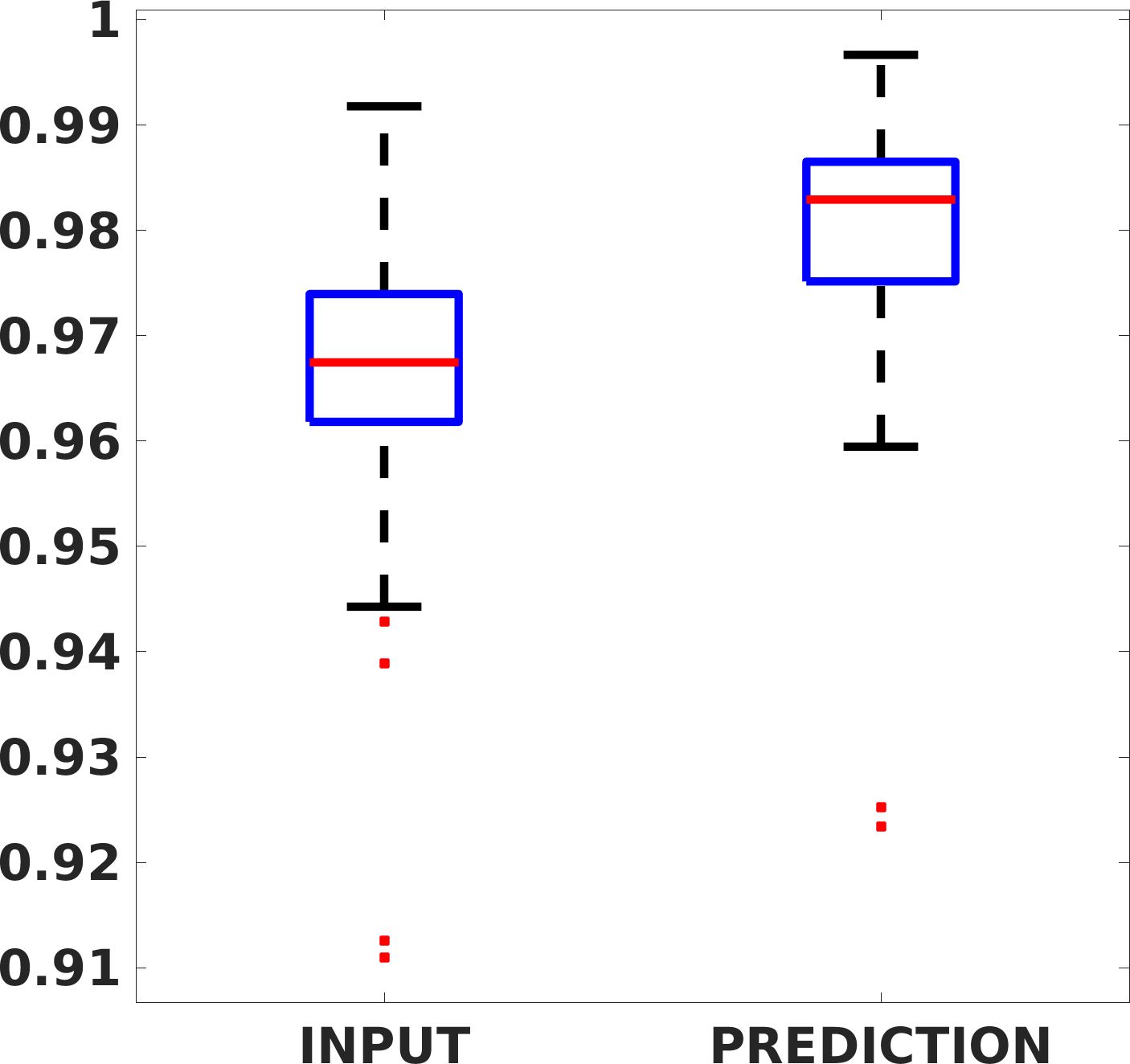} &
\includegraphics[width=0.28\columnwidth]{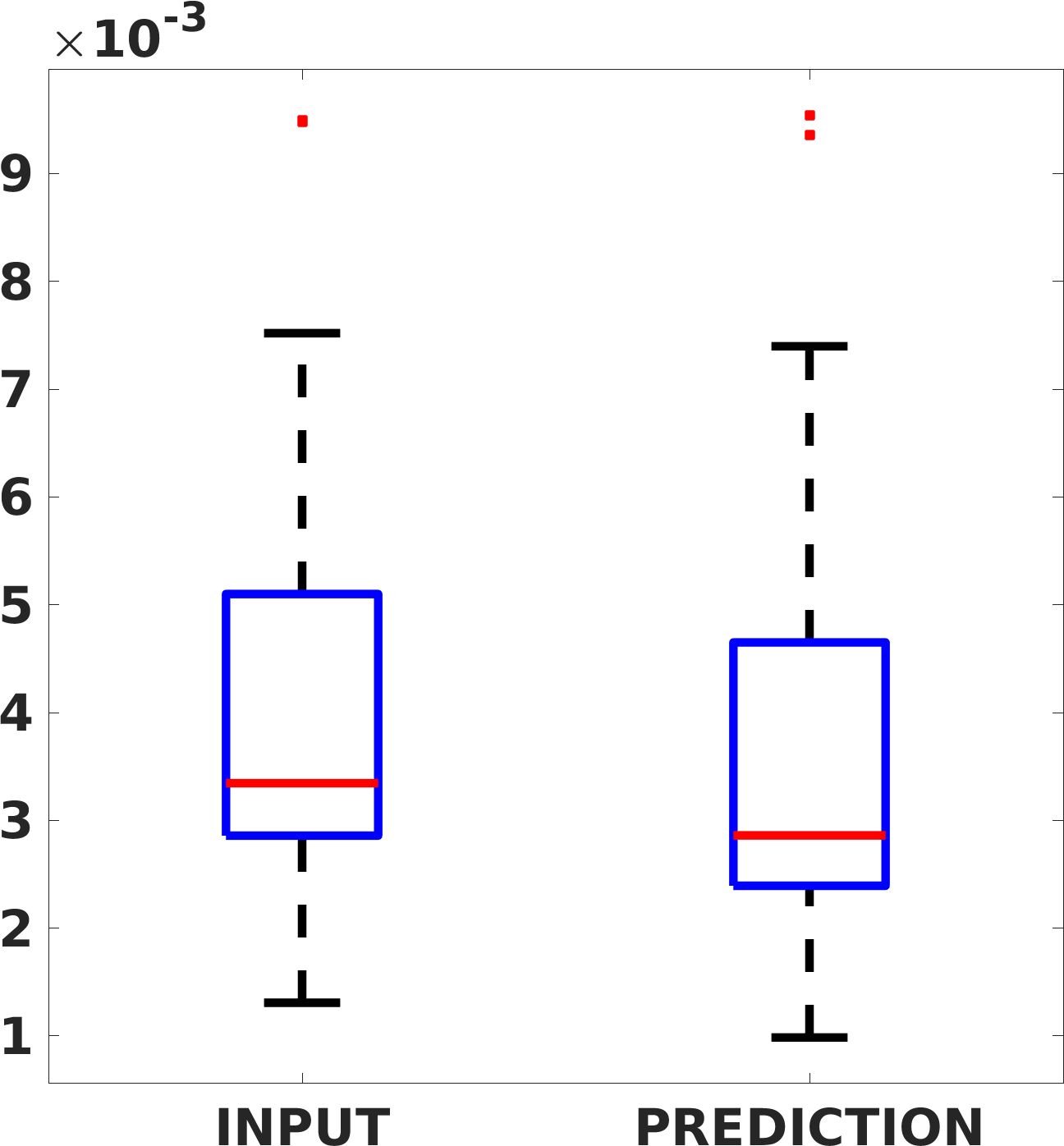} 
\end{tabular}
\caption{SSIM box-plot (left) and MAE box-plot (right) with denoised images of the (a) obstetric, (b) cardiac, and (c) abdominal districts:  2X (first line) and 4X (second line) results.\label{FIG:OBDENQUANTITATIVE2}}
\end{figure*}

Fig.~\ref{FIG:OBDENQUANTITATIVE} (right) shows the histogram of the absolute value of the error with respect to the target, of the prediction and \emph{Cubic convolution} respectively. This result shows that our framework increase of~$1.7\%$ and~$14\%$ (2X and 4X, respectively) the number of pixels where the prediction error is lower than 5, which is very similar to the target when visually analysing the images, and improved with respect to the learning framework applied to raw images. According to Fig.~\ref{FIG:OBDENQUANTITATIVE2}, our method improves the accuracy of \emph{Cubic convolution}. For example, the SSIM increases of~$1.3\%$ on cardiac 2X and the MAE increases of~$8.2\%$ on abdominal 4X.

\subsection{Execution time and computational cost\label{SEC:EXTIME}}
\begin{figure*}[t]
\centering
\begin{tabular}{cc}
\includegraphics[width=0.45\textwidth]{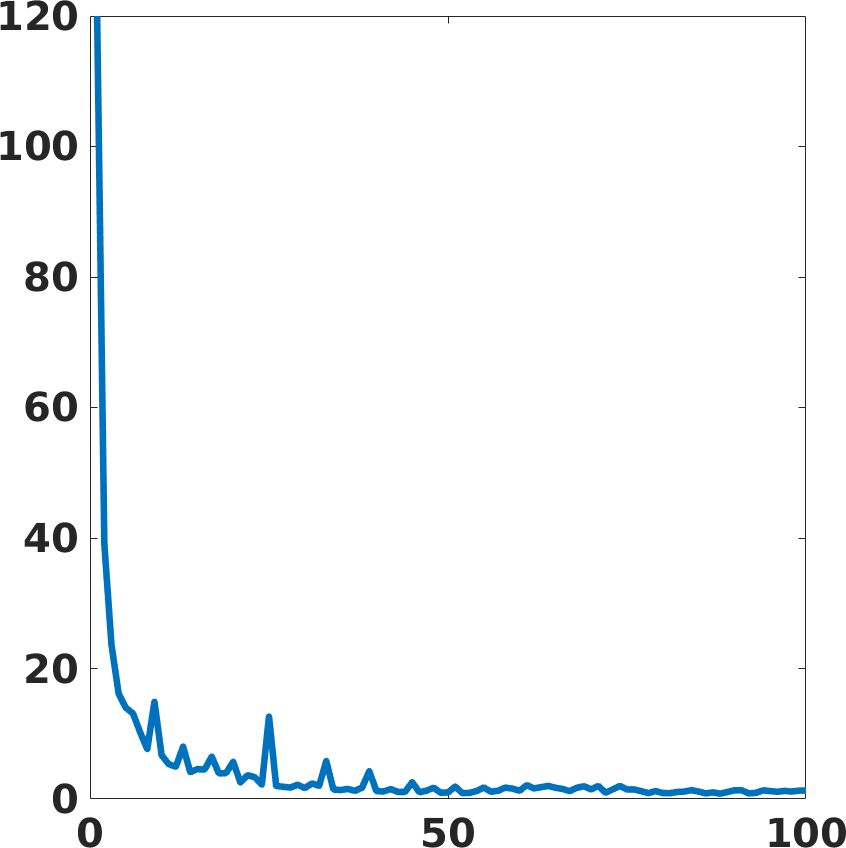} &
\includegraphics[width=0.45\textwidth]{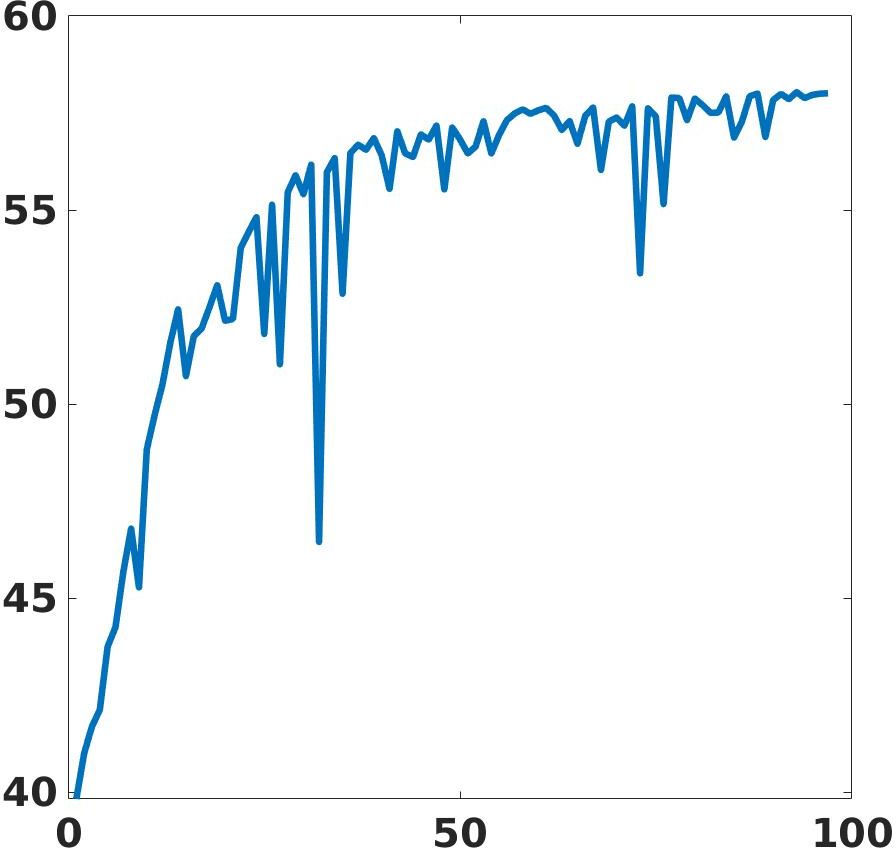} 
\end{tabular}
\caption{Training (left) and validation (right) loss ($y-$axis) with respect to the number of epochs ($x-$axis).\label{FIG:LOSS}}
\end{figure*}

We define an HPC implementation of the proposed framework on the CINECA-Marconi100 cluster, exploiting both CPUs (IBM POWER9 AC922) and GPUs (NVIDIA Volta V100). We design a parallel and distributed implementation in TensorFlow2, and we train multiple networks with large data sets for the target medical application. To test the training phase of the learning-based networks in the HPC environment, we exploit 8 nodes, each one composed of 32 cores and 4 accelerators, for a theoretical computational performance of 260 TFLOPS, and 220 GB of memory per node. The parallel implementation of the deep learning framework and the high hardware performance reduce the computation time of the training phase by at least 100 orders less than a serial implementation on a standard workstation. Fig.~\ref{FIG:LOSS} shows the training loss and validation PSNR. Both metrics show convergence property within 100 epochs iteration. In particular, the validation PSNR goes from a value of 41 to a value of 58 after 100 epochs.

The computational cost of the prediction depends on the resolution of the input image and the architecture of the network: in particular, the computational cost of a convolution operation is~$\mathcal{O} (r/s_r \cdot c/s_c)\cdot(f_r \cdot f_c)\cdot f$; in our application, the input images have variable resolutions, with a maximum value of~$r=c=600$, the kernel-filter size on rows and columns is~$f_r = f_c = 3$ on 2X applications and~$f_r = f_c = 5$ on 4X applications, the stride on rows and columns is~$s_r = s_c = 1$, we use 16 convolution operators and 10 kernel filters.

We test the prediction on GPU-based hardware, which replicates the hardware of a US scanner currently in use. Given a set of US input images from different districts at different resolutions, the average execution time is 8 milliseconds. Finally, the denoise pre-processing can be performed in real-time through a learning-based method~\cite{cammarasana2022real}.

\section{Discussion\label{SEC:DISCUSSION}}
SOTA super-resolution methods approximate the unknown values with deep-learning models that take advantage of large data sets or interpolating models that account for the neighbouring points through kernel functions. Learning-based models tend to generate artefacts while interpolating algorithms are general-purpose models that may be less accurate on anatomical districts with complex geometries and may not be robust to noise images. Our method combines the two approaches: first, we up-sample the low-resolution image through an interpolating method; then, we apply a learning-based network to improve the visual accuracy of the up-sampling on the specific anatomical district without generating artefacts and improving super-resolution results in comparison with the SOTA methods. Furthermore, neural network models such as WDSR can be used in super-resolution problems not only for interpolation but also for the specialisation and fine-tuning of the results. As the main requirement for our two-steps approach, the up-sampling algorithm (i.e., the first step) must have a low execution time, to keep the entire pipeline in real-time.

Through learning and high-performance computing, the proposed super-resolution is specialised to different anatomical districts by training multiple networks. Furthermore, we can improve the offline training with new data, a-priori and/or additional information on the input data (e.g., anatomical district, image resolution, acquisition methodology/protocol). The training data set can be periodically updated with the up-sampled images after the expert validation of the super-resolution results or with new data to further specialise in the individual networks. The use of deep learning on large data sets overcomes the limitations of vision-based algorithms that are general and do not fully encode the characteristics of the data. Each network is separately trained from scratch on each anatomical district and up-sampling factor. If a small data set is available for a certain anatomical district, we can train a general-purpose network and then specialise dedicated networks with a fine-tuning stage to the specific anatomical districts.

HPC is widespread for the training of learning models in US processing; for example, for the localisation of common carotid artery transverse section through RCNN~\cite{jain2020localization}, automatic segmentation of the carotid artery and internal jugular veins~\cite{groves2020automatic}, fetal standard planes recognition~\cite{pu2021automatic}, and segmentation and classification of anatomical structures~\cite{patel2019performance}. HPC and cloud computing also poses new challenges in terms of reorganisation of the medical analysis pipeline, where the computational demand is shifted to centralised hardware resources with a real-time execution of the network’s prediction on local devices~\cite{caballero2019deep}.

\section{Conclusions\label{SEC:CONCLUSIONS}}
We introduce a novel deep learning framework for the super-resolution of US images, which improves the quality of the up-sampling of a selected state-of-the-art algorithm, by training a neural network to match the target high-resolution image. Our method is tested on different anatomical districts (e.g., obstetric, cardiac, obstetric) and up-sampling factors (e.g., 2X, 4X), and it is general with respect to the up-sampling algorithm and the learning model, as long as it complies with the real-time prediction requirement. We analyse the results on 2D images and videos, on both raw and denoised signals, discussing the improvement of the denoising in terms of up-sampling accuracy, at the cost of a small loss of details on the US signal. Our method specialises trained networks to predict the high-resolution target through the design of the network architecture and the loss function, taking into account the anatomical district and the up-sampling factor and exploiting a large ultrasound data set. 

In future work, we want to extend the framework to US 3D images and perform with Esaote quality department and expert radiologists a clinical validation of the method through more formalised qualitative survey and evaluation methods~\cite{rahimi2022ct, ghulam2020clinical} through an interdisciplinary approach that involves engineering, medical science, physics, and computer science.

{\small{\paragraph{\textbf{Acknowledgements}} 
This work has been partially supported by the European Commission, NextGenerationEU, Missione 4 Componente 2, ``\emph{Dalla ricerca all’impresa}'', Innovation Ecosystem RAISE ``\emph{Robotics and AI for Socio-economic Empowerment}'', ECS00000035. Tests on CINECA Cluster are supported by the ISCRA-C Project HP10CVHIXD.}}
%
\bibliographystyle{alpha}
\bibliography{refs}

\newcommand{\etalchar}[1]{$^{#1}$}
\begin{thebibliography}{ANMM{\etalchar{+}}17}

\bibitem[AMP{\etalchar{+}}11]{alessandrini2011restoration}
Martino Alessandrini, Simona Maggio, Jonathan Por{\'e}e, Luca De~Marchi, Nicolo
  Speciale, Emilie Franceschini, Olivier Bernard, and Olivier Basset.
\newblock A restoration framework for ultrasonic tissue characterization.
\newblock {\em Transactions on Ultrasonics, Ferroelectrics, and Frequency
  Control}, 58(11):2344--2360, 2011.

\bibitem[ANMM{\etalchar{+}}17]{abdel2017breast}
Mohamed Abdel-Nasser, Jaime Melendez, Antonio Moreno, Osama~A Omer, and Domenec
  Puig.
\newblock Breast tumor classification in ultrasound images using texture
  analysis and super-resolution methods.
\newblock {\em Engineering Applications of Artificial Intelligence}, 59:84--92,
  2017.

\bibitem[ANO16]{abdel2016ultrasound}
Mohamed Abdel-Nasser and Osama~Ahmed Omer.
\newblock Ultrasound image enhancement using a deep learning architecture.
\newblock In {\em International Conference on Advanced Intelligent Systems and
  Informatics}, pages 639--649. Springer, 2016.

\bibitem[BGH20]{brown2020deep}
Katherine~G Brown, Debabrata Ghosh, and Kenneth Hoyt.
\newblock Deep learning of spatiotemporal filtering for fast super-resolution
  ultrasound imaging.
\newblock {\em Transactions on Ultrasonics, Ferroelectrics, and Frequency
  Control}, 67(9):1820--1829, 2020.

\bibitem[BLM{\etalchar{+}}08]{basarab2008method}
Adrian Basarab, Herv{\'e} Liebgott, Fabrice Morestin, Andrej Lyshchik, Tatsuya
  Higashi, Ryo Asato, and Philippe Delachartre.
\newblock A method for vector displacement estimation with ultrasound imaging
  and its application for thyroid nodular disease.
\newblock {\em Medical Image Analysis}, 12(3):259--274, 2008.

\bibitem[CGB19]{caballero2019deep}
Monica Caballero, Jon~Ander G{\'o}mez, and Aimilia Bantouna.
\newblock Deep-learning and hpc to boost biomedical applications for health
  (deephealth).
\newblock In {\em 2019 IEEE 32nd International Symposium on Computer-Based
  Medical Systems (CBMS)}, pages 150--155. IEEE, 2019.

\bibitem[CHH05]{clement2005superresolution}
GT~Clement, J~Huttunen, and K~Hynynen.
\newblock Superresolution ultrasound imaging using back-projected
  reconstruction.
\newblock {\em The Journal of the Acoustical Society of America},
  118(6):3953--3960, 2005.

\bibitem[CKH{\etalchar{+}}18]{choi2018deep}
Woosuk Choi, Mina Kim, Jae HakLee, Jungho Kim, and Jong BeomRa.
\newblock Deep cnn-based ultrasound super-resolution for high-speed
  high-resolution b-mode imaging.
\newblock In {\em International Ultrasonics Symposium}, pages 1--4. IEEE, 2018.

\bibitem[CNP22]{cammarasana2022real}
Simone Cammarasana, Paolo Nicolardi, and Giuseppe Patan{\`e}.
\newblock Real-time denoising of ultrasound images based on deep learning.
\newblock {\em Medical \& Biological Engineering \& Computing}, pages 1--16,
  2022.

\bibitem[C{\"O}MS19]{cuneyitouglu2019single}
Mine C{\"u}neyito{\u{g}}lu~{\"O}zkul, {\"U}nal~Erkan Mumcuo{\u{g}}lu, and
  {\.I}brahim~Tanzer Sancak.
\newblock Single-image bayesian restoration and multi-image super-resolution
  restoration for b-mode ultrasound using an accurate system model involving
  correlated nature of the speckle noise.
\newblock {\em Ultrasonic Imaging}, 41(6):368--386, 2019.

\bibitem[CP22]{cammarasana2022learning}
Simone Cammarasana and Giuseppe Patane.
\newblock Learning-based low-rank denoising.
\newblock {\em Signal, Image and Video Processing}, pages 1--7, 2022.

\bibitem[DGA{\etalchar{+}}17]{diamantis2017super}
Konstantinos Diamantis, Alan~H Greenaway, Tom Anderson, J{\o}rgen~Arendt
  Jensen, Paul~A Dalgarno, and Vassilis Sboros.
\newblock Super-resolution axial localization of ultrasound scatter using
  multi-focal imaging.
\newblock {\em IEEE Transactions on Biomedical Engineering}, 65(8):1840--1851,
  2017.

\bibitem[DZTN21]{ding2021ultrasound}
Jianrui Ding, Shili Zhao, Fenghe Tang, and Chunping Ning.
\newblock Ultrasound image super-resolution with two-stage zero-shot cyclegan.
\newblock In {\em Journal of Physics: Conference Series}, volume 2031, page
  012015. IOP Publishing, 2021.

\bibitem[EVW10]{ellis2010super}
Michael~A Ellis, Francesco Viola, and William~F Walker.
\newblock Super-resolution image reconstruction using diffuse source models.
\newblock {\em Ultrasound in medicine \& biology}, 36(6):967--977, 2010.

\bibitem[GG84]{geman1984stochastic}
Stuart Geman and Donald Geman.
\newblock Stochastic relaxation, gibbs distributions, and the bayesian
  restoration of images.
\newblock {\em Transactions on Pattern Analysis and Machine Intelligence},
  (6):721--741, 1984.

\bibitem[GKOS20]{ghulam2020clinical}
Qasam~M Ghulam, Sashi Kilaru, San-San Ou, and Henrik Sillesen.
\newblock Clinical validation of three-dimensional ultrasound for abdominal
  aortic aneurysm.
\newblock {\em Journal of Vascular Surgery}, 71(1):180--188, 2020.

\bibitem[GVV{\etalchar{+}}20]{groves2020automatic}
Leah~A Groves, Blake VanBerlo, Natan Veinberg, Abdulrahman Alboog, Terry~M
  Peters, and Elvis Chen.
\newblock Automatic segmentation of the carotid artery and internal jugular
  vein from 2{D} ultrasound images for 3{D} vascular reconstruction.
\newblock {\em International Journal of Computer Assisted Radiology and
  Surgery}, 15(11):1835--1846, 2020.

\bibitem[IZZE17]{isola2017image}
Phillip Isola, Jun-Yan Zhu, Tinghui Zhou, and Alexei~A Efros.
\newblock Image-to-image translation with conditional adversarial networks.
\newblock In {\em Conf. on Computer Vision and Pattern Recognition}, pages
  1125--1134, 2017.

\bibitem[JGB{\etalchar{+}}20]{jain2020localization}
Pankaj~K Jain, Saurabh Gupta, Arnav Bhavsar, Aditya Nigam, and Neeraj Sharma.
\newblock Localization of common carotid artery transverse section in {B}-mode
  ultrasound images using faster {RCNN}: a deep learning approach.
\newblock {\em Medical \& Biological Engineering \& Computing}, 58(3):471--482,
  2020.

\bibitem[KAR18]{khavari2018non}
Parviz Khavari, Amir Asif, and Hassan Rivaz.
\newblock Non-local super resolution in ultrasound imaging.
\newblock In {\em International Workshop on Multimedia Signal Processing},
  pages 1--6. IEEE, 2018.

\bibitem[Key81]{keys1981cubic}
Robert Keys.
\newblock Cubic convolution interpolation for digital image processing.
\newblock {\em Transactions on Acoustics, Speech, and Signal processing},
  29(6):1153--1160, 1981.

\bibitem[Lin04]{lingvall2004method}
Fredrik Lingvall.
\newblock A method of improving overall resolution in ultrasonic array imaging
  using spatio-temporal deconvolution.
\newblock {\em Ultrasonics}, 42(1-9):961--968, 2004.

\bibitem[LKO06]{lavarello2006regularized}
Roberto Lavarello, Farzad Kamalabadi, and William~D O'Brien.
\newblock A regularized inverse approach to ultrasonic pulse-echo imaging.
\newblock {\em Transactions on Medical Imaging}, 25(6):712--722, 2006.

\bibitem[LL18]{lu2018unsupervised}
Jingfeng Lu and Wanyu Liu.
\newblock Unsupervised super-resolution framework for medical ultrasound images
  using dilated convolutional neural networks.
\newblock In {\em 3rd International Conference on Image, Vision and Computing},
  pages 739--744. IEEE, 2018.

\bibitem[LLH{\etalchar{+}}21]{liu2021perception}
Heng Liu, Jianyong Liu, Shudong Hou, Tao Tao, and Jungong Han.
\newblock Perception consistency ultrasound image super-resolution via
  self-supervised cyclegan.
\newblock {\em Neural Computing and Applications}, pages 1--11, 2021.

\bibitem[LSK{\etalchar{+}}17]{lim2017enhanced}
Bee Lim, Sanghyun Son, Heewon Kim, Seungjun Nah, and Kyoung Mu~Lee.
\newblock Enhanced deep residual networks for single image super-resolution.
\newblock In {\em Proc. of the IEEE Conf. on Computer Vision and Pattern
  Recognition Workshops}, pages 136--144, 2017.

\bibitem[LTH{\etalchar{+}}17]{ledig2017photo}
Christian Ledig, Lucas Theis, Ferenc Husz{\'a}r, Jose Caballero, Andrew
  Cunningham, Alejandro Acosta, Andrew Aitken, Alykhan Tejani, Johannes Totz,
  Zehan Wang, et~al.
\newblock Photo-realistic single image super-resolution using a generative
  adversarial network.
\newblock In {\em Proceedings of the IEEE Conference on Computer Vision and
  Pattern Recognition}, pages 4681--4690, 2017.

\bibitem[MBK12]{morin2012alternating}
Renaud Morin, Adrian Basarab, and Denis Kouam{\'e}.
\newblock Alternating direction method of multipliers framework for
  super-resolution in ultrasound imaging.
\newblock In {\em International Symposium on Biomedical Imaging}, pages
  1595--1598. IEEE, 2012.

\bibitem[MBPK12]{morin2012post}
Renaud Morin, Adrian Basarab, Marie Ploquin, and Denis Kouam{\'e}.
\newblock Post-processing multiple-frame super-resolution in ultrasound
  imaging.
\newblock In {\em Medical Imaging: Ultrasonic Imaging, Tomography, and
  Therapy}, volume 8320, pages 433--440. SPIE, 2012.

\bibitem[NWY10]{ng2010solving}
Michael~K Ng, Pierre Weiss, and Xiaoming Yuan.
\newblock Solving constrained total-variation image restoration and
  reconstruction problems via alternating direction methods.
\newblock {\em Journal on Scientific Computing}, 32(5):2710--2736, 2010.

\bibitem[PBA{\etalchar{+}}19]{patel2019performance}
Trupesh~R. Patel, Sandeep Bodduluri, Thomas Anthony, William~S. Monroe,
  Pravinkumar~G. Kandhare, John-Paul Robinson, Arie Nakhmani, Chengcui Zhang,
  Surya~P. Bhatt, and Purushotham~V. Bangalore.
\newblock Performance characterization of single and multi {GPU} training of
  {U}-{N}et architecture for medical image segmentation tasks.
\newblock In {\em Proceedings of the Practice and Experience in Advanced
  Research Computing on Rise of the Machines (Learning)}. ACM, 2019.

\bibitem[PE14]{peleg2014statistical}
Tomer Peleg and Michael Elad.
\newblock A statistical prediction model based on sparse representations for
  single image super-resolution.
\newblock {\em Transactions on Image Processing}, 23(6):2569--2582, 2014.

\bibitem[PLLZ21]{pu2021automatic}
Bin Pu, Kenli Li, Shengli Li, and Ningbo Zhu.
\newblock Automatic fetal ultrasound standard plane recognition based on deep
  learning and {II}o{T}.
\newblock {\em IEEE Transactions on Industrial Informatics}, 17(11):7771--7780,
  2021.

\bibitem[RKFL22]{rahimi2022ct}
Aisyah Rahimi, Azira Khalil, Amir Faisal, and Khin~W Lai.
\newblock Ct-mri dual information registration for the diagnosis of liver
  cancer: A pilot study using point-based registration.
\newblock {\em Current Medical Imaging}, 18(1):61--66, 2022.

\bibitem[RR20]{rakotonirina2020esrgan+}
Nathana{\"e}l~Carraz Rakotonirina and Andry Rasoanaivo.
\newblock Esrgan+: Further improving enhanced super-resolution generative
  adversarial network.
\newblock In {\em IEEE Inter. Conf. on Acoustics, Speech and Signal
  Processing}, pages 3637--3641. IEEE, 2020.

\bibitem[SZ14]{simonyan2014very}
Karen Simonyan and Andrew Zisserman.
\newblock Very deep convolutional networks for large-scale image recognition.
\newblock {\em arXiv:1409.1556}, 2014.

\bibitem[SZA{\etalchar{+}}21]{schoen2021morphological}
Scott Schoen, Zhigen Zhao, Ashley Alva, Chengwu Huang, Shigao Chen, and Costas
  Arvanitis.
\newblock Morphological reconstruction improves microvessel mapping in
  super-resolution ultrasound.
\newblock {\em IEEE Transactions on Ultrasonics, Ferroelectrics, and Frequency
  Control}, 68(6):2141--2149, 2021.

\bibitem[TB20]{temiz2020super}
Hakan Temiz and Hasan~S Bilge.
\newblock Super resolution of b-mode ultrasound images with deep learning.
\newblock {\em Access}, 8:78808--78820, 2020.

\bibitem[TJ04]{taxt2004superresolution}
Torfinn Taxt and Radovan Jirik.
\newblock Superresolution of ultrasound images using the first and second
  harmonic signal.
\newblock {\em Transactions on Ultrasonics, Ferroelectrics, and Frequency
  control}, 51(2):163--175, 2004.

\bibitem[VEW07]{viola2007time}
Francesco Viola, Michael~A Ellis, and William~F Walker.
\newblock Time-domain optimized near-field estimator for ultrasound imaging:
  Initial development and results.
\newblock {\em IEEE Transactions on Medical Imaging}, 27(1):99--110, 2007.

\bibitem[VSSB{\etalchar{+}}19]{van2019deep}
Ruud~JG Van~Sloun, Oren Solomon, Matthew Bruce, Zin~Z Khaing, Yonina~C Eldar,
  and Massimo Mischi.
\newblock Deep learning for super-resolution vascular ultrasound imaging.
\newblock In {\em International Conference on Acoustics, Speech and Signal
  Processing}, pages 1055--1059. IEEE, 2019.

\bibitem[WYW{\etalchar{+}}18]{wang2018esrgan}
Xintao Wang, Ke~Yu, Shixiang Wu, Jinjin Gu, Yihao Liu, Chao Dong, Yu~Qiao, and
  Chen Change~Loy.
\newblock Esrgan: Enhanced super-resolution generative adversarial networks.
\newblock In {\em Proceedings of the European Conf. on Computer Vision}, 2018.

\bibitem[YFH20]{yu2020wide}
Jiahui Yu, Yuchen Fan, and Thomas Huang.
\newblock Wide activation for efficient image and video super-resolution.
\newblock In {\em British Machine Vision Conf.}, 2020.

\bibitem[YY18]{yoon2018deep}
Yeo~Hun Yoon and Jong~Chul Ye.
\newblock Deep learning for accelerated ultrasound imaging.
\newblock In {\em International Conference on Acoustics, Speech and Signal
  Processing}, pages 6673--6676. IEEE, 2018.

\bibitem[YZX12]{yu2012envelope}
Chengpu Yu, Cishen Zhang, and Lihua Xie.
\newblock An envelope signal based deconvolution algorithm for ultrasound
  imaging.
\newblock {\em Signal processing}, 92(3):793--800, 2012.

\bibitem[ZBKT15]{zhao2015joint}
Ningning Zhao, Adrian Basarab, Denis Kouame, and Jean-Yves Tourneret.
\newblock Joint bayesian deconvolution and pointspread function estimation for
  ultrasound imaging.
\newblock In {\em International Symposium on Biomedical Imaging}, pages
  235--238. IEEE, 2015.

\bibitem[ZWB{\etalchar{+}}16]{zhao2016single}
Ningning Zhao, Qi~Wei, Adrian Basarab, Denis Kouam{\'e}, and Jean-Yves
  Tourneret.
\newblock Single image super-resolution of medical ultrasound images using a
  fast algorithm.
\newblock In {\em International Symposium on Biomedical Imaging}, pages
  473--476. IEEE, 2016.

\end{thebibliography}
%
\begin{description}
\item[\textbf{Simone Cammarasana}] is research fellow at CNR-IMATI. He obtained a PhD in Computer Science at the University of Genova-DIBRIS, a post-lauream Master in Scientific Computing at the University of Sapienza-Roma, and a Master's degree in Engineering at the University of Pisa. His research interests include signals analysis, optimisation problems, and medical images.
\item[\textbf{Paolo Nicolardi}] is image processing and algorithms technical leader at Esaote. He obtained a Master degree in Engineering at Politecnico di Milano, in 2005. His research interests include image processing, computer vision, pattern recognition, machine learning, and medical images.
\item[\textbf{Giuseppe Patan\'e}] is senior researcher at CNR-IMATI. Since 2001, his research is mainly focused on Computer Graphics and Shape Modelling. He is the author of scientific publications in international journals and conference proceedings, and a tutor of PhD and Post.Doc students. He is responsible for R$\&$D activities in national and European projects.
\end{description}
\end{document}